\documentclass[runningheads]{llncs}

\usepackage{eccv}

\usepackage{eccvabbrv}

\usepackage{graphicx}
\usepackage{amsmath,amssymb} 

\usepackage{color}
\usepackage[ruled,vlined]{algorithm2e}
\usepackage{booktabs}
\usepackage{wrapfig}
\usepackage{needspace}
\usepackage{multirow}
\usepackage{xfrac}

\usepackage{mathrsfs}
\usepackage{xcolor}

\usepackage{float}

\usepackage[accsupp]{axessibility}

\usepackage{hyperref}

\usepackage{colortbl}

\definecolor{dominate}{RGB}{198, 239, 206}  \definecolor{win}{RGB}{255, 255, 255}        

\newcommand{\dom}[1]{\cellcolor{dominate}\textbf{#1}}
\newcommand{\win}[1]{\textbf{#1}}
\newcommand{\lose}[1]{#1}

\usepackage{orcidlink}
\newcommand{\orcidmobile}[1]{\orcidlink{#1}}

\begin{document}

\title{FrescoDiffusion: 4K Image-to-Video with Prior-Regularized Tiled Diffusion}

\titlerunning{FrescoDiffusion}

\author{
\mbox{Hugo Caselles-Dupré}\inst{1}\orcidmobile{0000-0003-2711-5732}\thanks{These authors contributed equally.} \and
\mbox{Mathis Koroglu}\inst{1,2}\orcidmobile{0009-0000-6589-3151}$^{\star}$ \and
\mbox{Guillaume Jeanneret}\inst{2}\orcidmobile{0000-0002-0055-7816}$^{\star}$ \and
\mbox{Arnaud Dapogny}\inst{2} \and
\mbox{Matthieu Cord}\inst{2,3}\orcidmobile{0000-0002-0627-5844}
}

\authorrunning{H.~Caselles-Dupré et al.}

\institute{Obvious Research, Paris, France \and
Sorbonne Université, CNRS, ISIR, F-75005 Paris, France \and
Valeo.ai, France}

\maketitle

\vspace{-0.5cm}

\begin{abstract}

Diffusion-based image-to-video (I2V) models have improved substantially, but they still struggle with ultra-high-resolution inputs such as 4K images. Native-resolution generation often loses fine details, while tiled denoising preserves local structure but disrupts global layout consistency. This issue is especially challenging for fresco animation, where large artworks contain many characters, objects, and semantically distinct regions that must remain spatially coherent over time.
We introduce FrescoDiffusion, a training-free method for coherent large-format I2V generation from a single complex image. Our approach augments tiled denoising with a precomputed latent prior: a low-resolution video is first generated at the model’s native resolution, then its latent trajectory is upsampled to provide a global spatiotemporal reference. During 4K generation, per-tile velocity predictions are fused with this reference at every diffusion step through a weighted least-squares objective, producing a closed-form update that improves global coherence while preserving fine detail.
We also introduce a spatial regularization variable for region-level motion control. Experiments on VBench-I2V and our fresco I2V dataset show that FrescoDiffusion improves consistency, fidelity, and controllability over tiled baselines while remaining computationally efficient.

\end{abstract}

\section{Introduction}
\label{sec:intro}

\begin{figure}[th!]
    \centering
\includegraphics[width=0.95\linewidth]{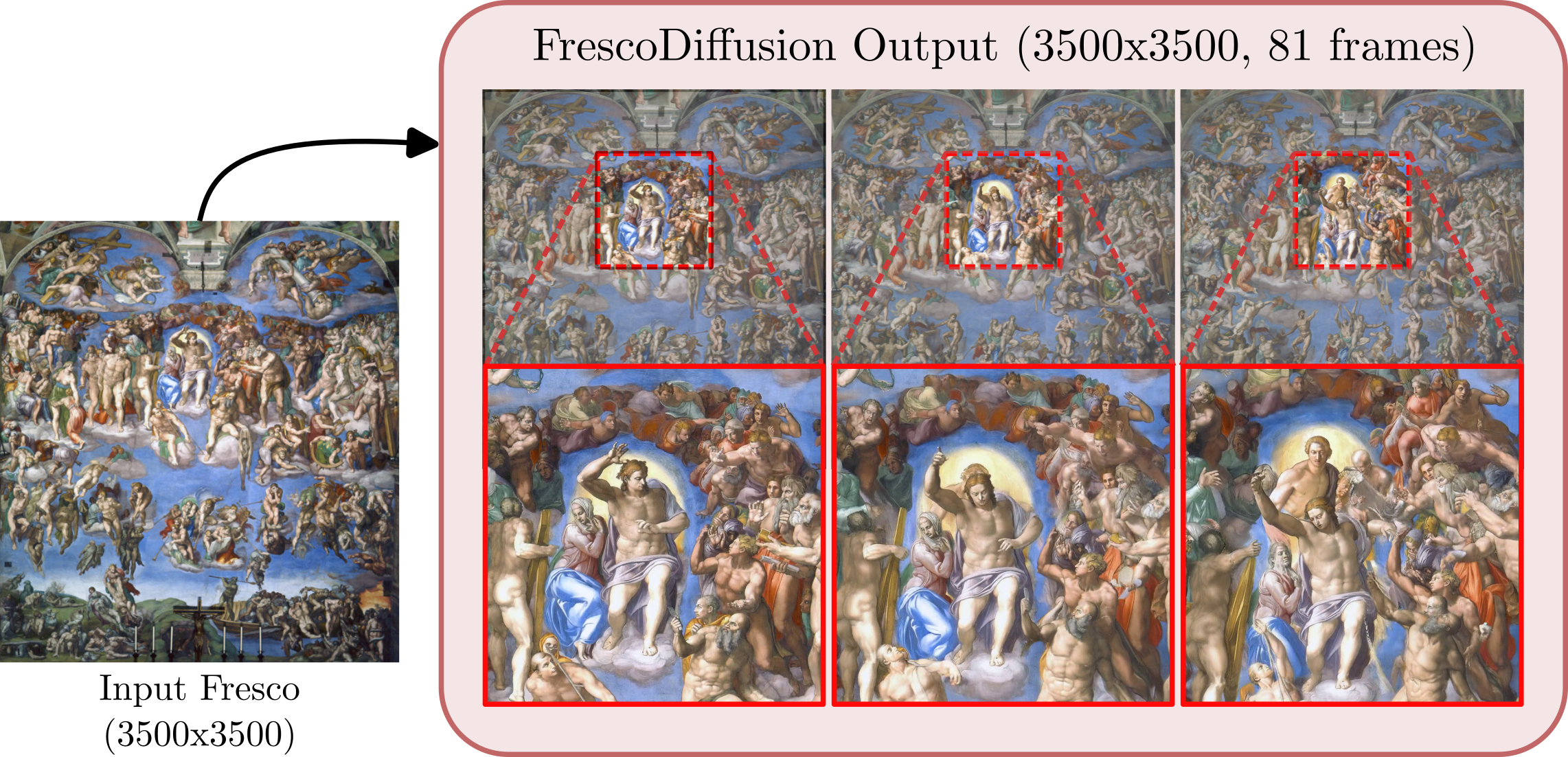}
    \caption{
From a single ultra-high-definition image ($3500\times3500$), FrescoDiffusion animates it at the same resolution.
    We show three frames from the generated video. The red box marks a fixed spatial region tracked across time, illustrating motion temporal consistency and fine-detail preservation.
    }
    \label{fig:teaser}
\end{figure}

Media is shifting toward ultra high-definition video, with resolutions around 4K ($3840\times2160$). In many applications, especially the creative industries such as cinema, animation, projections and art, there is a need for tools that can process and generate 4K content with sufficient detail. One example, and the focus of this work, is fresco animation. The same structural challenge also appears in projection mapping, animated films and motion comics, and urban scene simulation, although we evaluate only frescoes and VBench-I2V here.
In contrast to standard images, we define a \emph{\textbf{fresco}} as a large-scale image containing multiple scenes (\ie, up to hundreds) that blend seamlessly into a coherent visual.
To achieve animation of any picture, diffusion models~\cite{Singer2022MakeAVideo,rombach2022high,ho2022imagen,bar2024lumiere,batifol2025flux} have become the dominant approach for video synthesis, producing strong results in both text-to-video (T2V) and image-to-video (I2V) applications. While T2V models generate videos that faithfully follow an input prompt, I2V models follow the same paradigm but start with a first frame provided by the user.

Despite the progress in I2V, the most advanced models~\cite{wan2025wan,genmo2024mochi,hacohen2024ltx} operate under constraints that do not gracefully allow large-scale I2V. On the one hand, using regular video diffusion models on high-resolution images resized to their native spatial regime yields insufficient detail, as the input is too large and complex to be represented by the standard resolution of video diffusion models. On the other hand, memory and compute grow steeply with spatial and temporal dimensions, preventing the use of those models on 4K images.

Scaling image-to-video to very large videos containing many distinct regions remains largely unexplored in the literature. Existing work tries to address both issues, but the fresco setting makes it especially hard: simple tiling~\cite{bar2023multidiffusion,liu2025dynamicscaler} introduces cross-tile drift and visible seams, and post-hoc video super-resolution~\cite{he2024venhancer,liu2022video} cannot create scene-level content that never existed at the model’s native scale. A key difficulty specific to fresco animation is that different regions of the image play fundamentally different roles over time. Frescoes typically contain numerous loosely coupled scenes and characters: some regions are visually static and define architectural context, while others are semantically active and expected to exhibit motion. Rather than a simplified 4K-I2V setting, this exposes a distinct failure mode (adding local motion without breaking global structure) and motivates a region-aware treatment of global coherence.

In this paper, we introduce FrescoDiffusion to achieve 4K-I2V up to the fresco setting, as illustrated in \cref{fig:teaser}. Although animating the full-resolution fresco is not feasible directly, video models can animate a resized thumbnail of the same image at the model’s native scale. So, our core idea is to take advantage of this feature and use the resized thumbnail as a prior to guide the high-resolution denoising of the initial fresco with a novel loss. Our loss admits a closed-form solution that allows control of the balance between creativity (tiled denoising) and prior alignment. Moreover, we introduce a tool that adapts the generation process in active zones and the background differently with a mask to deal with frescoes specifically. It allows the generative model enough flexibility to animate the active zones while keeping the background close to the prior. In our experiments, we show the superiority, both in terms of quality and computation time, of FrescoDiffusion compared to other tiled-denoising methods through extensive qualitative and quantitative evaluation, and a user preference study, on our novel dataset FrescoArchive, and standard VBench 4K dataset~\cite{huang2025vbench++}. Videos generated with FrescoDiffusion are provided in the supplementary, with an interactive webpage for visualizing them conveniently.

To summarize, our contributions are as follows:

\begin{enumerate}
    \item We introduce FrescoDiffusion, a training-free 4K I2V method with a prior-regularized tile fusion applied at \emph{every} diffusion step. Our method leverages a closed-form solution of a newly introduced prior-based loss controlled with one interpretable strength $\lambda$.
    \item To demonstrate its use in an artistic domain, we propose a dynamic prior-strength schedule with a spatial gating mechanism that controls the trade-off between creativity and prior similarity, especially for fresco-to-video.
    \item We propose FrescoArchive, a new I2V dataset composed of complex multi-scene images for fresco-to-video evaluation.
\end{enumerate}

To contribute further to 4K-I2V, we will make our algorithm and dataset public with the paper release. Project page : \url{https://obvious-research.github.io/frescodiffusion/}.

\section{Related Work}
\label{sect:related}

\paragraph{Diffusion-based video generation.} Early text-to-video systems such as Make-A-Video~\cite{Singer2022MakeAVideo} and Imagen Video~\cite{ho2022imagen} generate short clips by cascading a base video diffusion model with spatial and temporal super-resolution modules. Subsequent work extends latent diffusion models~\cite{rombach2022high} to videos, e.g.\ Video Latent Diffusion Models (Video LDM)~\cite{blattmann2023stable}, which map videos to a compressed latent space for efficient high-resolution text-to-video generation, and Lumiere~\cite{bar2024lumiere}, whose space--time U-Net jointly denoises all frames. Recent foundation-scale models such as Wan~\cite{wan2025wan}, Mochi~\cite{genmo2024mochi} and LTX-Video~\cite{hacohen2024ltx} push video latent diffusion to higher quality and longer durations. Image-to-video methods (I2V) are usually supported by all these recent models. All these methods, including commercial systems (such as Kling, Sora2, Gen4.5 or Veo3), operate uniquely on their native spatial resolutions (typically $480$--$1080$p) and video lengths of a few seconds. In contrast, we formulate a training-free approach that takes an existing image-to-video diffusion model and extends it beyond its native resolution, bringing additional creative control.

\paragraph{Video super-resolution.}
Video super-resolution (VSR) seeks to reconstruct high-resolution videos from low-resolution inputs, typically with strong temporal coherence. 
While these methods~\cite{chan2021basicvsr,chan2022basicvsr++,chan2022investigating,liang2024vrt,zhou2024upscale,he2024venhancer,xie2025star}, including recent streaming diffusion VSR such as FlashVSR~\cite{flashvsr}, excel at producing globally coherent, temporally consistent videos, they are fundamentally constrained to staying close to the information present in the low-resolution input. Their objectives encourage fidelity to the input video under distortion and perceptual metrics, and any hallucinated detail is limited to local texture refinement. Starting from a 480p or 720p animation of a fresco, VSR can only upscale and slightly enrich this coarse representation; it cannot create hundreds of semantically distinct scenes that were never visible in the low-resolution video. In contrast, FrescoDiffusion performs tiled denoising directly on a large latent canvas and uses a thumbnail animation purely as a prior in latent space, allowing each tile to carry as much semantic content as a native-resolution video while maintaining global coherence.

\paragraph{Training-free high-resolution tiled denoising.}
A line of work studies generating large images and videos from pre-trained diffusion models without additional training. MultiDiffusion~\cite{bar2023multidiffusion} fuses overlapping diffusion trajectories via a weighted least-squares objective, enabling large-scale image generation with many scenes, but without explicit global coherence. Several approaches build upon this idea to improve spatial consistency. Mixture of Diffusers~\cite{jimenez2023mixture} runs multiple regional diffusions on a shared canvas to control high-resolution composition, while DiffCollage~\cite{zhang2023diffcollage} formulates generation as a factor graph over patches and overlaps. SpotDiffusion~\cite{frolov2025spotdiffusion} further reduces memory by denoising disjoint windows over time, trading overlap for efficiency. Recent training-free methods instead modify the sampling procedure of a single diffusion model to scale resolution. ScaleCrafter~\cite{he2023scalecrafter} and DemoFusion~\cite{du2024demofusion} progressively enlarge the effective receptive field through re-dilation, dispersed convolutions, or staged upscaling, treating the full canvas as a single sample rather than independent tiles. 

Closer to our work, DynamicScaler~\cite{liu2025dynamicscaler} (DS) proposes an offset-shifting denoising strategy for panoramic or $360^\circ$ video generation, where spatial windows are shifted across denoising steps to synchronize content and motion across fields of view.

It also relies on training-free tiling with global guidance from a low-resolution video, but differs from FrescoDiffusion in three ways. \textbf{(i)} DS injects the low-resolution prior \emph{once}, as a noisy initialization, whereas we regularize \emph{every} denoising step through a single quadratic objective. \textbf{(ii)} Our objective admits a single closed-form fused velocity via a per-step least-squares solution, whereas DS relies on heuristic window shifting and blending across steps rather than an explicit fusion objective. \textbf{(iii)} DS targets $360^\circ$ panoramas with a single shared global motion, while we allow for regional controllability of where motion should appear.

More broadly, FrescoDiffusion is related to latent-guidance methods such as Blended Latent Diffusion~\cite{avrahami2023blended}, which steer denoising toward a reference constraint. We instead regularize the fused velocity of an entire tiled video canvas and obtain a spatially varying closed-form update. Concurrent PixelWizard~\cite{li2026pixelwizard} learns a compact spatiotemporal anchor and a shortcut-trained native 2K/4K generator; FrescoDiffusion is training-free and reuses an existing I2V backbone.

\section{FrescoDiffusion Method}
\label{sec:method}

\begin{figure}[t!]
    \centering
\includegraphics[width=0.9\linewidth]{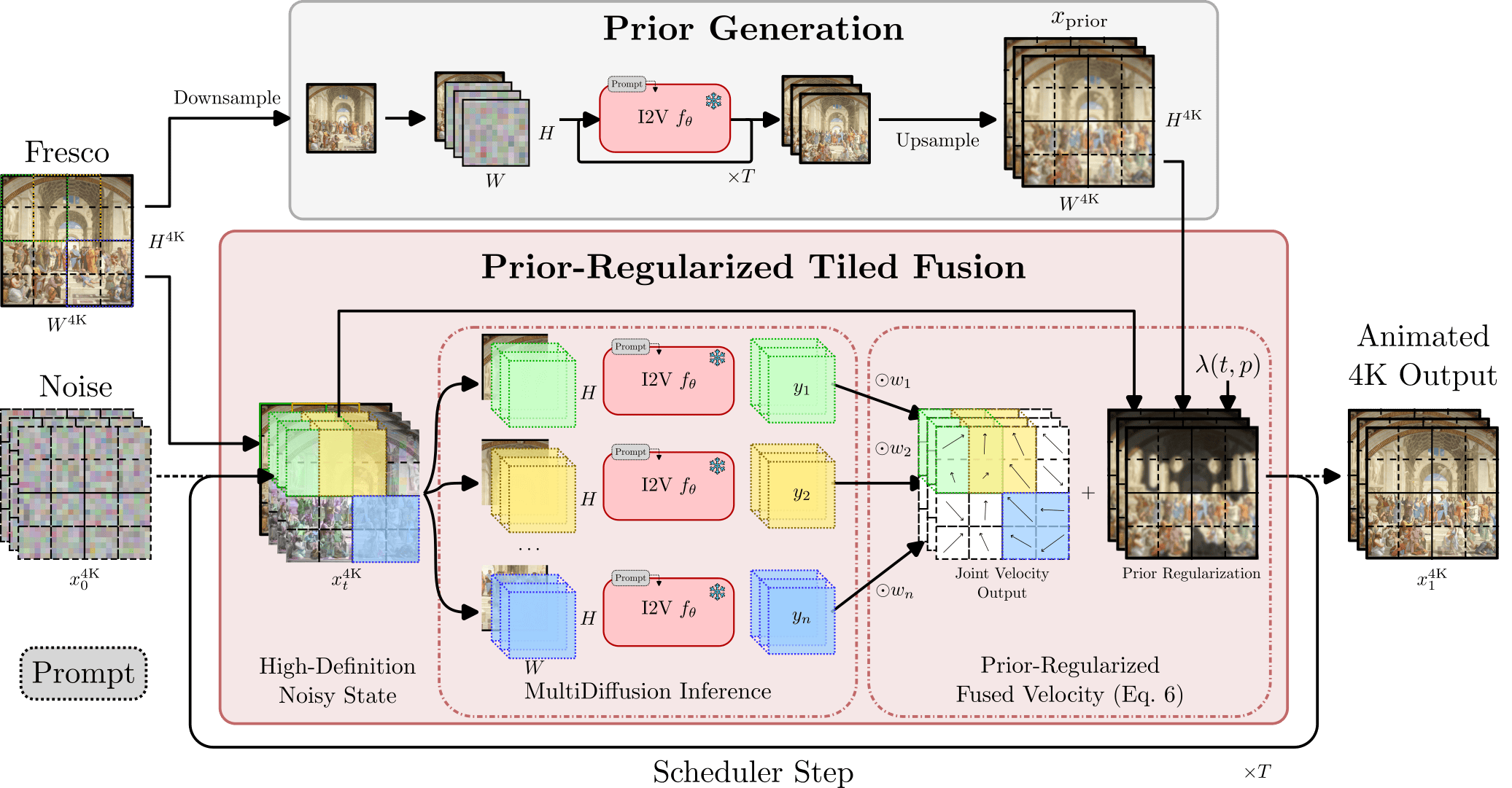}
    \caption{
    Overview of FrescoDiffusion. Starting from a 4K fresco image, we first build a global latent prior, by resizing the image to the native input size of the image-to-video backbone. Next, we upsample the prior latents \(x_{\text{prior}}\) to fit the 4K image size. We then apply tiled denoising to the large latent canvas, \(x_t^{\text{4K}}\), obtaining per-tile flow predictions, \(\{y_i\}\). We then use \(\{y_i\}\) and \(x_{\text{prior}}\) to compute the optimal output velocity field (\cref{eq:closedform}) according to our loss $\ell_{\text{FD}}$ (\cref{eq:f2v-energy}). This updated field is then used to update the large latent canvas, \(x_t^{\text{4K}}\), with the flow-matching scheduler. 
    }
    \label{fig:FrescoDiffusion}
\end{figure}

In this section, we introduce FrescoDiffusion, a method tailored for multi-scene 4K-I2V. Our proposed approach consists of two steps, as shown in \cref{fig:FrescoDiffusion}.  
First, we compute a prior thumbnail to guide the diffusion process. Second, during the denoising process, we analytically minimize the energy loss to produce the optimal fused output (\cref{sec:energy}). This output then guides the diffusion process. During this step, we optionally employ our novel masking strategy to direct the diffusion toward the prior by explicitly indicating which regions should be modified and which should converge to the prior (\cref{sec:lambda}).

\subsection{Ultra high-definition tiled denoising baseline framework}
\label{sec:md}

We start by introducing our baseline and notation.
Let $t\in[0,1]$ be the timestep, $c$ be the conditioning tuple (input image and prompt) for the I2V flow-matching model $f_\theta$, and $x_t\in\mathbb{R}^{C\times T\times H\times W}$ be the latent state. At each $t$, the model predicts a velocity field
\begin{equation}
    y\left(x_t\right)= f_\theta\left(x_t,t,c\right).
\end{equation}
Starting from $x_{t=0} \sim \mathcal{N}(0, I)$, a scheduler integrates these velocities up to $t=1$. The resulting latent $x_1$ is then decoded to produce the video.

Next, we introduce MultiDiffusion~\cite{bar2023multidiffusion} (MD), adapted to I2V as in prior work~\cite{liu2025dynamicscaler}. MD generates high-resolution latent codes,\allowbreak \ie $x_1^{\text{4K}} \in \allowbreak \mathbb{R}^{C \times T \times H^{\text{4K}} \times W^{\text{4K}}}$ with $H^{\text{4K}} \gg H$ and $W^{\text{4K}} \gg W$, by running $f_\theta$ on overlapping tiles of a large ``canvas'' $x_t^{\text{4K}}$ and merging the tile-wise predictions.

More specifically, let $x^\text{4K}_{0}$ be the initial 4K latent canvas. Let $\mathscr{C}_p$ \textit{crop} a window of shape $(C,T,H,W)$ at position $p$, and let $\mathscr{P}_p$ \textit{zero-pad} a tile back to shape $(C,T,H^\text{4K},W^\text{4K})$ at coordinates $p$. For each tile $i$, define the tile prediction
\begin{equation}\label{eq:crop-pad}
    y_i(x_t^\text{4K}) = \mathscr{P}_{p_i}\!\big(y\left(\mathscr{C}_{p_i}\left(x_t^\text{4K}\right)\right)\big),
\end{equation}
where $p_i$ is the coordinate of the $i^\text{th}$ tile. Given $x_t^\text{4K}$, and the tiled-velocity $\{y_i(x_t^\text{4K})\}_{i=1}^n$ predictions, MD solves for a single merged velocity $y^\star$ that best matches these overlapping tile predictions by minimizing the loss
\begin{equation}\label{eq:lmd}
    \ell_\text{MD}\left(y^\star;t\right) = \sum_{i=1}^{n}\big\|\,\sqrt{w_i}\odot\big(y^\star - y_i\left(x_t^\text{4K}\right)\big)\,\big\|_2^2,
\end{equation}
where $n$ is the number of windows, $p_i$ are the coordinates and $w_i$ are weight maps of window $i$ used to reduce seams between tiles.
This loss admits the closed-form solution
\begin{equation}\label{eq:md}
    y_{MD}\left(x_t^\text{4K}\right) = \frac{\displaystyle\sum_{i=1}^n w_i \odot y_i \left(x_t^\text{4K}\right)}{\displaystyle\sum_{i=1}^n w_i}.
\end{equation}
Finally, $y_{MD}$ is used to update the canvas, $x_t^\text{4K}$, using the iterative standard flow-matching sampling process.

\subsection{FrescoDiffusion: Prior-Regularized Tile Fusion}\label{sec:energy}
MD provides a solution for merging overlapping windows using a weighted sum. However, MD lacks the ability to regularize window merging with an existing prior, such as the initial frame, to create a cohesive scene.
To this end, we propose to extend $\ell_{\text{MD}}$ with a novel regularization term $\ell_{\text{prior}}$. Our new FrescoDiffusion loss is
\begin{equation}\label{eq:f2v-energy}
\ell_{\text{FD}}\left(y^\star; t\right)
\;=\; 
\underbrace{\big\|\,\sqrt{\lambda}\,\odot\,\left[\left(x_t^\text{4K} - \sigma_t y^\star\right) - x_{\text{prior}}\right]\big\|_2^2}_{\ell_\text{prior}(y^\star; t, x_\text{prior})} \;+\; \ell_\text{MD}\left(y^\star; t\right)
\end{equation}
Our loss has two terms: $\ell_\text{MD}$ reduces the disparity between the shifting-window outputs, while 
$\ell_\text{prior}$ minimizes the dissimilarity between the current-step prediction of the clean latent, \ie $(x_t^\text{4K} - \sigma_t y)$ in the flow matching formulation, and the prior $x_{\text{prior}} \in \mathbb{R}^{C\times T\times H^\text{4K}\times W^\text{4K}}$. Here, $\lambda$ is a regularization variable that can be either a constant ($\lambda\in\mathbb{R}$) or a tensor ($\lambda\in\mathbb{R}^{C \times T \times H^\text{4K} \times W^\text{4K}}$), whose design is discussed in the next section. Additionally, $\sigma_t$ denotes the scheduler’s discrete noise standard deviation at step $t$. Note that in other diffusion formulations, we just have to adapt the corresponding one-step prediction, see \cref{app:derivation-other}.

\Cref{eq:f2v-energy} is separable across canvas coordinates and strictly convex. Thus, the unique minimizer can be found in closed form by setting the derivative to zero. Therefore, the prior-regularized fused velocity is
\begin{equation}
y_{\text{FD}}(x_t^\text{4K})
\;=\;
\frac{\sigma_t\cdot\lambda\odot\,(x_t^\text{4K} - x_{\text{prior}})\;+\;\displaystyle\sum_{i=1}^n w_i\odot y_i(x_t^\text{4K})}
{\sigma_t^2 \cdot \lambda\;+\;\displaystyle\sum_{i=1}^n w_i}.
\label{eq:closedform}
\end{equation}
When $\lambda=0$, the closed-form solution in \cref{eq:closedform} reduces to the MultiDiffusion fusion in \cref{eq:md}. Prior errors enter the update proportionally to $\lambda$: a large value can propagate incorrect structure or motion from an unreliable thumbnail, whereas the decreasing schedule progressively releases this constraint. Reducing $\lambda_\text{base}$ or its cutoff moves the method toward MultiDiffusion when the prior is poor.

To create the global prior for $x_t^\text{4K}$, we resize the input fresco to the model’s native spatial size, generating a small image-to-video sequence. Then, we perform a per-frame trilinear upscale in latent space (to the large canvas size).
\cref{fig:FrescoDiffusion} illustrates FrescoDiffusion's generation process, on top of an algorithm in \cref{app:algo}.

\subsection{Spatio-Temporal Prior Strength Scheduling}\label{sec:lambda}

We will now discuss the design of the prior strength, $\lambda$, in \cref{eq:f2v-energy}. The parameter $\lambda$ addresses two objectives: (i) Remain structurally close to the prior while allowing the creation of new details.
(ii) Treat spatial regions differently in the case of frescoes. Background regions are supposed to remain structurally stable while other active regions should be animated. 

\begin{wrapfigure}{r}{0.47\columnwidth}
    \centering
    \includegraphics[width=\linewidth]{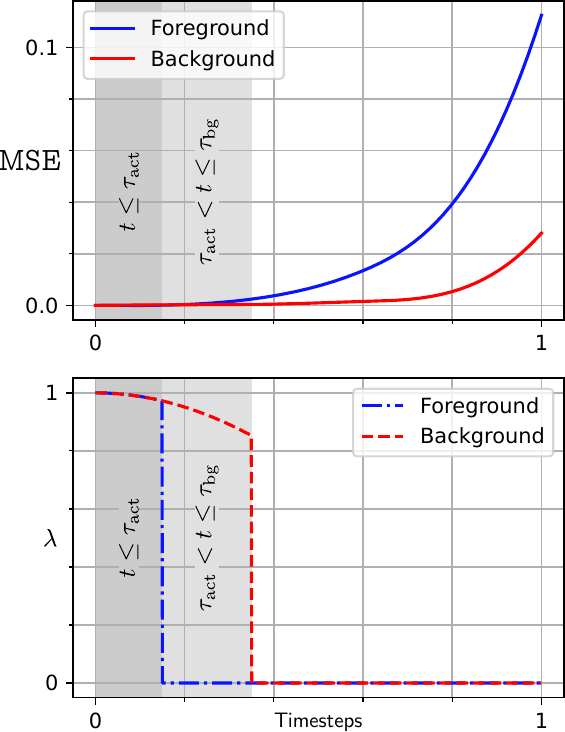}
    \caption{
    Spatial prior schedule $\lambda_R$ (\cref{eq:spatial-lambda}). (Top) MSE between the {\color{blue}foreground} / {\color{red}background} regions and the prior: the background stays bound to the prior longer than the foreground. (Bottom) Schedule for both regions.
    }
    \label{fig:spatial_lambda}
\end{wrapfigure}

To attain objective (i), we use a high $\lambda$ in the initial diffusion stages, keeping the model close to the prior, and decay it in the later stages to add detail. We thus model $\lambda$ as a global gated decreasing schedule of the diffusion step
\begin{equation}\label{eq:global-lambda}
\lambda_G(t,\tau)=\lambda_\text{base}\cdot\cos\!\left(t\frac{\pi}{2}\right)\cdot\mathbf{1}[t \le \tau]
\end{equation}
where $\tau$ is the gating and $\lambda_\text{base}$ is the strength of the regularization. When $\lambda = \lambda_G \in \mathbb{R}$, we name our method FrescoDiffusion. This schedule follows a clear principle: stay aligned to the prior while coarse structure forms, then decay to free fine detail. It introduces no free parameters tuned per image; a single fixed $(\lambda_\text{base},\tau)$ is reused across all images and datasets, which we find robust and interpretable. Both \cref{fig:spatial_lambda} Mean Square Error (MSE vs.\ $t$) and the component ablation in \cref{tab:vbench_scheduling} support each design choice.

To reach objective (ii), we compute a spatial activity map $A(p)$ to differentiate active zones from the background. Let \(A(p)\in\{0,1\}\) be a binary map, in which \(A(p)=1\) denotes active regions on the position $p$ (\eg, characters or local scenes expected to move) and \(A(p)=0\) denotes structurally static regions. Also, let \(\tau_{\text{act}}\) and \(\tau_{\text{bg}}\) be two temporal cutoffs, with \(\tau_{\text{act}} \leq \tau_{\text{bg}}\), which control the application of the prior to active and background regions, respectively. Hence, our prior strength factor becomes

\begin{equation}
\lambda_{R}\left(t,p\right) = 
\begin{cases} 
\lambda_{G}\left(t,\tau_{\text{act}}\right) & {\color{blue} \text{if}\;A\left(p\right)=1\;\text{(pixel $p$ in the foreground)}} \\ 
\lambda_{G}\left(t,\tau_{\text{bg}}\right) & {\color{red}\text{if}\;A\left(p\right)=0\;\text{(pixel $p$ in the background)}} 
\end{cases}
\label{eq:spatial-lambda}
\end{equation}

Please note that $\lambda_R$ is a tensor with shape $(C \times T \times H^\text{4K} \times W^\text{4K})$. We refer to this variant ($\lambda=\lambda_R$) as Regional-FrescoDiffusion (R-FrescoDiffusion).

\Needspace{4\baselineskip}

This design is motivated by \cref{fig:spatial_lambda}, which plots the MSE between the noised prior and the current latents $x_t^\text{4K}$. The gate enforces global coherence early ($t \le \tau_{\text{act}}$), then relaxes the prior first in active regions ($\tau_{\text{act}} < t \le \tau_{\text{bg}}$) to let motion and detail emerge, while background regions stay constrained longer. 
In late steps ($t > \tau_{\text{bg}}$), the prior is fully disabled everywhere, and sampling focuses purely on fine detail refinement. The end result is that the background MSE is much closer to the prior compared to the foreground. 
The coefficient \(\lambda_\text{base}\ge 0\) controls how strongly we adhere to the prior versus letting tiled denoising add new detail. FrescoDiffusion ($\lambda_G$) is our general default. R-FrescoDiffusion helps most when localized actors move against a stable background; with globally coupled motion or no reliable activity map, we use the global variant. When $\tau_{\text{act}}=\tau_{\text{bg}}$, $\lambda_R=\lambda_G$ and the regional variant reduces exactly to FrescoDiffusion. R-FrescoDiffusion does not require a deliberately static prior: the background stability emerges from the longer cutoff $\tau_{\text{bg}}$, and a dynamic prior remains compatible.

\section{Dataset and Evaluation Protocols}
\label{sec:dataset}

\subsubsection{Dataset}

We use the \textit{Image Suite} of VBench~\cite{huang2025vbench++}, as our first 4K-I2V dataset. Such datasets focus on one or a few objects, and we are looking for frescoes, \ie complex images with multiple intricate scenes, to evaluate our method thoroughly. Therefore, we propose a new dataset named \emph{FrescoArchive} for I2V techniques on a fresco scale. Starting with the LAION-2B Aesthetic Subset~\cite{schuhmann2022laion}, we filtered the images based on criteria such as pixel count, aesthetics, watermarks and NSFW scores. Next, we performed text-based filtering, followed by zero-shot classification to detect frescoes. Subsequently, we deduplicated~\cite{idealods2019imagededup} the dataset and generated captions using Qwen3-VL-32B~\cite{bai2025qwen3} with both the image and the LAION caption. Ultimately, we manually selected 371 pairs to achieve the best possible image-caption match. We provide details on this process in \cref{app:dataset}, along with statistics. This dataset will be used exclusively for validation.

\subsubsection{Evaluation Metrics}

We employ a user study at full resolution to quantify the human preference over the baselines. 
We used Amazon Mechanical Turk to conduct the study. Participants were carefully filtered to avoid bots and lower quality evaluators: we required a masters status and a task approval rate superior to $85\%$ to enroll.
They were shown pairs of videos generated with two concurrent methods, from the same input image, drawn from both the FrescoArchive and VBench-I2V datasets. Videos were displayed at identical resolution and duration, with randomized ordering and no method identification. We report preference percentages with 95\% confidence intervals. For each pair, participants answered two binary-choice questions: 
\begin{itemize}
\item Animation Fidelity: Which video most closely resembles a fresco artwork that has been smoothly and naturally animated?
\item Motion plausibility: Which video provides the most convincing animation of the input image, with appropriate and perceptible motion?
\end{itemize}

We also evaluate our method on the VBench protocols~\cite{huang2023vbench,zheng2025vbench2,huang2025vbench++}, following standard studies~\cite{liu2025dynamicscaler}. These metrics compute the similarity between the input image and each frame, as well as the similarity between consecutive frames. These protocols evaluate several criteria: Subject Consistency, Motion Smoothness, Aesthetic Score, and Imaging Quality. We complement it using VBench's I2V metrics (Image-to-Video Subject Consistency and Image-to-Video Background Consistency), to measure the similarity between input image and video.

While VBench is the standard for 480p/1080p, it provides incomplete 4K-I2V evaluations because it downsizes videos to fit metric models' requirement (DINOv2~\cite{oquab2023dinov2} and CLIP~\cite{radford2021learning}).
We thus complement our testbed with three metrics that specifically target 4K-I2V. (i) We perform standard sharpness measures at full scale to quantify fine-detail generation using the Tenengrad~\cite{pertuz2013analysis} function. (ii) We use a simple yet efficient Temporal Consistency metric, measuring the mean square error between consecutive downsized frames to quantify differences between frames. (iii) We compute a prior similarity metric using DINOv3~\cite{simeoni2025dinov3}, which is not limited to 1080p resolution compared to DINOv2 used in VBench.

\section{Experiments}

\subsection{Implementation Details and Baselines}

\paragraph{Video generation backbone.} All experiments use the 14B-parameter Wan2.2-I2V~\cite{wan2025wan} model. We generate 81 frames at 16 fps with six TurboDiffusion~\cite{zhang2025turbodiffusionacceleratingvideodiffusion} steps and guidance scale 1.0. FrescoDiffusion and MultiDiffusion use $480\times832$ pixel tiles with 30\% overlap. We select $\lambda_\text{base}=1.5$ as a central operating point in the $[0.01,3]$ sweep in \cref{fig:quant_metrics}, then freeze it and the $\tau$ cutoffs for every image and both datasets. Exact schedules, strides, fusion weights, and resolution handling are reported in \cref{app:fresco-implem-details}.

\paragraph{Baselines.} We compare our method to three tiled-diffusion methods: MultiDiffusion~\cite{bar2023multidiffusion}, DemoFusion~\cite{du2024demofusion} (state-of-the-art tiled image diffusion method adapted to a video setup), and DynamicScaler~\cite{liu2025dynamicscaler} (state-of-the-art tiled video diffusion method). All implementation details are available in \cref{app:implem-details}.
For a fair backbone comparison, we adapt every baseline to Wan2.2 and use identical prompts, sampling steps, guidance scales, seeds, output resolutions, and frame counts. Tile geometry is shared where each algorithm permits; method-specific settings follow the authors' released implementations and are listed in \cref{app:implem-details}. We did not perform independent per-baseline hyperparameter searches after transfer to Wan2.2, so the results compare documented default adaptations rather than separately tuned optima. No method is tuned per test image.

\paragraph{Spatial activity maps.}
\label{sec:setup-sam}

\begin{wrapfigure}{r}{0.4\linewidth}
    \centering
    \includegraphics[width=\linewidth]{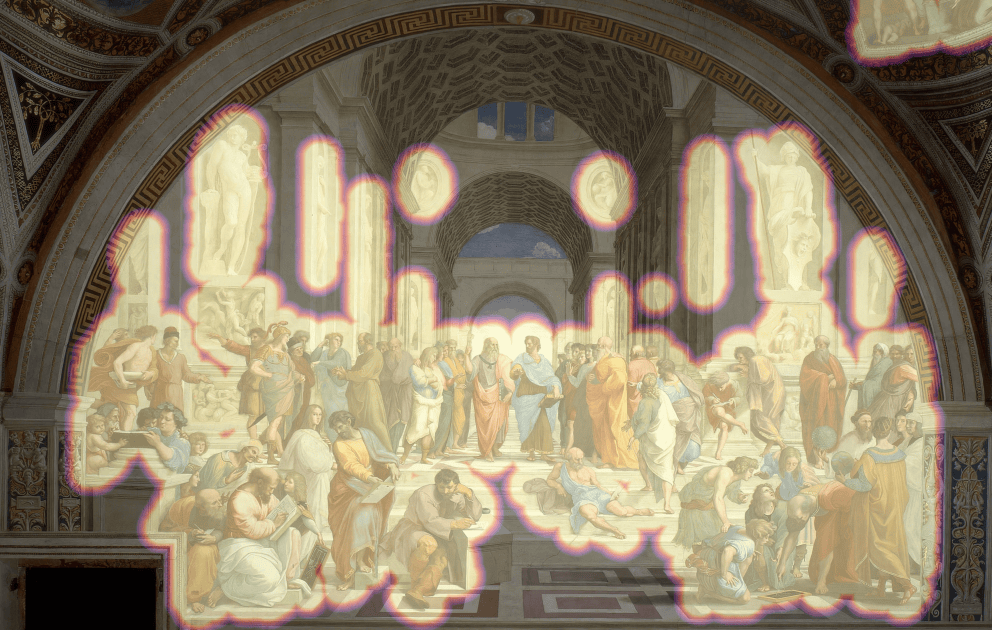}
    \caption{
    Overlay of the spatial activity map onto the input fresco.
    }
    \label{fig:sam3}
\end{wrapfigure}

R-FrescoDiffusion uses a spatially gated prior schedule to differentiate active regions from structurally static background.
For each input image, we compute the activity map $A(p)$ (see \cref{sec:lambda}) using the Segment Anything Model 3 (SAM3) \cite{carion2025sam}, as illustrated in \cref{fig:sam3}. We apply SAM3 with a fixed set of prompts producing a binary activity map in $[0,1]$, which is downsampled to latent resolution and used directly in Eq.~\eqref{eq:spatial-lambda}. See details in \cref{app:sam3}.
These activity maps are computed once per input image and remain fixed throughout sampling, without additional learnable parameters.

\subsection{Qualitative Evaluation}
\begin{figure}[ht!]
    \centering
\includegraphics[width=\linewidth]{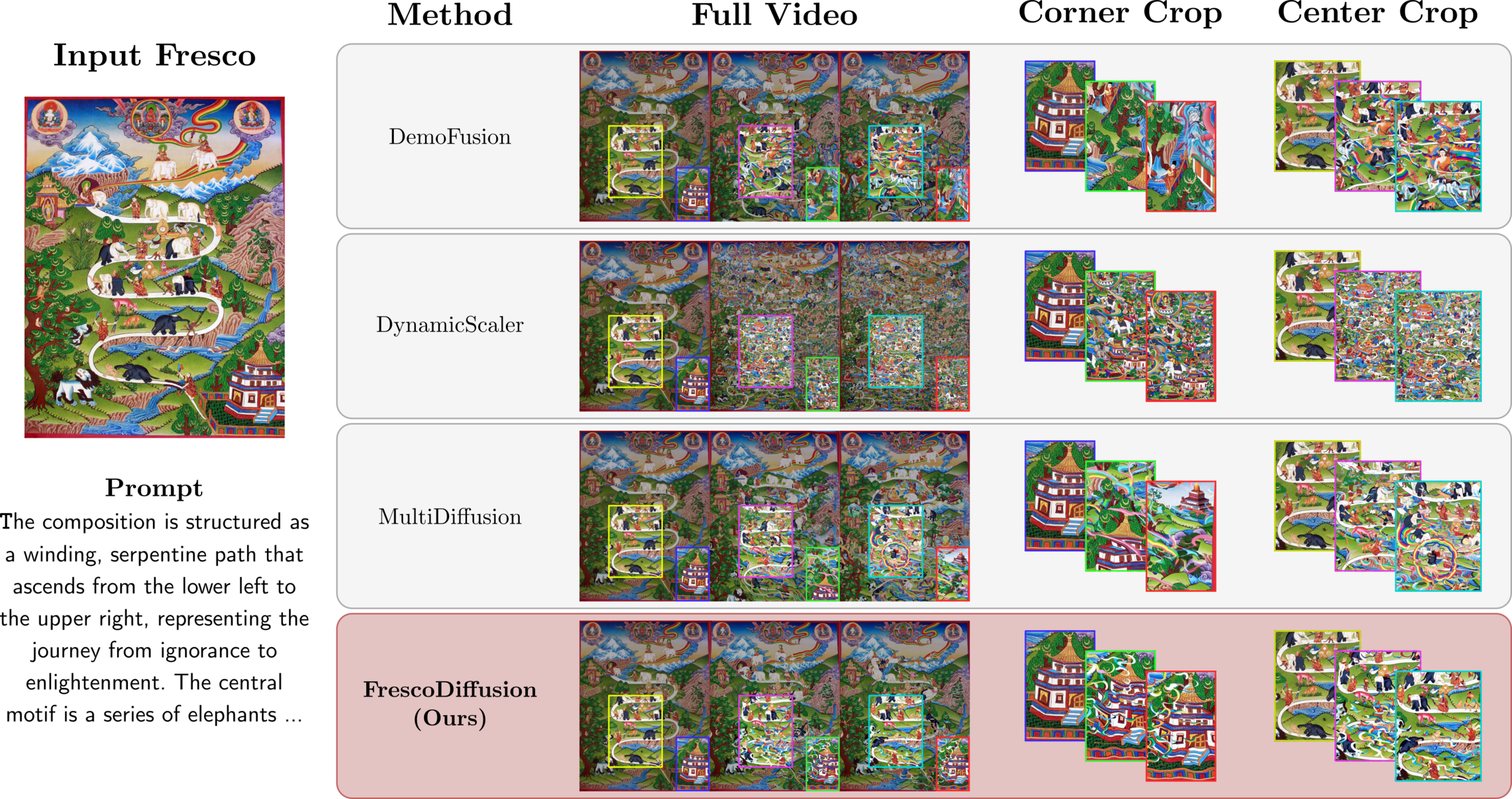}
\caption{
    A qualitative comparison of fresco-scale inputs. FrescoDiffusion generates coherent global scenes and animates details at a local level. By contrast, DemoFusion, DynamicScaler and MultiDiffusion only manage to produce either coherent scenes or high-quality details, but not both. }
    \label{fig:qualitative}
\end{figure}

We begin with a qualitative study. In \cref{fig:qualitative}, we show the first, 40th, and last frame for each model, with fixed-coordinate central and corner crops that expose fine-grained temporal changes at full output resolution.
DemoFusion preserves the global structure. Yet, some elements are modified, such as the path in the center crop and the hut in the corner crop. Visible wobbling is present when the video is playing. MultiDiffusion and DynamicScaler tend to introduce excessive novel content across tiles or sampling stages, which results in accumulation of structural inconsistencies and loss of temporal coherence. In contrast, FrescoDiffusion produces high quality videos while conserving the general layout of the video. Later, in \cref{sec:control}, we discuss the differences between FrescoDiffusion and its regional counterpart from a qualitative perspective. 
We also contrast with video super-resolution (VSR): methods such as FlashVSR~\cite{flashvsr} are tied to the trajectory of the base video at its native resolution, so they sharpen texture but cannot introduce the multi-scene content and motion that FrescoDiffusion synthesizes. VSR is therefore not adapted to our task: it can refine an already-determined animation but cannot animate a static fresco. Further full-resolution examples, a VSR comparison, and quantitative motion analysis are reported in \cref{app:extra-results,fig:vsr-comparison,app:motion-quantity}.

\subsection{State-of-the-Art quantitative comparison}

\subsubsection{High-Resolution Evaluation: User Study.}

We start with a user study to quantitatively measure human preference, run on both FrescoArchive (FA) and VBench-I2V. \Cref{tab:user_study} shows that both of our methods are strongly preferred over DynamicScaler and MultiDiffusion on both datasets, with average preference rates of 76\% to 84\% on FA and 61\% to 69\% on VBench-I2V, confirming that these baselines produce noticeably lower quality animations. Against the stronger DemoFusion baseline, the two variants are complementary: on FA, FrescoDiffusion is significantly preferred (62\% average), reflecting its more perceptible motion on frescoes, while on VBench-I2V R-FrescoDiffusion wins (62\% average) on more structured content. This split mirrors their design: FrescoDiffusion is the general default, and the regional schedule of R-FrescoDiffusion helps most when foreground and background play distinct roles. Both observations are consistent across the motion and fidelity questions.

\begin{table}[ht]
  \centering
  \caption{User study results: human preference rates (\% of annotators preferring our method over each baseline) on both FrescoArchive (FA; 371 source images, 1344 ratings) and VBench-I2V (361 source images, 1465 ratings). Each pair is drawn from the same source image, shown at identical resolution and duration in randomized order. \colorbox{dominate}{Green} cells mark a statistically significant preference ($>50\%$ at the 95\% level); all rates use 95\% confidence intervals of at most $\pm6\%$.}
  \resizebox{\linewidth}{!}{\begin{tabular}{l cccccc cccccc}
  \toprule
  & \multicolumn{6}{c}{\textit{FrescoArchive}}
  & \multicolumn{6}{c}{\textit{VBench-I2V}} \\
  \cmidrule(lr){2-7}\cmidrule(lr){8-13}
  & \multicolumn{3}{c}{FrescoDiffusion} & \multicolumn{3}{c}{R-FrescoDiffusion}
  & \multicolumn{3}{c}{FrescoDiffusion} & \multicolumn{3}{c}{R-FrescoDiffusion} \\
  \cmidrule(lr){2-4}\cmidrule(lr){5-7}\cmidrule(lr){8-10}\cmidrule(lr){11-13}
  & Motion & Fidelity & Avg.
  & Motion & Fidelity & Avg.
  & Motion & Fidelity & Avg.
  & Motion & Fidelity & Avg. \\
  \midrule
  vs.\ DemoFusion
  & \dom{63\%} & \dom{60\%} & \dom{62\%}
  & \win{54\%} & \dom{62\%} & \dom{58\%}
  & \win{55\%} & \lose{46\%} & \win{51\%}
  & \dom{66\%} & \dom{57\%} & \dom{62\%} \\
  vs.\ DynamicScaler
  & \dom{78\%} & \dom{80\%} & \dom{79\%}
  & \dom{88\%} & \dom{80\%} & \dom{84\%}
  & \dom{57\%} & \dom{64\%} & \dom{61\%}
  & \dom{68\%} & \dom{64\%} & \dom{66\%} \\
  vs.\ MultiDiffusion
  & \dom{78\%} & \dom{74\%} & \dom{76\%}
  & \dom{78\%} & \dom{74\%} & \dom{76\%}
  & \dom{61\%} & \dom{66\%} & \dom{64\%}
  & \dom{73\%} & \dom{64\%} & \dom{69\%} \\
  \bottomrule
  \end{tabular}}
  \label{tab:user_study}
  \end{table}

\subsubsection{Standard Low-Resolution I2V Metrics}

Next, as a sanity check, we compare our approach with modern methods (\cref{tab:quant_results}) using standard lower-resolution I2V metrics (VBench and VBench-I2V). These metrics are not designed for 4K videos as the evaluated videos are downscaled aggressively ($\sim10-20\times$) to fit the backbones' resolution. As a result, the reported scores provide only a coarse proxy for performance at the original 4K resolution. We perform this evaluation on the FrescoArchive and 4K \textit{Image Suite} VBench dataset, using both regional and non-regional configurations. 
In the FrescoArchive dataset, our method slightly outperforms the baselines on average.
On the \textit{Image Suite} VBench dataset, our approach performs second best on average, slightly outperformed by DemoFusion. Full results for each metric are available in \cref{sec:full-vbench}. Note that for DynamicScaler, our re-implementation performs better, as the I2V backbone in their original implementation, VideoCrafter~\cite{chen2023videocrafter1}, is much older and clearly underperforms Wan2.2. Our conclusion on this benchmark is consistent with what we observed in our user study and qualitatively: we outperform all methods on the fresco-to-video task and are competitive with DemoFusion on the 4K-I2V task.

\begin{table}
    \centering
    \caption{Standard low-resolution I2V metrics. The table shows the average performance (higher is better) using the VBench evaluation suite on both FrescoArchive and VBench-I2V image sets. We also display the average generation time in minutes.}
    \begin{tabular}{l|c|c|c|c}\toprule
    Method                        & FrescoArchive     & VBench-I2V        & Time (min)            & Rel.\ Time \\\midrule
    DynamicScaler (original)      & 0.857             & 0.862             & 18.45                 & 2.27$\times$ \\
    DynamicScaler$^{*}$ (CVPR'25) & 0.871             & 0.865             & 10.25                 & 1.26$\times$ \\
    DemoFusion$^{*}$ (CVPR'24)    & 0.903             & \textbf{0.879}    & 13.5                  & 1.66$\times$ \\\midrule
    MultiDiffusion$^{*}$ (ICML'23)& 0.876             & 0.860             & \textbf{8.15}         & \textbf{1.00}$\times$ \\
    FrescoDiffusion               & \underline{0.904} & 0.875             & \underline{8.58}      & \underline{1.05$\times$} \\
    R-FrescoDiffusion             & \textbf{0.907}    & \underline{0.878} & 9.08                  & 1.11$\times$ \\\bottomrule
\end{tabular}
    \label{tab:quant_results}
\end{table}

\subsubsection{Computational efficiency.} We report average wall-clock time over all runs on both datasets on a single H100 GPU. FrescoDiffusion and R-FrescoDiffusion remain within 5\% and 11\% of MultiDiffusion runtime, respectively, while reducing generation time by 36\% and 33\% relative to DemoFusion (13.5 vs.\ 8.58/9.08 minutes). Our Wan2.2 DynamicScaler adaptation is faster than the original but remains slower than both variants.

\subsection{Controlling FrescoDiffusion}\label{sec:control}

\subsubsection{Ablation of the prior strength schedule.}

\begin{figure}[ht]
    \centering
    \includegraphics[width=0.9\linewidth]{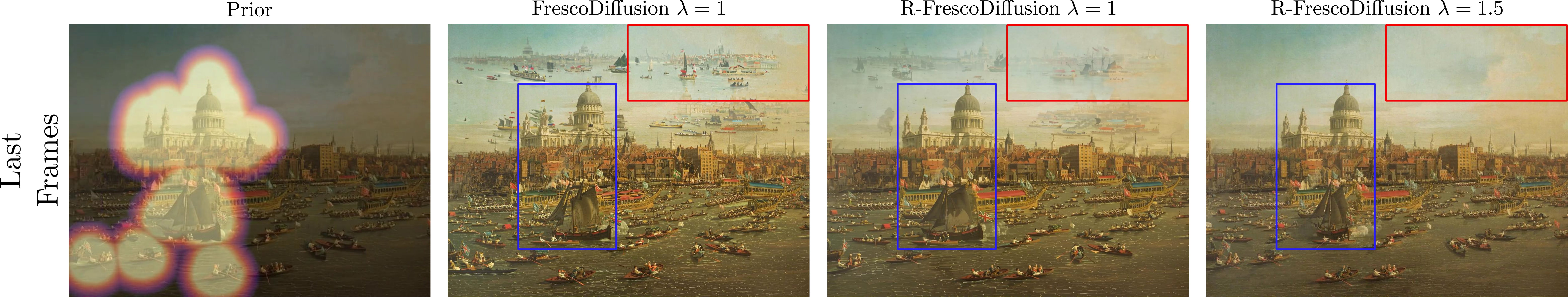}
\caption{Regional constraint. Prior shows an overlay of the activity map. The {\color{red}red}/{\color{blue}blue} boxes represent background/foreground regions. Our regional loss pulls the background toward the prior while allowing new details in foreground regions.}
    \label{fig:fd-vs-rfd}
\end{figure}

We perform an ablation study of the spatio-temporal prior strength schedule's design, $\lambda(t,p)$, using VBench metrics on FrescoArchive. We start with MultiDiffusion (no $\lambda$) and add a constant regularization $\lambda=\lambda_\text{base}=1.5$. This actually worsens performance. Next, we add the cosine schedule $\lambda=\lambda_\text{base} \, \cos(t\frac{\pi}{2})\in\mathbb{R}$, and obtain significant gains over MD. Then, we build FrescoDiffusion by setting $\lambda = \lambda_G$ (see \cref{eq:global-lambda}), and we finish with R-FrescoDiffusion by setting $\lambda=\lambda_R$ (see \cref{eq:spatial-lambda}). This results in the best measured performance. Crucially, the global prior-regularized fusion \emph{alone, without any spatial mask} ($\lambda_G$) already lifts MultiDiffusion from $0.876$ to $0.904$ ($+0.028$ avg), whereas the regional SAM3 mask adds a smaller further $+0.003$ avg ($+0.019$ SC). The prior-regularization term, not the mask, therefore drives most of the gain; R-FrescoDiffusion is a complementary control axis, consistent with the user study, where its advantage concentrates on content that requires more creativity. Qualitatively, we show the difference between FrescoDiffusion and R-FrescoDiffusion in \cref{fig:fd-vs-rfd}. When adding our regional regularization, R-FrescoDiffusion is more similar to the prior on background regions, as intended. We provide further visualization of that effect in \cref{app:extra-results}.

\begin{table}[ht]
  \centering
  \caption{Prior strength schedule ablation study on FrescoArchive. Best in \textbf{bold}. The results suggest that including the schedule, the gating, and the spatial regularization enhances the quantitative performance.}
  \label{tab:vbench_scheduling}
  \resizebox{0.9\linewidth}{!}{\begin{tabular}{l|c|cccccc|c}
    \toprule
    Method & $\lambda$ function & SC & MS & A & I & ISC & IBC & Avg \\
    \midrule
    MultiDiffusion  & $0$  & 0.876          & 0.974          & 0.686          & 0.754 & 0.981          & 0.989          & 0.876 \\ \midrule
    FrescoDiffusion & $\lambda_\text{base}$ & 0.942 & \textbf{0.991} & 0.645 & 0.598 & 0.979 & 0.983 & 0.856 \\
    FrescoDiffusion & $\lambda_\text{base} \, \cos(t\frac{\pi}{2})$ & 0.946 & \textbf{0.991} & 0.724 & 0.730 & 0.987 & 0.992 & 0.895 \\
    FrescoDiffusion & $\lambda_G$ (\cref{eq:global-lambda}) & 0.958 & 0.990 & \textbf{0.738} & \textbf{0.753} & \textbf{0.991} & \textbf{0.995} & 0.904 \\
    R-FrescoDiffusion & $\lambda_R$ (\cref{eq:spatial-lambda})  & \textbf{0.977} & \textbf{0.991} & 0.736 & \textbf{0.753} & \textbf{0.991} & 0.994 & \textbf{0.907} \\
    \bottomrule
  \end{tabular}}
\end{table}

\subsubsection{$\lambda$-Controlled Pareto Trade-off Between Creativity and Prior Similarity}

\begin{figure}[h]
    \centering
    \begin{subfigure}[t]{0.49\linewidth}
        \centering
        \includegraphics[width=\linewidth]{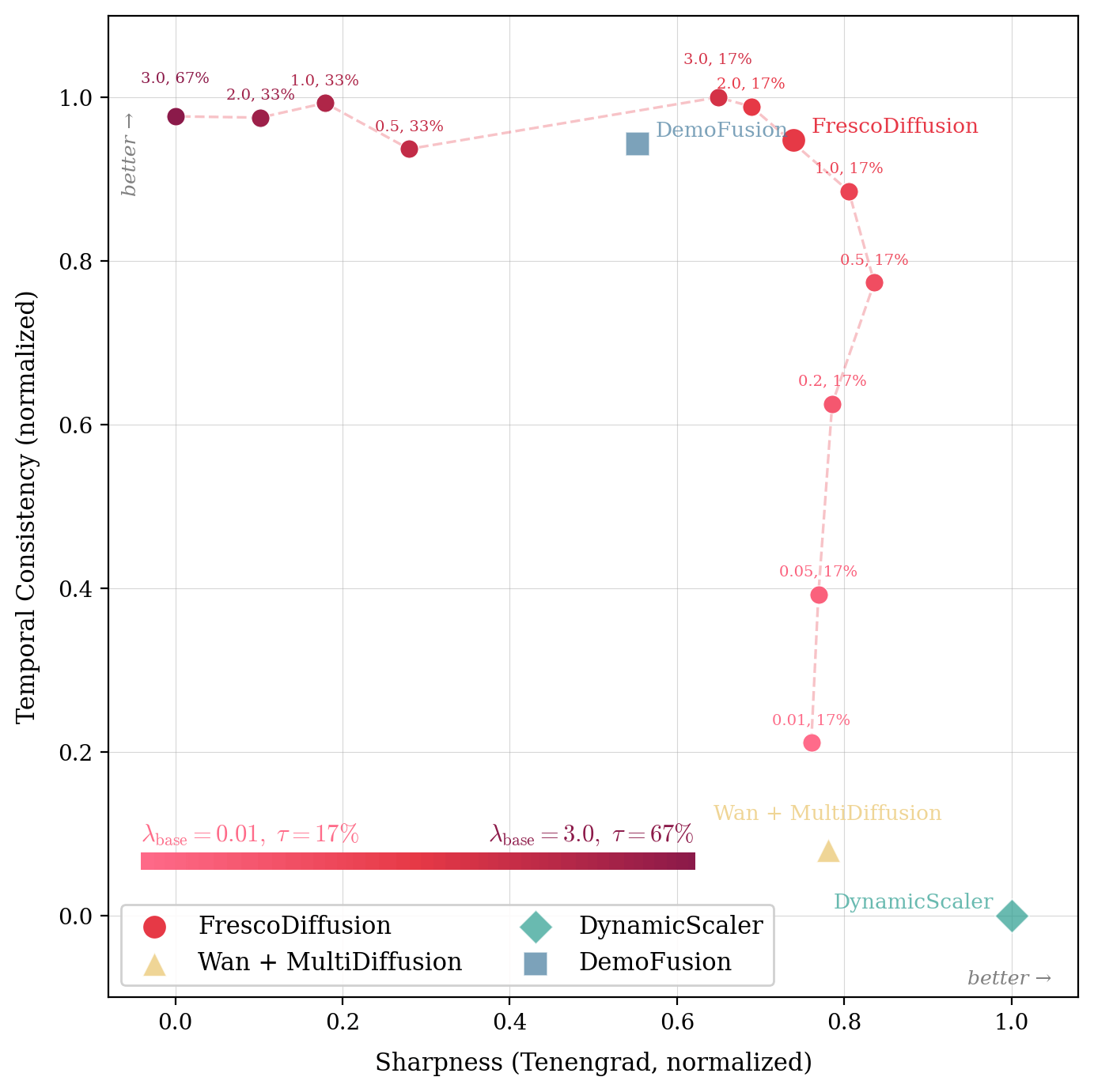}
        \caption{Temporal consistency versus sharpness.}
        \label{fig:quant_metrics_a}
    \end{subfigure}
    \hfill
    \begin{subfigure}[t]{0.49\linewidth}
        \centering
        \includegraphics[width=\linewidth]{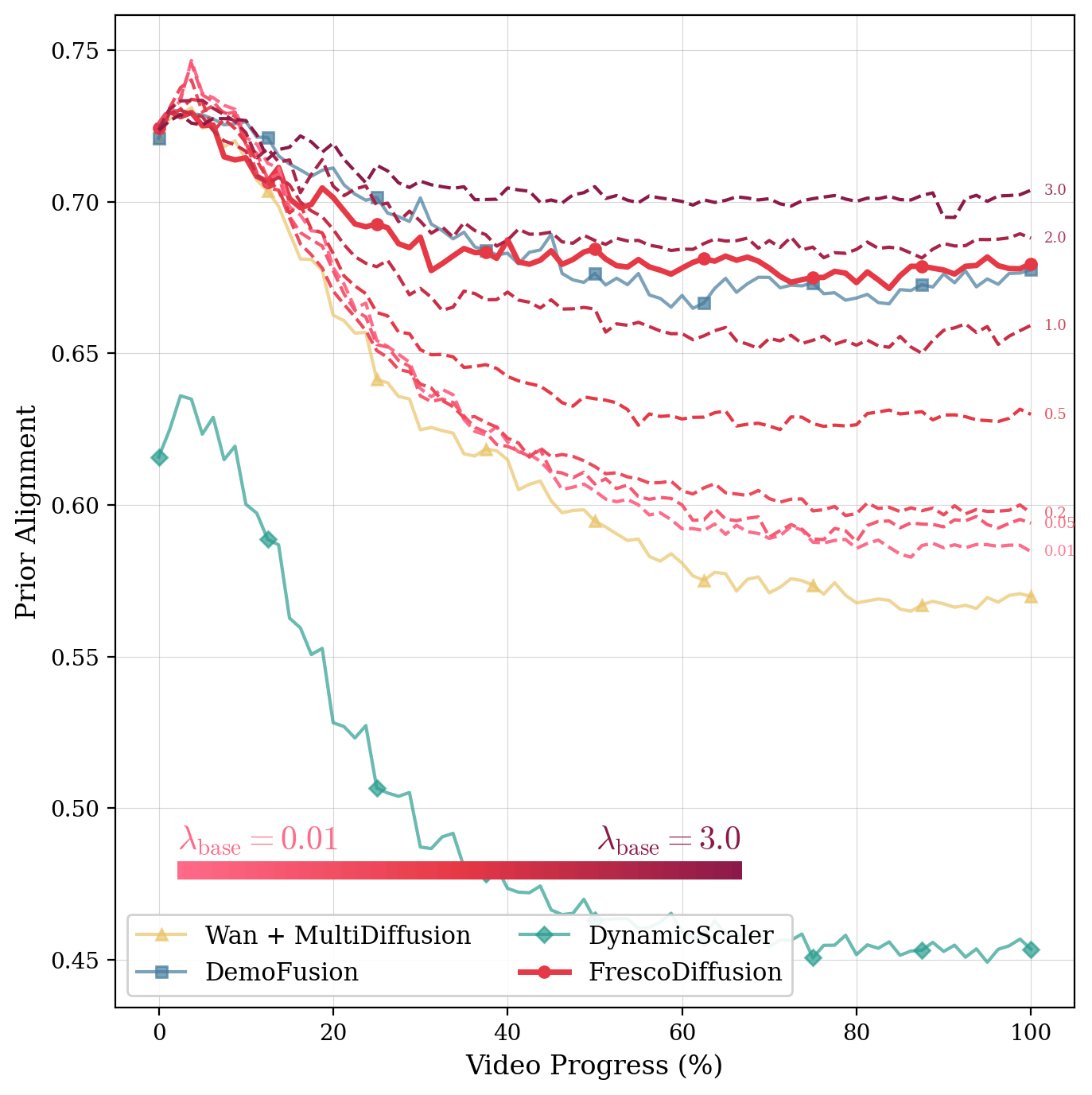}
        \caption{Prior similarity evolution over time.}
        \label{fig:quant_metrics_b}
    \end{subfigure}
    \caption{Quantitative evaluation of the trade-off between creativity and prior similarity controlled by $\lambda$. (a) Temporal consistency versus sharpness, illustrating the Pareto frontier between preserving temporal coherence and maintaining high image sharpness. (b) Evolution of prior similarity over time, showing how increasing prior strength and temporal gating progressively aligns the generated outputs with the prior.
    }
    \label{fig:quant_metrics}
\end{figure}

Creativity and prior similarity are essentially contrary objectives. One cannot improve one without hurting the other. This inherent tradeoff creates a Pareto frontier composed of the set of optimal compromises between the two objectives. 
To navigate this frontier using $\lambda_G(t,\tau)$ (\cref{eq:global-lambda}), we linearly modify the prior strength ($\lambda_\text{base}\in[0.01,3]$), and the temporal gating ($\tau\in[0, 1]$).
To represent the creativity objective we use the sharpness metric as a proxy, and for the prior similarity we use both temporal consistency and our prior similarity metric presented in \cref{sec:dataset}.
The results in \cref{fig:quant_metrics} show two expected behaviors. (i) When $\lambda_\text{base}$ and the temporal gating, $\tau$, increase, the outputs equal those of the prior. (ii) On the contrary, when those parameters decrease, we reach the same performance as MultiDiffusion (full creativity, no prior). The curve formed between these two opposites creates the Pareto frontier which allows a trade-off between the two objectives. Thus, FrescoDiffusion allows full control over this crucial trade-off.

\clearpage
\section{Limitations and Conclusion}  

\paragraph{\quad Limitations.} FrescoDiffusion relies on a meaningful low-resolution prior; an inaccurate thumbnail can propagate erroneous structure or motion in proportion to $\lambda$, although the decreasing schedule limits this effect. Content demanding globally coupled geometric motion (sustained camera motion or live sports) also falls outside its current scope. Our experiments use only Wan2.2; while \cref{app:derivation-other} gives the $\epsilon$-prediction extension, empirical cross-backbone validation remains future work. Finally, tiled denoising remains computationally expensive despite reduced-step sampling and low-precision optimizations.

\paragraph{\quad Conclusion.} FrescoDiffusion is a simple, training-free solution that uses existing video diffusion models to animate 4K, multi-scene images. It combines tiled denoising with a latent prior derived from a thumbnail animation to preserve global coherence while introducing local detail at large scales. In user studies, our variants are preferred over MultiDiffusion and DynamicScaler on both datasets, preferred over DemoFusion on FrescoArchive, and competitive with it on VBench-I2V, while using less wall-clock time than the stronger baselines. Our approach also exposes an interpretable balance between creativity and fidelity for large-scale image animation.

\section*{Acknowledgements}

\begin{sloppypar}
This work has been partially supported by the VISA DEEP chair (grant $\text{ANR-20-CHIA-0022}$), the PostGenAI@Paris cluster (grant $\text{ANR-23-IACL-0007}$; France 2030), and funded by the French National Research Agency (ANR) under the Renaissance project (grant $\text{ANR-23-CE23-0023}$; France 2030).
\end{sloppypar}

Computing and storage resources were provided by GENCI at IDRIS under grant $\text{2025-AD011016538}$ on the Jean Zay H100 partition.

A CC BY 4.0 public copyright license has been applied by the authors to the present document and will be applied to all subsequent versions up to the Author Accepted Manuscript (AAM) arising from this submission, in accordance with the grant’s open access conditions.

\bibliographystyle{splncs04}
\bibliography{main}

@String(PAMI  = {IEEE Trans. Pattern Anal. Mach. Intell.})

@String(CVPR  = {IEEE Conf. Comput. Vis. Pattern Recog.})

@String(ICCV  = {Int. Conf. Comput. Vis.})

@String(NeurIPS = {Adv. Neural Inform. Process. Syst.})

@String(ICML  = {Int. Conf. Mach. Learn.})

@String(ICLR  = {Int. Conf. Learn. Represent.})

@String(TOG   = {ACM Trans. Graph.})

@String(TIP   = {IEEE Trans. Image Process.})

@String(PR    = {Pattern Recognition})

@String(PAMI  = {IEEE TPAMI})

@String(CVPR  = {CVPR})

@String(ICCV  = {ICCV})

@String(NeurIPS = {NeurIPS})

@String(ICML  = {ICML})

@String(ICLR  = {ICLR})

@String(TOG   = {ACM TOG})

@String(TIP   = {IEEE TIP})

@String(PR    = {PR})

@Article{avrahami2023blended,
  author = {Avrahami, Omri and Fried, Ohad and Lischinski, Dani},
  title = {Blended latent diffusion},
  journal = TOG,
  year = {2023},
  volume = {42},
  number = {4},
  pages = {1--11}
}

@Article{bai2025qwen3,
  author = {Bai, Shuai and Cai, Yuxuan and Chen, Ruizhe and Chen, Keqin and Chen, Xionghui and Cheng, Zesen and Deng, Lianghao and Ding, Wei and Gao, Chang and Ge, Chunjiang and others},
  title = {Qwen3-VL Technical Report},
  journal = {arXiv preprint arXiv:2511.21631},
  year = {2025}
}

@InProceedings{bar2023multidiffusion,
  author = {Bar-Tal, Omer and Yariv, Lior and Lipman, Yaron and Dekel, Tali},
  title = {{MultiDiffusion}: Fusing Diffusion Paths for Controlled Image Generation},
  booktitle = ICML,
  year = {2023},
  pages = {1737--1752},
  volume = {202}
}

@InProceedings{bar2024lumiere,
  author = {Bar-Tal, Omer and Chefer, Hila and Tov, Omer and Herrmann, Charles and Paiss, Roni and Zada, Shiran and Ephrat, Ariel and Hur, Junhwa and Liu, Guanghui and Raj, Amit and others},
  title = {Lumiere: A space-time diffusion model for video generation},
  booktitle = {SIGGRAPH Asia 2024 Conference Papers},
  year = {2024},
  pages = {1--11}
}

@Article{batifol2025flux,
  author = {Batifol, Stephen and Blattmann, Andreas and Boesel, Frederic and Consul, Saksham and Diagne, Cyril and Dockhorn, Tim and English, Jack and English, Zion and Esser, Patrick and Kulal, Sumith and others},
  title = {FLUX. 1 Kontext: Flow Matching for In-Context Image Generation and Editing in Latent Space},
  journal = {arXiv e-prints},
  year = {2025},
  pages = {arXiv--2506}
}

@Article{blattmann2023stable,
  author = {Blattmann, Andreas and Dockhorn, Tim and Kulal, Sumith and Mendelevitch, Daniel and Kilian, Maciej and Lorenz, Dominik and Levi, Yam and English, Zion and Voleti, Vikram and Letts, Adam and others},
  title = {Stable video diffusion: Scaling latent video diffusion models to large datasets},
  journal = {arXiv preprint arXiv:2311.15127},
  year = {2023}
}

@InProceedings{bolya2025perception,
  author = {Bolya, Daniel and Huang, Po-Yao and Sun, Peize and Cho, Jang Hyun and Madotto, Andrea and Wei, Chen and Ma, Tengyu and Zhi, Jiale and Rajasegaran, Jathushan and Rasheed, Hanoona Abdul and others},
  title = {Perception Encoder: The best visual embeddings are not at the output of the network},
  booktitle = NeurIPS,
  year = {2025}
}

@Article{carion2025sam,
  author = {Carion, Nicolas and Gustafson, Laura and Hu, Yuan-Ting and Debnath, Shoubhik and Hu, Ronghang and Suris, Didac and Ryali, Chaitanya and Alwala, Kalyan Vasudev and Khedr, Haitham and Huang, Andrew and others},
  title = {Sam 3: Segment anything with concepts},
  journal = {arXiv preprint arXiv:2511.16719},
  year = {2025}
}

@InProceedings{chan2021basicvsr,
  author = {Chan, Kelvin CK and Wang, Xintao and Yu, Ke and Dong, Chao and Loy, Chen Change},
  title = {Basicvsr: The search for essential components in video super-resolution and beyond},
  booktitle = CVPR,
  year = {2021},
  pages = {4947--4956}
}

@InProceedings{chan2022basicvsr++,
  author = {Chan, Kelvin CK and Zhou, Shangchen and Xu, Xiangyu and Loy, Chen Change},
  title = {Basicvsr++: Improving video super-resolution with enhanced propagation and alignment},
  booktitle = CVPR,
  year = {2022},
  pages = {5972--5981}
}

@InProceedings{chan2022investigating,
  author = {Chan, Kelvin CK and Zhou, Shangchen and Xu, Xiangyu and Loy, Chen Change},
  title = {Investigating tradeoffs in real-world video super-resolution},
  booktitle = CVPR,
  year = {2022},
  pages = {5962--5971}
}

@Article{chen2023videocrafter1,
  author = {Chen, Haoxin and Xia, Menghan and He, Yingqing and Zhang, Yong and Cun, Xiaodong and Yang, Shaoshu and Xing, Jinbo and Liu, Yaofang and Chen, Qifeng and Wang, Xintao and others},
  title = {Videocrafter1: Open diffusion models for high-quality video generation},
  journal = {arXiv preprint arXiv:2310.19512},
  year = {2023}
}

@InProceedings{du2024demofusion,
  author = {Du, Ruoyi and Chang, Dongliang and Hospedales, Timothy and Song, Yi-Zhe and Ma, Zhanyu},
  title = {Demofusion: Democratising high-resolution image generation with no \$\$\$},
  booktitle = CVPR,
  year = {2024},
  pages = {6159--6168}
}

@InProceedings{frolov2025spotdiffusion,
  author = {Frolov, Stanislav and Moser, Brian B and Dengel, Andreas},
  title = {Spotdiffusion: A fast approach for seamless panorama generation over time},
  booktitle = {2025 IEEE/CVF Winter Conference on Applications of Computer Vision (WACV)},
  year = {2025},
  pages = {2073--2081}
}

@Misc{genmo2024mochi,
  author = {Genmo Team},
  title = {Mochi 1},
  howpublished = {\url{https://github.com/genmoai/models}},
  journal = {GitHub repository},
  year = {2024}
}

@Article{hacohen2024ltx,
  author = {HaCohen, Yoav and Chiprut, Nisan and Brazowski, Benny and Shalem, Daniel and Moshe, Dudu and Richardson, Eitan and Levin, Eran and Shiran, Guy and Zabari, Nir and Gordon, Ori and others},
  title = {Ltx-video: Realtime video latent diffusion},
  journal = {arXiv preprint arXiv:2501.00103},
  year = {2024}
}

@InProceedings{he2023scalecrafter,
  author = {He, Yingqing and Yang, Shaoshu and Chen, Haoxin and Cun, Xiaodong and Xia, Menghan and Zhang, Yong and Wang, Xintao and He, Ran and Chen, Qifeng and Shan, Ying},
  title = {Scalecrafter: Tuning-free higher-resolution visual generation with diffusion models},
  booktitle = ICLR,
  year = {2023}
}

@Article{he2024venhancer,
  author = {He, Jingwen and Xue, Tianfan and Liu, Dongyang and Lin, Xinqi and Gao, Peng and Lin, Dahua and Qiao, Yu and Ouyang, Wanli and Liu, Ziwei},
  title = {VEnhancer: Generative Space-Time Enhancement for Video Generation},
  journal = {arXiv preprint arXiv:2407.07667},
  year = {2024}
}

@Article{ho2022imagen,
  author = {Ho, Jonathan and Chan, William and Saharia, Chitwan and Whang, Jay and Gao, Ruiqi and Gritsenko, Alexey and Kingma, Diederik P and Poole, Ben and Norouzi, Mohammad and Fleet, David J and others},
  title = {Imagen video: High definition video generation with diffusion models},
  journal = {arXiv preprint arXiv:2210.02303},
  year = {2022}
}

@InProceedings{huang2023vbench,
  author = {Huang, Ziqi and He, Yinan and Yu, Jiashuo and Zhang, Fan and Si, Chenyang and Jiang, Yuming and Zhang, Yuanhan and Wu, Tianxing and Jin, Qingyang and Chanpaisit, Nattapol and Wang, Yaohui and Chen, Xinyuan and Wang, Limin and Lin, Dahua and Qiao, Yu and Liu, Ziwei},
  title = {{VBench}: Comprehensive Benchmark Suite for Video Generative Models},
  booktitle = CVPR,
  year = {2024}
}

@Article{huang2025vbench++,
  author = {Huang, Ziqi and Zhang, Fan and Xu, Xiaojie and He, Yinan and Yu, Jiashuo and Dong, Ziyue and Ma, Qianli and Chanpaisit, Nattapol and Si, Chenyang and Jiang, Yuming and Wang, Yaohui and Chen, Xinyuan and Chen, Ying-Cong and Wang, Limin and Lin, Dahua and Qiao, Yu and Liu, Ziwei},
  title = {{VBench++}: Comprehensive and Versatile Benchmark Suite for Video Generative Models},
  journal = PAMI,
  year = {2025},
  doi = {10.1109/TPAMI.2025.3633890}
}

@Misc{idealods2019imagededup,
  author = {Tanuj Jain and Christopher Lennan and Zubin John and Dat Tran},
  title = {Imagededup},
  howpublished = {\url{https://github.com/idealo/imagededup}},
  year = {2019}
}

@Article{jimenez2023mixture,
  author = {Jim{\'e}nez, {\'A}lvaro Barbero},
  title = {Mixture of diffusers for scene composition and high resolution image generation},
  journal = {arXiv preprint arXiv:2302.02412},
  year = {2023}
}

@Article{liang2024vrt,
  author = {Liang, Jingyun and Cao, Jiezhang and Fan, Yuchen and Zhang, Kai and Ranjan, Rakesh and Li, Yawei and Timofte, Radu and Van Gool, Luc},
  title = {Vrt: A video restoration transformer},
  journal = TIP,
  year = {2024},
  volume = {33},
  pages = {2171--2182}
}

@Article{liu2022video,
  author = {Liu, Hongying and Ruan, Zhubo and Zhao, Peng and Dong, Chao and Shang, Fanhua and Liu, Yuanyuan and Yang, Linlin and Timofte, Radu},
  title = {Video super-resolution based on deep learning: a comprehensive survey},
  journal = {Artificial Intelligence Review},
  year = {2022},
  volume = {55},
  number = {8},
  pages = {5981--6035}
}

@InProceedings{liu2025dynamicscaler,
  author = {Liu, Jinxiu and Lin, Shaoheng and Li, Yinxiao and Yang, Ming-Hsuan},
  title = {Dynamicscaler: Seamless and scalable video generation for panoramic scenes},
  booktitle = CVPR,
  year = {2025},
  pages = {6144--6153}
}

@Misc{oquab2023dinov2,
  author = {Oquab, Maxime and Darcet, Timothée and Moutakanni, Theo and Vo, Huy V. and Szafraniec, Marc and Khalidov, Vasil and Fernandez, Pierre and Haziza, Daniel and Massa, Francisco and El-Nouby, Alaaeldin and Howes, Russell and Huang, Po-Yao and Xu, Hu and Sharma, Vasu and Li, Shang-Wen and Galuba, Wojciech and Rabbat, Mike and Assran, Mido and Ballas, Nicolas and Synnaeve, Gabriel and Misra, Ishan and Jegou, Herve and Mairal, Julien and Labatut, Patrick and Joulin, Armand and Bojanowski, Piotr},
  title = {DINOv2: Learning Robust Visual Features without Supervision},
  journal = {arXiv:2304.07193},
  year = {2023}
}

@Article{pertuz2013analysis,
  author = {Pertuz, Said and Puig, Domenec and Garcia, Miguel Angel},
  title = {Analysis of focus measure operators for shape-from-focus},
  journal = PR,
  year = {2013},
  volume = {46},
  number = {5},
  pages = {1415--1432}
}

@InProceedings{radford2021learning,
  author = {Radford, Alec and Kim, Jong Wook and Hallacy, Chris and Ramesh, Aditya and Goh, Gabriel and Agarwal, Sandhini and Sastry, Girish and Askell, Amanda and Mishkin, Pamela and Clark, Jack and others},
  title = {Learning transferable visual models from natural language supervision},
  booktitle = ICML,
  year = {2021},
  pages = {8748--8763}
}

@InProceedings{rombach2022high,
  author = {Rombach, Robin and Blattmann, Andreas and Lorenz, Dominik and Esser, Patrick and Ommer, Bj{\"o}rn},
  title = {High-resolution image synthesis with latent diffusion models},
  booktitle = CVPR,
  year = {2022},
  pages = {10684--10695}
}

@Article{schuhmann2022laion,
  author = {Schuhmann, Christoph and Beaumont, Romain and Vencu, Richard and Gordon, Cade and Wightman, Ross and Cherti, Mehdi and Coombes, Theo and Katta, Aarush and Mullis, Clayton and Wortsman, Mitchell and others},
  title = {Laion-5b: An open large-scale dataset for training next generation image-text models},
  journal = {Advances in neural information processing systems},
  year = {2022},
  volume = {35},
  pages = {25278--25294}
}

@Article{simeoni2025dinov3,
  author = {Sim{\'e}oni, Oriane and Vo, Huy V and Seitzer, Maximilian and Baldassarre, Federico and Oquab, Maxime and Jose, Cijo and Khalidov, Vasil and Szafraniec, Marc and Yi, Seungeun and Ramamonjisoa, Micha{\"e}l and others},
  title = {Dinov3},
  journal = {arXiv preprint arXiv:2508.10104},
  year = {2025}
}

@InProceedings{Singer2022MakeAVideo,
  author = {Singer, Uriel and Polyak, Adam and Hayes, Thomas and Yin, Xi and An, Jie and Zhang, Songyang and Hu, Qiyuan and Yang, Harry and Ashual, Oron and Gafni, Oran and Parikh, Devi and Gupta, Sonal and Taigman, Yaniv},
  title = {Make-A-Video: Text-to-Video Generation without Text-Video Data},
  booktitle = ICLR,
  year = {2023}
}

@Article{wan2025wan,
  author = {Wan, Team and Wang, Ang and Ai, Baole and Wen, Bin and Mao, Chaojie and Xie, Chen-Wei and Chen, Di and Yu, Feiwu and Zhao, Haiming and Yang, Jianxiao and others},
  title = {Wan: Open and advanced large-scale video generative models},
  journal = {arXiv preprint arXiv:2503.20314},
  year = {2025}
}

@Article{wang2025internvl3,
  author = {Wang, Weiyun and Gao, Zhangwei and Gu, Lixin and Pu, Hengjun and Cui, Long and Wei, Xingguang and Liu, Zhaoyang and Jing, Linglin and Ye, Shenglong and Shao, Jie and others},
  title = {Internvl3. 5: Advancing open-source multimodal models in versatility, reasoning, and efficiency},
  journal = {arXiv preprint arXiv:2508.18265},
  year = {2025}
}

@InProceedings{xie2025star,
  author = {Xie, Rui and Liu, Yinhong and Zhou, Penghao and Zhao, Chen and Zhou, Jun and Zhang, Kai and Zhang, Zhenyu and Yang, Jian and Yang, Zhenheng and Tai, Ying},
  title = {Star: Spatial-temporal augmentation with text-to-video models for real-world video super-resolution},
  booktitle = ICCV,
  year = {2025},
  pages = {17108--17118}
}

@InProceedings{zhang2023diffcollage,
  author = {Zhang, Qinsheng and Song, Jiaming and Huang, Xun and Chen, Yongxin and Liu, Ming-Yu},
  title = {Diffcollage: Parallel generation of large content with diffusion models},
  booktitle = CVPR,
  year = {2023},
  pages = {10188--10198}
}

@Misc{zhang2025turbodiffusionacceleratingvideodiffusion,
  author = {Jintao Zhang and Kaiwen Zheng and Kai Jiang and Haoxu Wang and Ion Stoica and Joseph E. Gonzalez and Jianfei Chen and Jun Zhu},
  title = {TurboDiffusion: Accelerating Video Diffusion Models by 100-200 Times},
  year = {2025},
  eprint = {2512.16093},
  archiveprefix = {arXiv},
  primaryclass = {cs.CV}
}

@Article{zheng2025vbench2,
  author = {Zheng, Dian and Huang, Ziqi and Liu, Hongbo and Zou, Kai and He, Yinan and Zhang, Fan and Zhang, Yuanhan and He, Jingwen and Zheng, Wei-Shi and Qiao, Yu and Liu, Ziwei},
  title = {{VBench-2.0}: Advancing Video Generation Benchmark Suite for Intrinsic Faithfulness},
  journal = {arXiv preprint arXiv:2503.21755},
  year = {2025}
}

@InProceedings{zhou2024upscale,
  author = {Zhou, Shangchen and Yang, Peiqing and Wang, Jianyi and Luo, Yihang and Loy, Chen Change},
  title = {Upscale-a-video: Temporal-consistent diffusion model for real-world video super-resolution},
  booktitle = CVPR,
  year = {2024},
  pages = {2535--2545}
}

@article{flashvsr,
  title={{FlashVSR}: Towards Real-Time Diffusion-Based Streaming Video Super-Resolution},
  author={Zhuang, Junhao and Guo, Shi and Cai, Xin and Li, Xiaohui and Liu, Yihao and Yuan, Chun and Xue, Tianfan},
  journal={arXiv preprint arXiv:2510.12747},
  year={2025}
}

@inproceedings{teed2020raft,
  title={{RAFT}: Recurrent all-pairs field transforms for optical flow},
  author={Teed, Zachary and Deng, Jia},
  booktitle={European conference on computer vision},
  pages={402--419},
  year={2020},
  organization={Springer}
}

@Article{li2026pixelwizard,
  author = {Li, Wenxue and Ren, Jingjing and Zhang, Peng and Ye, Tian and Zhou, Daiguo and Luan, Jian and Zhu, Lei},
  title = {{PixelWizard}: Towards Efficient High-Fidelity Video Generation at Ultra-Large Spatial Resolution},
  journal = {arXiv preprint arXiv:2605.25801},
  year = {2026}
}

\clearpage

\appendix
\makeatletter
\renewcommand{\theHsection}{appendix.\thesection}
\renewcommand{\theHsubsection}{appendix.\thesubsection}
\renewcommand{\theHsubsubsection}{appendix.\thesubsubsection}
\makeatother
\begin{center}{\Large\bfseries\boldmath
 \pretolerance=10000
 FrescoDiffusion: 4K Image-to-Video with Prior-Regularized Tiled Diffusion - Supplementary Material\par}
\end{center}

\section{FrescoDiffusion Additional Details}

\subsection{Implementation Details}
\label{app:fresco-implem-details}

All experiments use the Wan2.2-I2V 14B backbone with the same accelerated 6-step TurboDiffusion setting as in the main paper. We generate $81$ frames at $16$ fps with guidance scale $1.0$, i.e., without effective classifier-free guidance. For an input image of size $(H,W)$, we first generate the low-resolution prior at fixed target area $480\times832 = 399{,}360$ pixels while preserving aspect ratio:
\[
\hat H=\mathrm{round}\!\left(\sqrt{399360\,H/W}\right),\qquad
\hat W=\mathrm{round}\!\left(\sqrt{399360\,W/H}\right),
\]
and then snap both dimensions down to the nearest multiple of $16$. The full-resolution pass uses $H^{4K}=\max(16,\lfloor H/16\rfloor\cdot16)$ and $W^{4K}=\max(16,\lfloor W/16\rfloor\cdot16)$, after an optional isotropic downscaling when the input exceeds ultra-HD resolution. This multiple-of-$16$ constraint comes from the latent video backbone: the Wan VAE reduces spatial resolution by a factor $8$, and the downstream latent grid is processed in spatial patches of size $2$, yielding an effective validity constraint of $8\times2=16$. With spatial and temporal compression factors $8$ and $4$, the latent tensor therefore has shape $16\times21\times(H^{4K}/8)\times(W^{4K}/8)$ for the default $81$-frame setting. The low-resolution latent prior is resized to the large latent canvas with endpoint-aligned trilinear interpolation in latent space, VAE tiling is enabled during inference, and the decoded output is resized back to the original image size only if snapping changed the resolution.

\paragraph{Tiled denoising and regularization.}
The high-resolution pass uses MultiDiffusion windows of size $480\times832$ pixels with $30\%$ overlap, giving nominal pixel strides $336\times582$. After conversion to latent coordinates and rounding to the valid latent grid, this becomes $60\times104$ latent windows with strides $42\times72$; extra final windows are added whenever needed so that the right and bottom boundaries are exactly covered. Tile fusion uses linear ramps with minimum border weight $0.1$. Standard MultiDiffusion uses $\sum_i w_i y_i / \sum_i w_i$, whereas the prior-regularized implementation additionally accumulates $\sum_i w_i y_i$ and $\sum_i w_i$ for the one-shot closed-form update in model-output space. In all reported FrescoDiffusion runs we use a cosine prior schedule with $\lambda_{\mathrm{base}}=1.5$ and cutoff $\tau_{\mathrm{end}}=0.1$, where $\tau=i/(N-1)$ and $N=6$ (number of steps); hence the prior is active only at the first denoising step. In R-FrescoDiffusion, the active regions cutoff is $\tau_{\mathrm{act}}=0.1$ and the inactive regions cutoff is $\tau_{\mathrm{bg}}=0.35$. Since the six normalized step positions are $\{0,0.2,0.4,0.6,0.8,1.0\}$, the foreground prior is active only at $i=0$, while the background prior is active at $i=0$ and $i=1$. The active or inactive parts of the video are determined based on masks computed according to \cref{app:sam3}.

\subsection{Sampling Procedure}
\label{app:algo}

At each step, we run the transformer on each crop of the canvas latent \(x_0^\text{4K}\) to obtain a per-tile prediction \(y_i\). Then, we accumulate two canvas-shaped tensors: \(\sum_i w_i\odot y_i\) and \(\sum_i w_i\). We then add the prior term and divide as in \eqref{eq:closedform} to produce a single fused prediction on the full canvas. Finally, we invoke the scheduler exactly once to obtain \(x_{t+\Delta t}\). This preserves the original sampler while adding only light overhead: one extra reduction per pixel, a single pointwise rational fuse, and no additional network passes beyond those already required by tiled denoising. The full step is summarized in \cref{alg:f2v}.

\begin{algorithm}[th!]
\caption{FrescoDiffusion: one sampler step at time \(t\)}
\label{alg:f2v}
\DontPrintSemicolon
\KwIn{
canvas latent \(x^\text{4K}_t\in\mathbb{R}^{C\times T\times H^\text{4K}\times W^\text{4K}}\);
tile positions \(\{p_i\}_{i=1}^n\);
weight maps \(\{w_i\}_{i=1}^n\);
upscaled prior \(x_{\text{prior}}\in\mathbb{R}^{C\times T\times H^\text{4K}\times W^\text{4K}}\);
noise level \(\sigma_t\);
prior-strength schedule \(\lambda\);
flow-matching step size \(\Delta t\);
flow-matching model $f_\theta$
}
\KwOut{updated canvas latent \(x_{t+\Delta t}^\text{4K}\)}
\nl Initialize \(\textit{num}\leftarrow 0\in\mathbb{R}^{C\times T\times H^\text{4K}\times W^\text{4K}}\),
\(\textit{den}\leftarrow 0\in\mathbb{R}^{1\times 1\times H^\text{4K}\times W^\text{4K}}\)\;
\nl \For{\(i=1,\dots,n\)}{
    \nl \(\tilde{x}_t \leftarrow \mathscr{C}_{p_i}(x_t^\text{4K})\)\tcp*[r]{Crop canvas at $p_i$}
    \nl \(\tilde{y}_i \leftarrow f_\theta(\tilde{x}_t, t, c)\)\tcp*[r]{Estimates crop flow}
    \nl \(y_i \leftarrow \mathscr{P}_{p_i}(\tilde{y}_i)\)\tcp*[r]{Zero-pads the output}
    \nl \(\textit{num}\leftarrow \textit{num} + w_i \odot y_i\) \tcp*[r]{Updates the numerator}
    \nl \(\textit{den}\leftarrow \textit{den} + w_i\) \tcp*[r]{Updates denominator} 
}
\nl \tcp*[r]{Add prior regularization}
\[
y \leftarrow
\frac{\textit{num} + \lambda\,\sigma_t\,(x_t^\text{4K} - x_{\text{prior}})}
{\textit{den} + \lambda\,\sigma_t^2}
\]
\nl \(x_{t+\Delta t}^\text{4K}\leftarrow \textsc{SchedulerStep}(x_t^\text{4K},\,y,\,t)\) \tcp*[r]{Update noisy states}
\end{algorithm}

\section{Fresco Evaluation Dataset Construction}
\label{app:dataset}

To facilitate the creation of UHD-I2V, we noticed that no dataset fits our requirements for UHD-I2V. Unlike VBench high-definition~\cite{huang2025vbench++} set or panoramas~\cite{liu2025dynamicscaler} that focus on one object or less, we search for images to animate with multiple intricate scenes. Here, we propose a new set to generate and evaluate UHD-I2V techniques at a fresco-like scale. We name our dataset FrescoArchive. We start with the LAION-2B Aesthetic Subset~\cite{schuhmann2022laion}. The first step is then filtering the images based on several criteria: having more than a million pixels, an aesthetic score of at least 5.8, and a watermark and unsafe scores of less than 0.8 and 0.5, respectively. This initial filtering process yielded a total of two million images. Next, we performed a semantic filtering process. To do this, for each image, we compute the average cosine similarity between the target instance and the prompts
\begin{quote}
\ttfamily
\raggedright"A large detailed fresco" \\
 "A magnificent fresco with many different scenes" \\
"A narrative composition" \\
"A fresco with lots of details" \\
"A large polyptych and composite image fresco" \\
"A large metapicture with several compositions" \\
"A fresco tableau" \\
"A painting fresco"
\end{quote}
using the PerceptionEncoder~\cite{bolya2025perception} G14-448 variant similarity model, and we finish by selecting the top 50,000 images. After, we used Intern-VL-3.5~\cite{wang2025internvl3} as a classifier to detect frescoes. To do so, we used the following prompt:
\begin{quote}
\ttfamily
\raggedright
You are a visual classifier. Decide whether the image is a fresco-like composition.

Definition (for this task):
A qualifying image resembles a large, detailed, integrated scene (like historical frescoes). Modern photos or digital works count if they share these traits.

Answer yes only if all are true:

\begin{itemize}
    \item The image shows high apparent resolution / detail density (many fine, precise elements).
    \item There are multiple distinct sub-scenes or groups (from a few to dozens+) distributed across the same frame.
    \item These elements are blended into one coherent composition (no panel borders or obvious collage seams).
\end{itemize}

Answer no if any of the following:

\begin{itemize}
    \item Single subject or minimal detail.
    \item The fresco is not the main content of the image (e.g. the photo shows a wall, room, or museum scene where the fresco only appears as a small part, rather than the fresco itself being the full image).
    \item Simple graphics, logos, posters, or text-only images.
    \item Comics/manga with separate panels, tiled grids, or collages with hard borders.
    \item Diagrams, charts, UI/screenshots, or patterns.
\end{itemize}
Unsure: answer no.

Output format:
Respond with exactly yes or no (lowercase, no punctuation, no extra words).
\end{quote}
This filtering results in a total of 10,000 images. Finally, we deduplicated the dataset using both perceptual hashing and CNN-based deduplication techniques using the ImageDedup library~\cite{idealods2019imagededup}. Then, to generate UHD image captions, we used Qwen3-VL-32B~\cite{bai2025qwen3} with both the image and LAION caption, resulting in 6,734 UHD image-caption pairs. To prompt Qwen3-VL-32B, we used the following text:
\begin{quote}
\ttfamily
\raggedright
Using the existing caption below as context, write a long, highly detailed, precise, and fluent caption that thoroughly describes the image. Give some precise contextual information, relative positional information, subjects, objects and elements description and identification. Caution: the caption provided can be false or wrong; use the image as the only source of truth. The caption is only here to help you be more precise. Respond with the caption only (no preface, no metadata, no quotes). Existing caption:  \{caption\}
\end{quote}
Finally, we manually selected 371 pairs to get the best qualitative image-captions pairs.

\begin{table}
    \centering
    \begin{tabular}{c|cccc}\toprule
        Dataset   & Samples & Words         & Image Width     & Image Height \\\midrule
        FrescoArchive    & 371         & 355.79 $\pm$ 88.0 & 2265 $\pm$ 1257 & 1552 $\pm$ 864 \\
        VBench & 361         & 14.70 $\pm$ 2.24  & 4592 $\pm$ 1305 & 3748 $\pm$ 1214 \\
        \bottomrule
    \end{tabular}
    \caption{Dataset statistics. We compare FrescoArchive and VBench datasets' low-level statistics.}
    \label{tab:stats}
\end{table}

\begin{figure}
    \centering
    \includegraphics[width=0.45\linewidth]{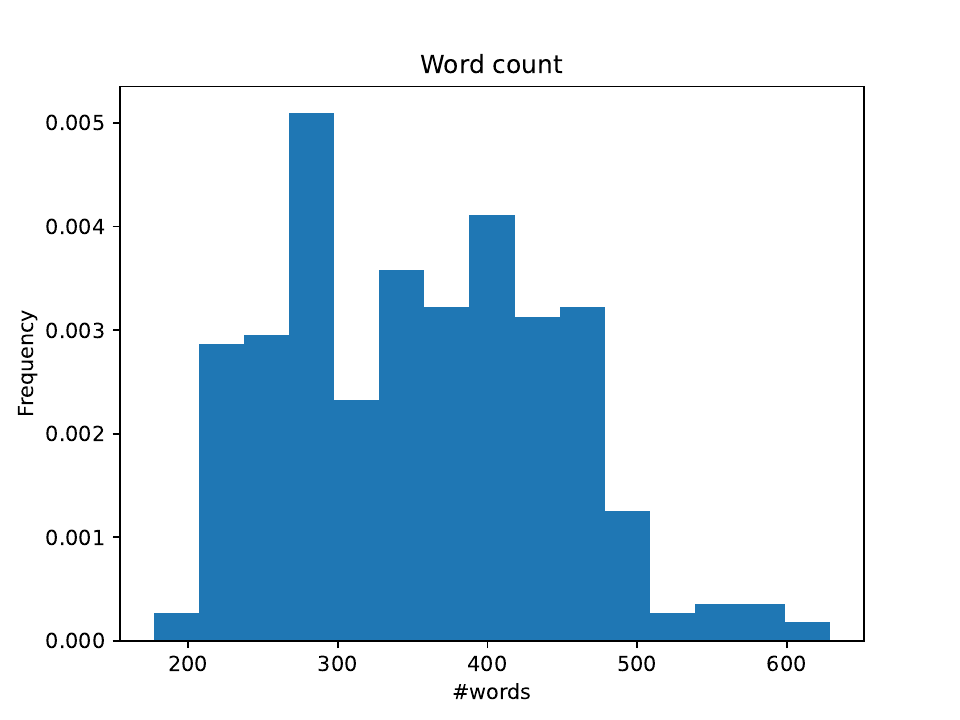}
    \includegraphics[width=0.45\linewidth]{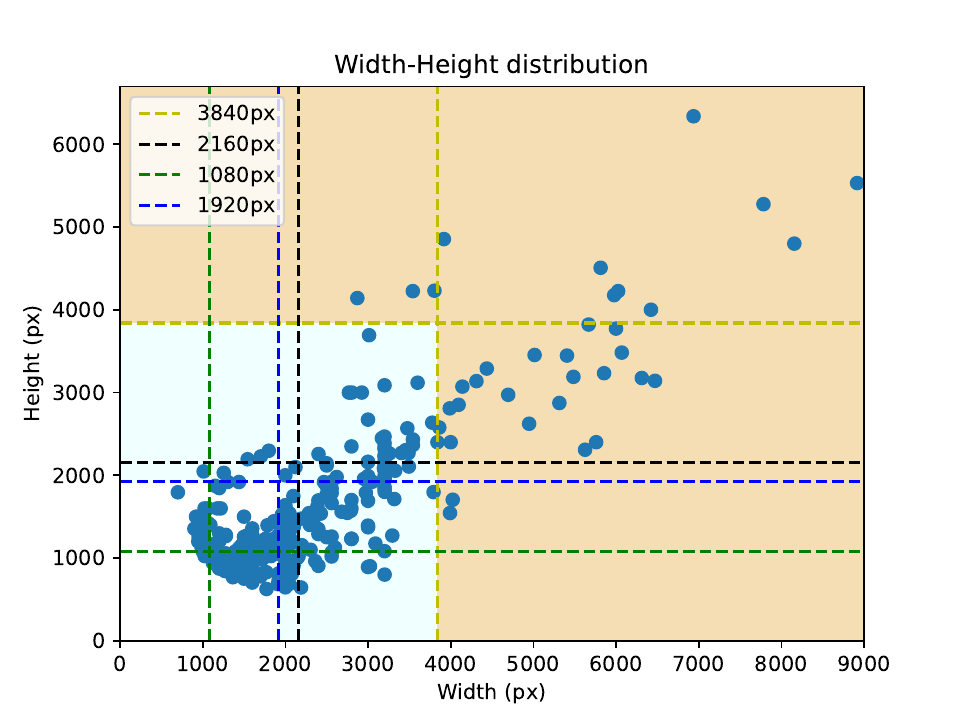}
    \caption{Analysis of FrescoArchive dataset. Left: distribution of word in caption. Right: Resolution of images distribution.}
    \label{fig:more_stats}
\end{figure}

For the statistics, we compute several metrics (text and image-wise) and quantitatively compare with VBench I2V~\cite{huang2025vbench++} to mark a reference point. First, we compute high-level statistics, seen in \cref{tab:stats}. We measure the number of samples, average number of words per prompt, and average width and height. As we can see, our set contains a similar number of images as VBench. Yet, our prompts contain an order of magnitude more than VBench, depicting more precise and detailed prompts. As for the average image shape, our set contains smaller images, but with more complex scenery. In \cref{fig:more_stats}, you can see the number of words and shape distribution of our dataset.

\begin{figure}
    \centering
\includegraphics[width=0.55\linewidth]{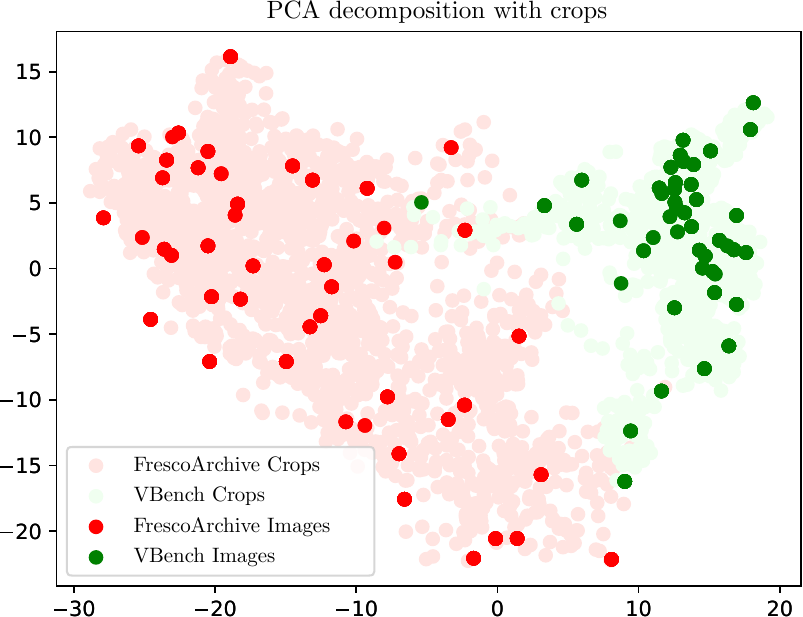}
    \caption{PCA between FrescoArchive and VBench I2V dataset. We visualize the \mbox{DINOv2~\cite{oquab2023dinov2}} feature PCA between our proposed dataset and VBench's images (red and green points, respectively). Also, in light green and red, random crops of the images. This plot shows that our dataset contains images with different statistics than standard ones, found in the VBench I2V set. 
}
    \label{fig:pca}
\end{figure}

Next, we qualitatively study FrescoArchive's complexity with reference to VBench. To do this, we first encode all images using the DINOv2~\cite{oquab2023dinov2} model to get a global view of each image. Then, we perform a PCA dimensionality reduction to visualize their distribution. Finally, using the resulting reduction, we further visualize individual crops.
As can be seen in \cref{fig:pca}, the resulting PCA shows that our dataset covers a wider span of the main axes, unlike VBench. This suggests that our dataset has more diversity in shared components using the global DINO features.

\section{Experiment Details}
\label{app:exp}

\subsection{Metrics}
\label{app:metrics}

\subsubsection{Sharpness metric.}

We assess per-frame sharpness using the Tenengrad~\cite{pertuz2013analysis} measure, a classical no-reference focus metric based on directional gradient energy. For each frame, the luminance channel is extracted and horizontal and vertical gradients are computed using a 3×3 Sobel operator. The Tenengrad score is defined as the mean squared gradient magnitude $T = \frac{1}{HW}\sum_{i,j}(G_x^2(i,j) + G_y^2(i,j))$, where $G_x$ and $G_y$ are the responses of the Sobel filters along each axis. This quantity is sensitive to high-frequency spatial structure and penalizes blurry outputs where gradient energy is suppressed. Per-frame scores are averaged over all frames of a video to yield a single video-level score. Higher values indicate sharper, more detail-preserving outputs.

\subsubsection{Temporal consistency metric.}

Each frame is first converted to grayscale and then downsampled to $128 \times 128$ pixels using area-averaging interpolation, retaining only coarse spatial structure. The temporal inconsistency score for a video of $T$ frames is defined as the mean squared error between consecutive downscaled frames: $\mathcal{C} = \frac{1}{T-1}\sum_{t=1}^{T-1}|\hat{f}_{t} - \hat{f}_{t-1}|_F^2 / \left(64^2\right)$, where $\hat{f}_t$ denotes the downscaled grayscale frame at time $t$. By operating at this coarse resolution, the metric captures global consistency while remaining agnostic to legitimate scene motion. 

\subsubsection{Prior alignment metric.}

We measure how faithfully each generated video preserves the global semantic content of the prior using a frame-level feature alignment score based on DINOv3~\cite{simeoni2025dinov3}. For each frame pair $(f_t^{\text{prior}}, f_t^{\text{gen}})$, both frames are independently forwarded through a frozen DINOv3 ViT-S/16 encoder without any resizing (\ie, at native resolution), and we extract the CLS token from each: the pooled global representation output by the transformer. Prior alignment is then defined as the mean cosine similarity between corresponding CLS tokens across all $T$ frames of a video: $\mathcal{A}
=
\frac{1}{T}\sum_{t=1}^{T}
\frac{\langle \mathbf{z}_t^{\text{prior}}, \mathbf{z}_t^{\text{gen}} \rangle}
     {\|\mathbf{z}_t^{\text{prior}}\|\,\|\mathbf{z}_t^{\text{gen}}\|}$, which equivalently measures the cosine of the angle between the two CLS token directions in feature space. We report the mean of this score across all videos in the benchmark; higher values indicate that the generated video remains semantically aligned to the prior along its temporal trajectory. A key advantage of DINOv3 over its predecessor DINOv2 is its ability to process high-resolution inputs, including 4K frames, without interpolating positional embeddings or resizing the input. This property is essential in our setting, where the generated videos are high-resolution and downscaling prior to feature extraction would discard fine-grained content that may be critical for alignment assessment.

\subsection{Standard low-resolution I2V metrics}\label{sec:full-vbench}

On \Cref{tab:quant_results} in the main paper, we presented the average results on both FrescoArchive and VBench-I2V datasets. Here, in \Cref{tab:quant_results_full}, we detail each metric: Subject Consistency, Motion Smoothness, Aesthetic, Imaging, I2V Subject Consistency, and I2V Background Consistency.

\begin{table}[ht]                  
    \centering
    \caption{Quantitative evaluation on FrescoArchive and VBench-I2V. Higher is better for all metrics except time. Best in \textbf{bold} and second best \underline{underlined}. $^*$ denotes methods adapted to the video prior setting. SC, MS, A, I, ISC, and IBC stand for Subject Consistency, Motion Smoothness, Aesthetic, Imaging, I2V Subject Consistency, and I2V Background Consistency, respectively.}                                 
    \label{tab:quant_results_full}
    \resizebox{0.9\linewidth}{!}{
    \begin{tabular}{l|cccccc|c|c}
      \toprule
      \multirow{2}{*}{Method} & \multicolumn{4}{c|}{\textbf{Quality Metrics}} & \multicolumn{2}{c|}{\textbf{I2V Metrics}} & \multirow{2}{*}{Avg} & \multirow{2}{*}{Time} \\\cmidrule{2-7}
       & SC & MS & A & I & ISC & IBC & & \\
      \midrule
      \multicolumn{9}{c}{\textit{FrescoArchive}} \\
      \midrule
      DynamicScaler (original)      & 0.945 & 0.975 & 0.693 & 0.706          & 0.893 & 0.930 & 0.857 & 18.45 \\
      DynamicScaler$^{*}$ (CVPR'25) & 0.852 & 0.971 & 0.681 & \textbf{0.754} & 0.980 & 0.989 & 0.871 & 10.25 \\
      DemoFusion$^{*}$ (CVPR'24)    & \underline{0.960} & 0.987 & 0.734 & 0.752          & \underline{0.990} & \underline{0.994} & 0.903 & 13.5  \\
      \midrule
      MultiDiffusion$^{*}$ (ICML'23)      & 0.876          & 0.974          & 0.686          & \textbf{0.754} & 0.981          & 0.989            & 0.876          & \textbf{8.15} \\
      FrescoDiffusion               & 0.958          & \underline{0.990}          & \textbf{0.738} & \underline{0.753}          & \textbf{0.991} & \textbf{0.995}    & \underline{0.904}          & \underline{8.58}          \\
      R-FrescoDiffusion             & \textbf{0.977} & \textbf{0.991} & \underline{0.736} & \underline{0.753}          & \textbf{0.991} & \underline{0.994} & \textbf{0.907} & 9.08          \\
      \midrule
      \multicolumn{9}{c}{\textit{VBench-I2V}} \\
      \midrule
      DynamicScaler (original)          & 0.949          & 0.976          & 0.707          & 0.713          & 0.893          & 0.932          & 0.862          & 18.45         \\
      DynamicScaler$^{*}$ (CVPR'25)               & 0.904          & \underline{0.985}          & 0.621          & 0.701          & 0.988          & 0.991          & 0.865          & 10.25         \\
      DemoFusion$^{*}$ (CVPR'24)                  & 0.943          & \textbf{0.989} & \textbf{0.639} & \textbf{0.720} & \textbf{0.990} & \textbf{0.993} & \textbf{0.879} & 13.5 \\
      \midrule
      MultiDiffusion$^{*}$ (ICML'23) & 0.893        & 0.984          & 0.611          & 0.698          & 0.987          & 0.989          & 0.860          & \textbf{8.15} \\
      FrescoDiffusion                       & \underline{0.933}          & \textbf{0.989} & 0.632          & \underline{0.716}          & \underline{0.989}          & \underline{0.992}          & 0.875          & \underline{8.58}          \\
      R-FrescoDiffusion                     & \textbf{0.956} & \textbf{0.989} & \underline{0.634}          & 0.712          & 0.988          & 0.991          & \underline{0.878}          & 9.08          \\
      \bottomrule
    \end{tabular}}
\end{table}

\subsection{Baseline Implementation Details}
\label{app:implem-details}

\paragraph{MultiDiffusion.} Our implementation follows the original MultiDiffusion~\cite{bar2023multidiffusion} procedure without additional heuristics. The only modifications are (i) replacing the base backbone with Wan2.2 I2V in place of the original denoiser, and (ii) applying a linear decay blending mask outside each tile to smoothly attenuate contributions near tile borders and reduce seam artifacts when merging overlapping predictions.

\paragraph{DemoFusion.} DemoFusion~\cite{du2024demofusion} introduces three techniques for high-resolution image generation. We re-implemented (i) progressive phase upsampling and (ii) skip-residual global guidance, reusing the exact hyper-parameters from the authors’ official code to enable a faithful comparison. These parameters control how resolution increases across phases (number of phases, per-phase upsampling factor, and per-scale denoising schedule) and the strength of global-structure guidance during refinement.
In contrast, we did not observe reliable gains from DemoFusion-style dilated sampling when transferring it from SDXL’s UNet denoiser to Wan2.2’s DiT-based denoiser. We hypothesize this is because dilated sampling assumes updates are roughly separable across interleaved sub-lattices an assumption that fits UNets’ local, convolutional structure but breaks for DiT models with global self-attention. Evaluating Wan2.2 on sparse lattices changes the attention context and likely shifts inputs off-distribution, leading to inconsistent offsets that merge into visible artifacts (seams/checkerboards) rather than improved coherence.

\paragraph{DynamicScaler.} We follow the authors’ official implementation of DynamicScaler~\cite{liu2025dynamicscaler} and reuse their released hyper-parameters for both the offset-shifting denoiser and the global motion-guidance module. We condition motion guidance on the same low-resolution video that we use for our FrescoDiffusion regularization prior, so both signals rely on an identical motion reference. For the sliding/rotating denoising window, we set the per-step offset (stride) to half the window size, \ie a 50\% overlap between consecutive windows, which stabilizes stitching across steps and mitigates boundary artifacts.

\subsection{Spatial Activity Map Computation}
\label{app:sam3}

Given an input frame, we compute a \emph{spatial activity map} using SAM3~\cite{carion2025sam}.
The region to animate can also be explicitly specified by the user.
For automated processing and ease of use, we employ a segmentation pipeline that identifies plausible dynamic entities and converts them into a spatial activity map.

The activity map is computed once per image, stored at latent resolution, clamped to $[0,1]$, resized to the current latent size with standard trilinear interpolation, binarized using the test $A>0$, and then kept fixed throughout sampling.

\paragraph{Prompt-based segmentation.}
Our default pipeline queries SAM3 using a fixed set of prompts corresponding to categories that commonly exhibit motion.
Specifically, we provide the following textual prompts to the model:

\begin{itemize}
\item person
\item vegetation
\item vehicles
\end{itemize}

For each prompt, SAM3 predicts candidate spatial masks corresponding to instances of the queried category. The masks are then averaged. Visualizations of such masks are provided in \cref{fig:additional_sam3_masks}.

\paragraph{Mask extraction and filtering.}
SAM3 predictions are filtered using a score threshold $\tau_s = 0.45$.
We discard masks whose relative area exceeds $0.30$ of the image area, as such regions typically correspond to overly coarse detections.
We additionally remove masks with excessive boundary contact, defined as cases where more than $80\%$ of the mask pixels lie within a $10$-pixel margin of the image border.

\paragraph{Spatial support.}
For each retained mask, we construct a spatial support region by dilating the mask using a Euclidean distance transform with radius $r = 75$ pixels.
The resulting regions are merged to produce the final spatial activity map.

\paragraph{Exploratory variants.}
We explored two extensions of this pipeline. First, we experimented with extending the spatial mask across time. Using SAM3 together with the generated prior video, the mask predicted on the input frame was propagated to cover the full temporal extent of the video, as seen in \cref{fig:additional_sam3_overlay}. Second, we evaluated a variant in which the prompts provided to SAM3 are generated automatically using a vision-language model.
In this setup, Qwen3-VL-32B~\cite{bai2025qwen3} analyzes both the input image and the generated prior video to produce textual prompts corresponding to entities that could plausibly support animation. These prompts are then used to condition SAM3, while mask extraction, filtering, and spatial support construction remain identical to the default pipeline.
In practice, neither the temporal mask propagation nor the VLM-guided prompting produced measurable improvements over the fixed-prompt approach.
As both variants introduce additional computational overhead, they were not used for the final results reported in the paper.

\clearpage

\begin{center}
\captionsetup{type=figure}

\begin{tabular}{@{}>{\centering\arraybackslash}m{0.45\textwidth} >{\centering\arraybackslash}m{0.45\textwidth}@{}}
\toprule

\includegraphics[width=0.35\textwidth]{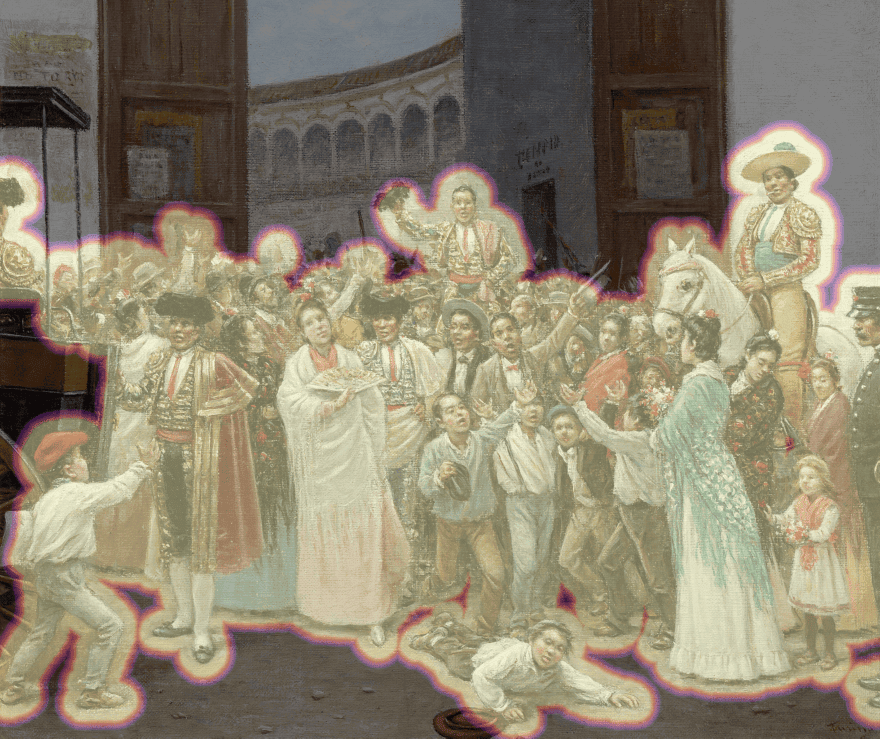} &
\includegraphics[width=0.35\textwidth]{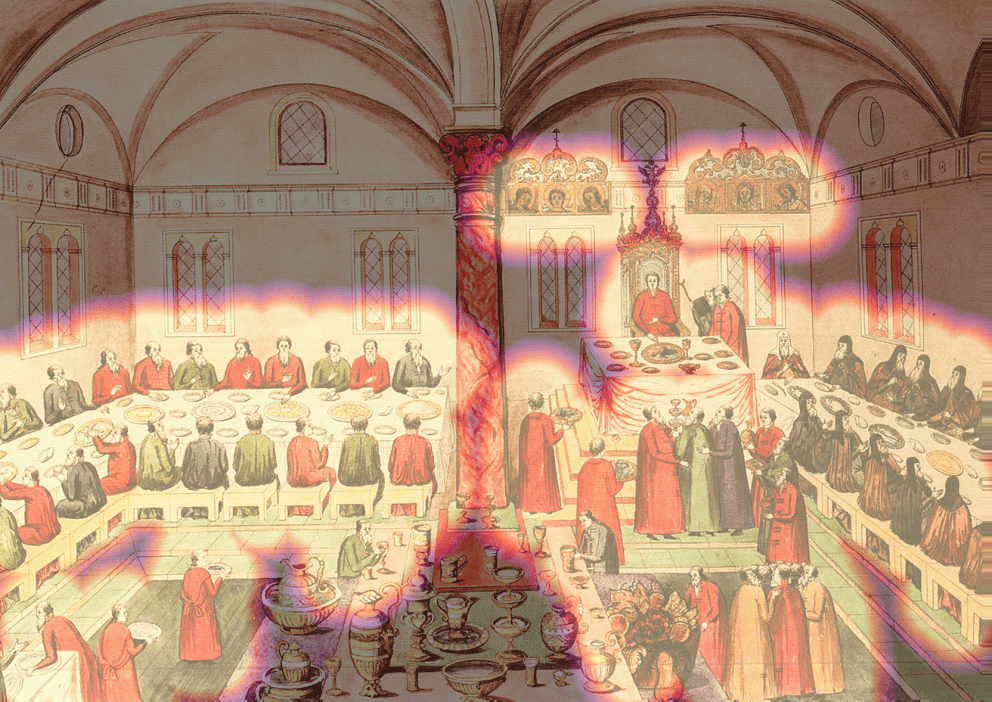} \\

\includegraphics[width=0.35\textwidth]{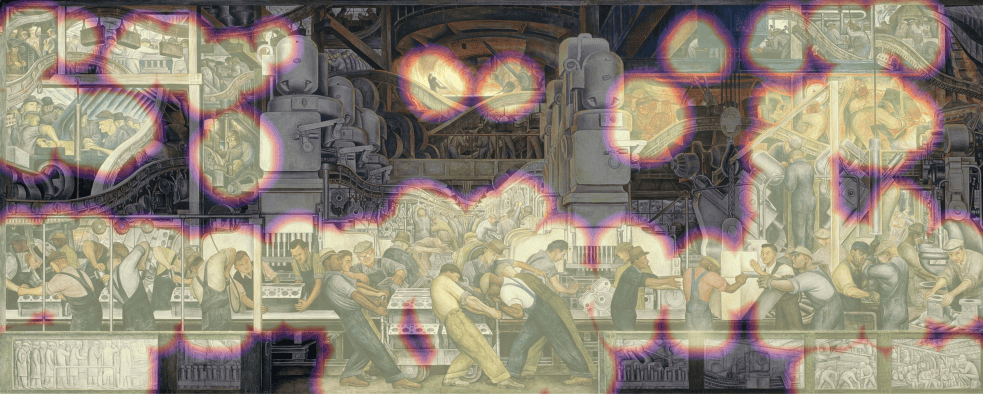} &
\includegraphics[width=0.35\textwidth]{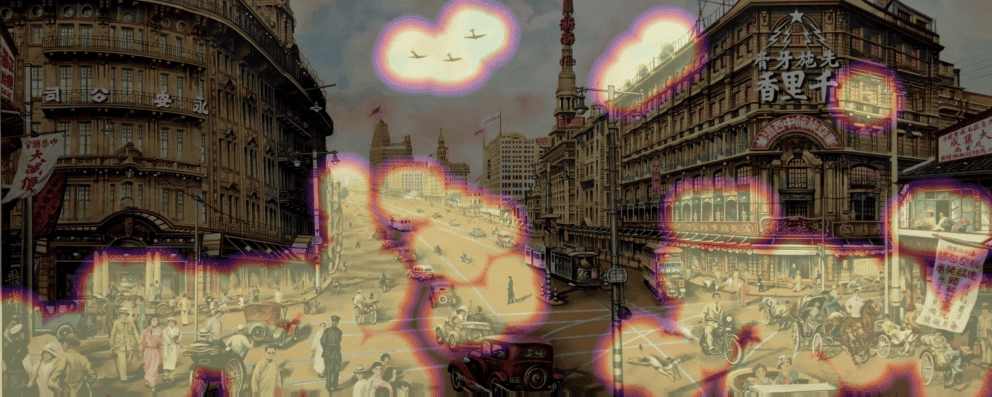} \\

\bottomrule
\end{tabular}

\caption{Additional qualitative examples of spatial activity maps obtained with SAM3. Each image shows the overlay used to identify regions likely to contain dynamic content, which are then used to guide the active/inactive prior regularization in R-FrescoDiffusion.}
\label{fig:additional_sam3_masks}
\end{center}

\begin{center}
\captionsetup{type=figure}

\begin{tabular}{@{}ccc@{}}
\toprule
\textbf{frame 1} & \textbf{frame 40} & \textbf{frame 80} \\
\midrule

\includegraphics[width=0.30\textwidth]{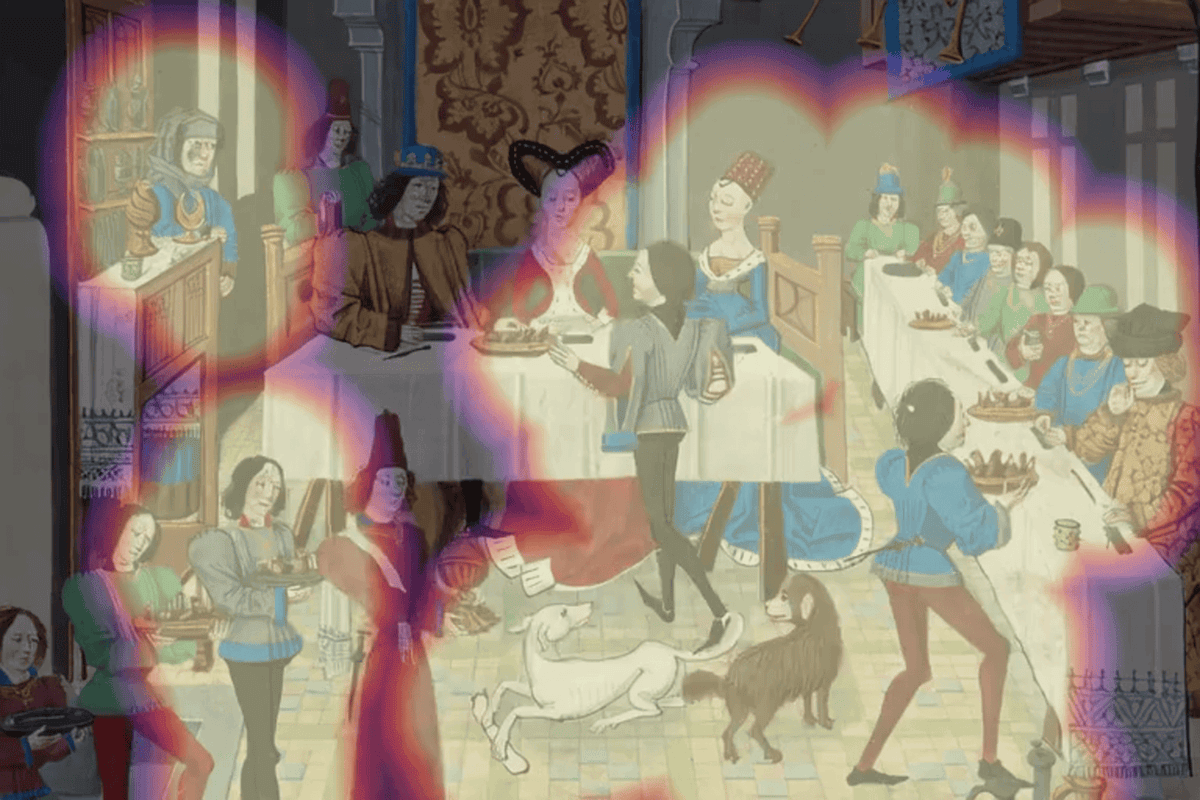} &
\includegraphics[width=0.30\textwidth]{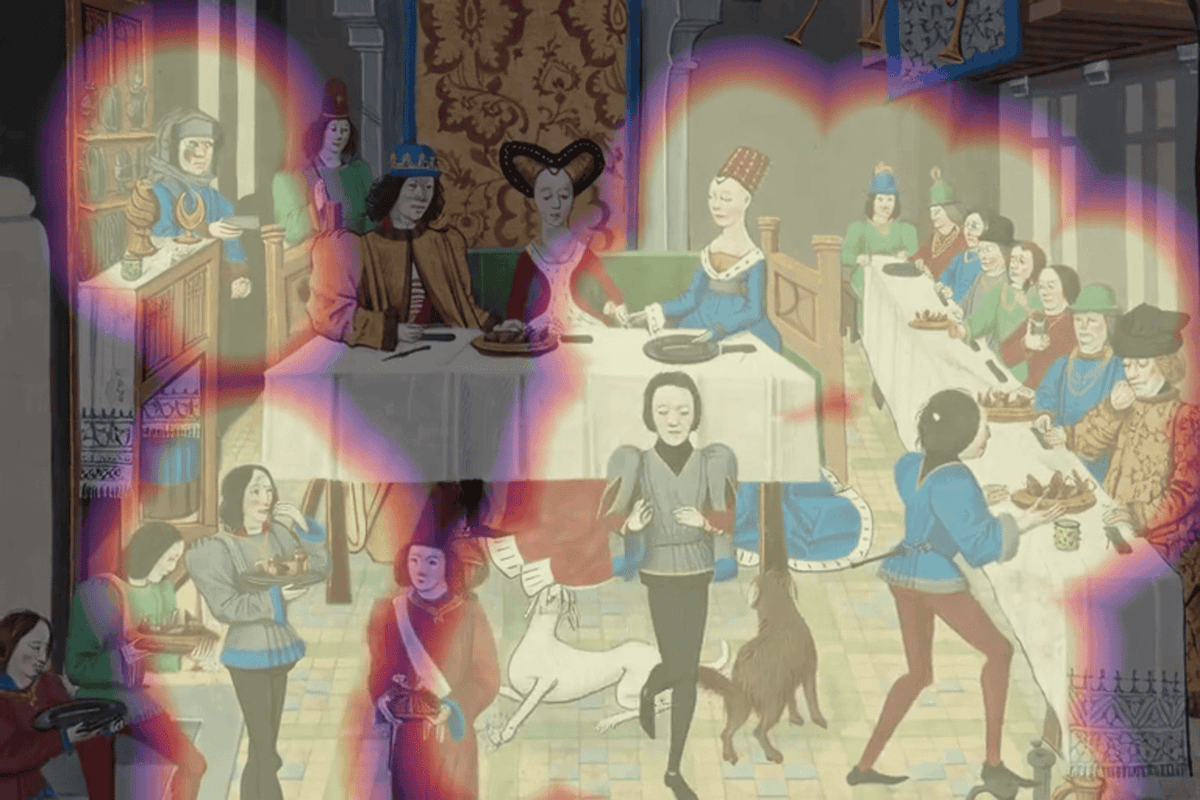} &
\includegraphics[width=0.30\textwidth]{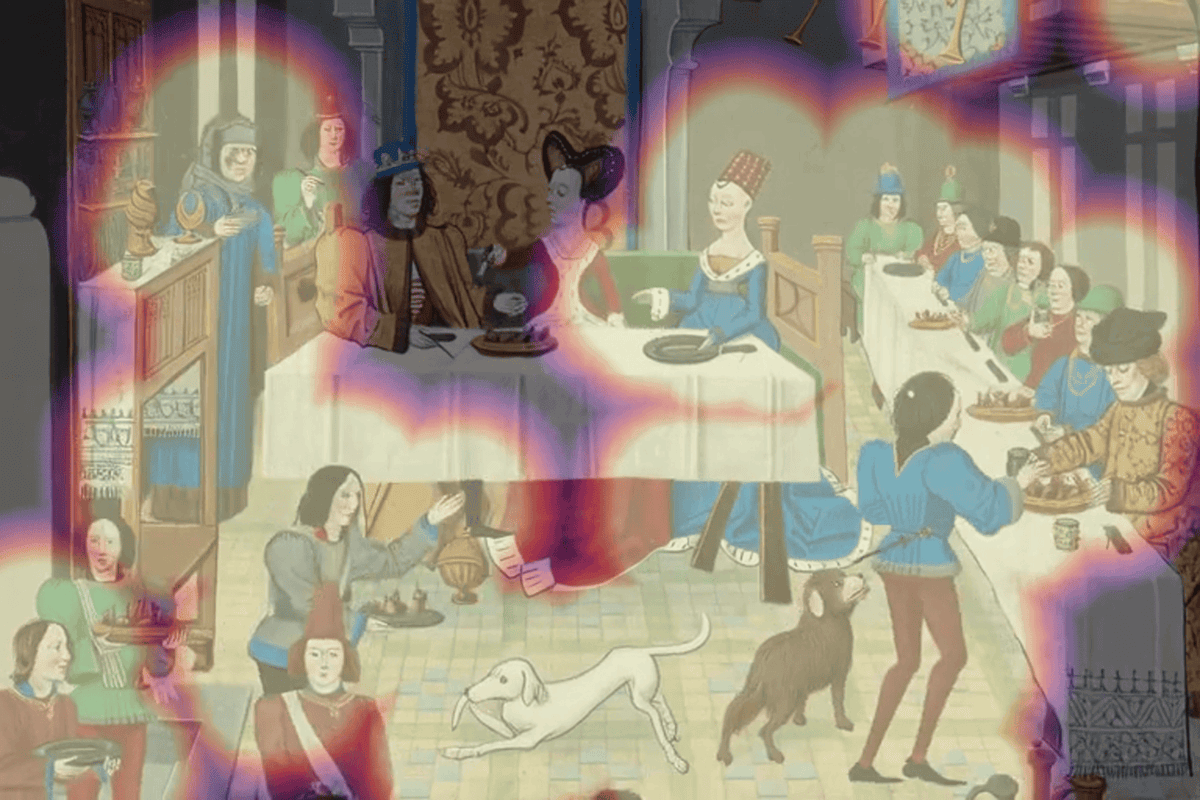} \\

\includegraphics[width=0.30\textwidth]{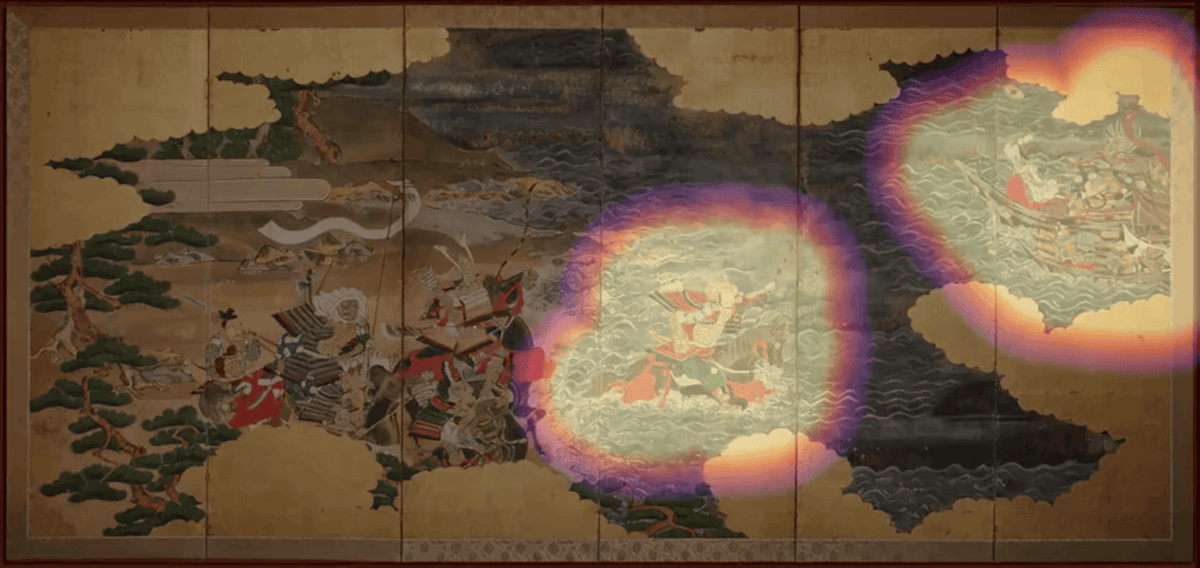} &
\includegraphics[width=0.30\textwidth]{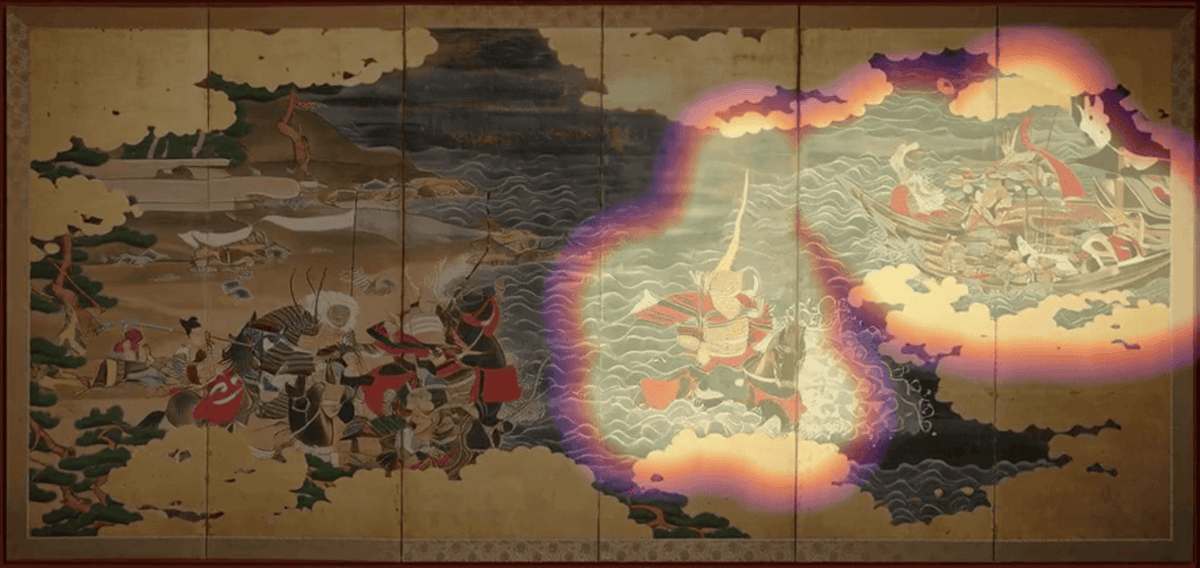} &
\includegraphics[width=0.30\textwidth]{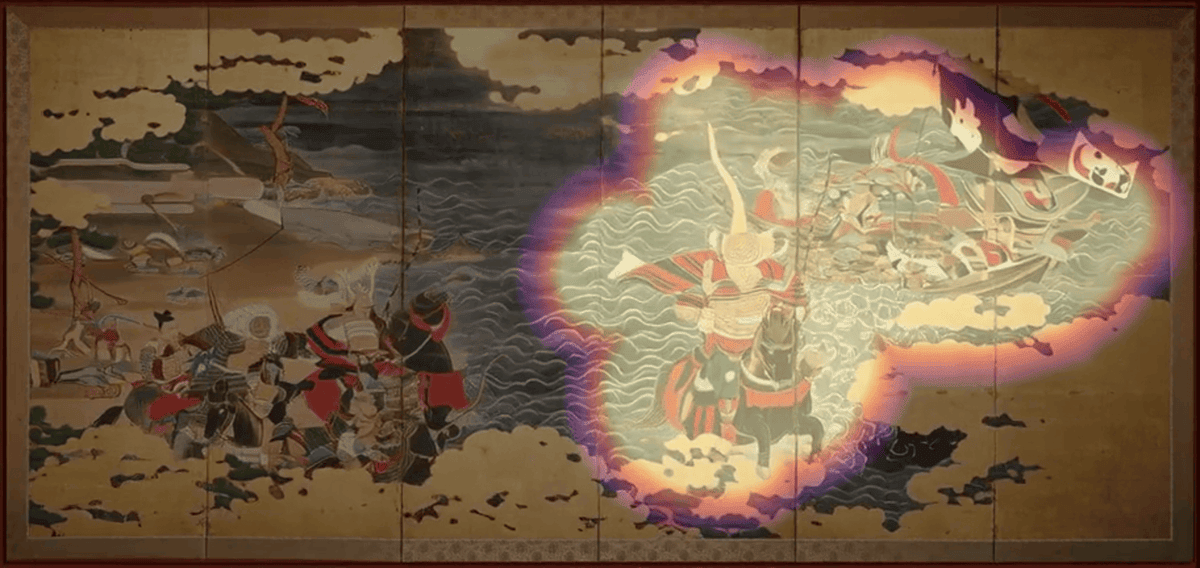} \\

\bottomrule
\end{tabular}

\caption{Two qualitative examples showing the temporal masks obtained with SAM3 at frames 1, 40, and 80. The video overlaid is the prior generated with Wan model.}
\label{fig:additional_sam3_overlay}
\end{center}

\clearpage
\section{VSR Baseline Qualitative Comparison}
\label{app:vsr-comparison}

To illustrate the VSR discussion in \cref{sect:related}, \cref{fig:vsr-comparison} compares, at $t=2$s on the same FrescoArchive input, the low-resolution prior (Wan2.2-I2V at native resolution), FlashVSR applied to that prior, and FrescoDiffusion. The bottom row zooms on the lower-left quadrant of each frame. FlashVSR sharpens the texture of the prior but reproduces its layout almost exactly, including the same static figures and the empty foreground. FrescoDiffusion instead populates the same region with new content (e.g., the horses and additional figures), confirming that VSR can refine an already-determined trajectory but cannot invent content beyond it.

\begin{center}
\captionsetup{type=figure}

\begin{tabular}{@{}ccc@{}}
\toprule
\textbf{Prior (Wan2.2-I2V)} & \textbf{FlashVSR} & \textbf{FrescoDiffusion} \\
\midrule
\includegraphics[width=0.30\textwidth]{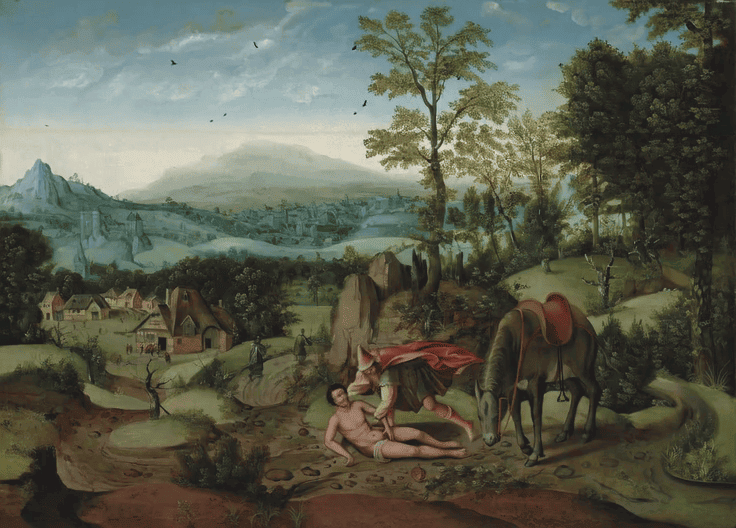} &
\includegraphics[width=0.30\textwidth]{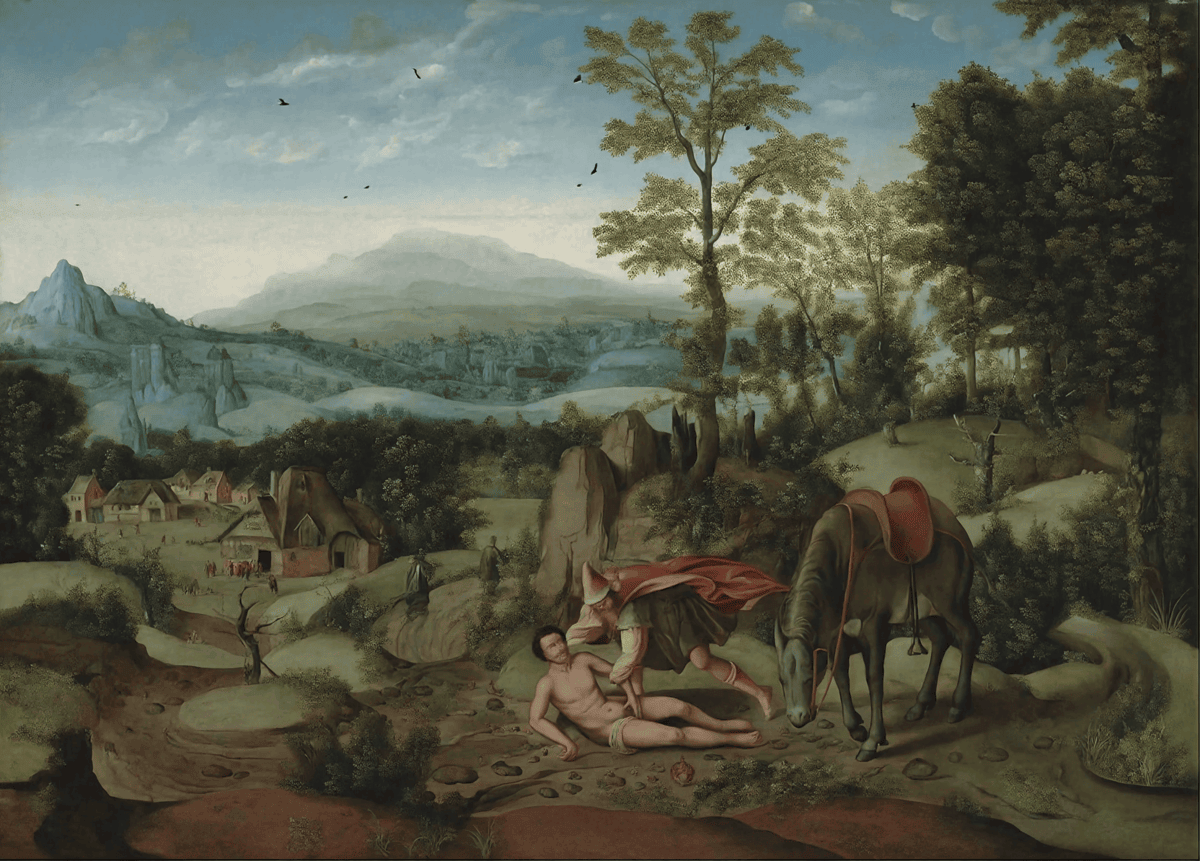} &
\includegraphics[width=0.30\textwidth]{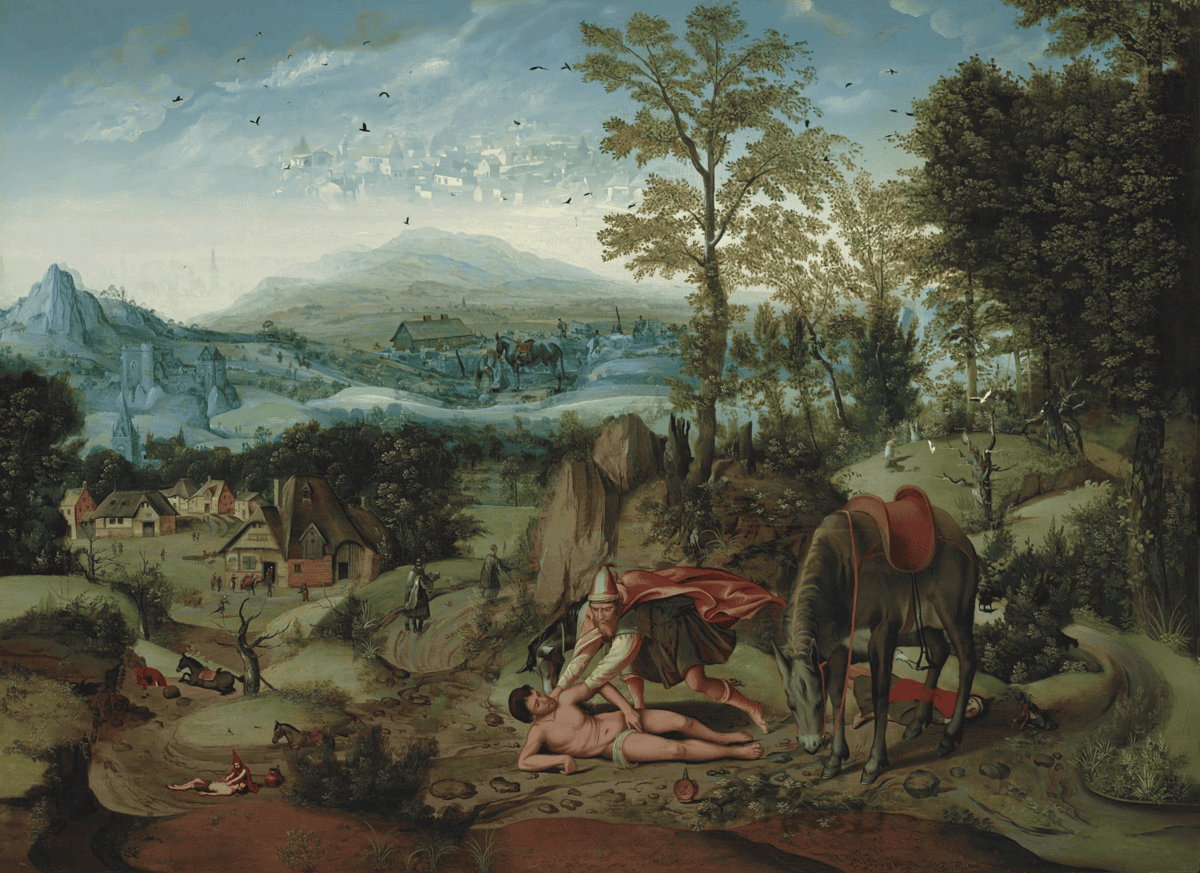} \\
\includegraphics[width=0.30\textwidth]{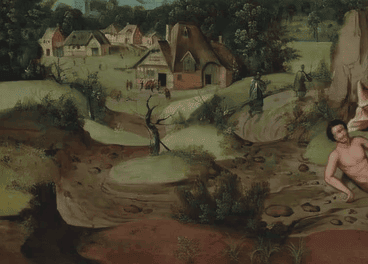} &
\includegraphics[width=0.30\textwidth]{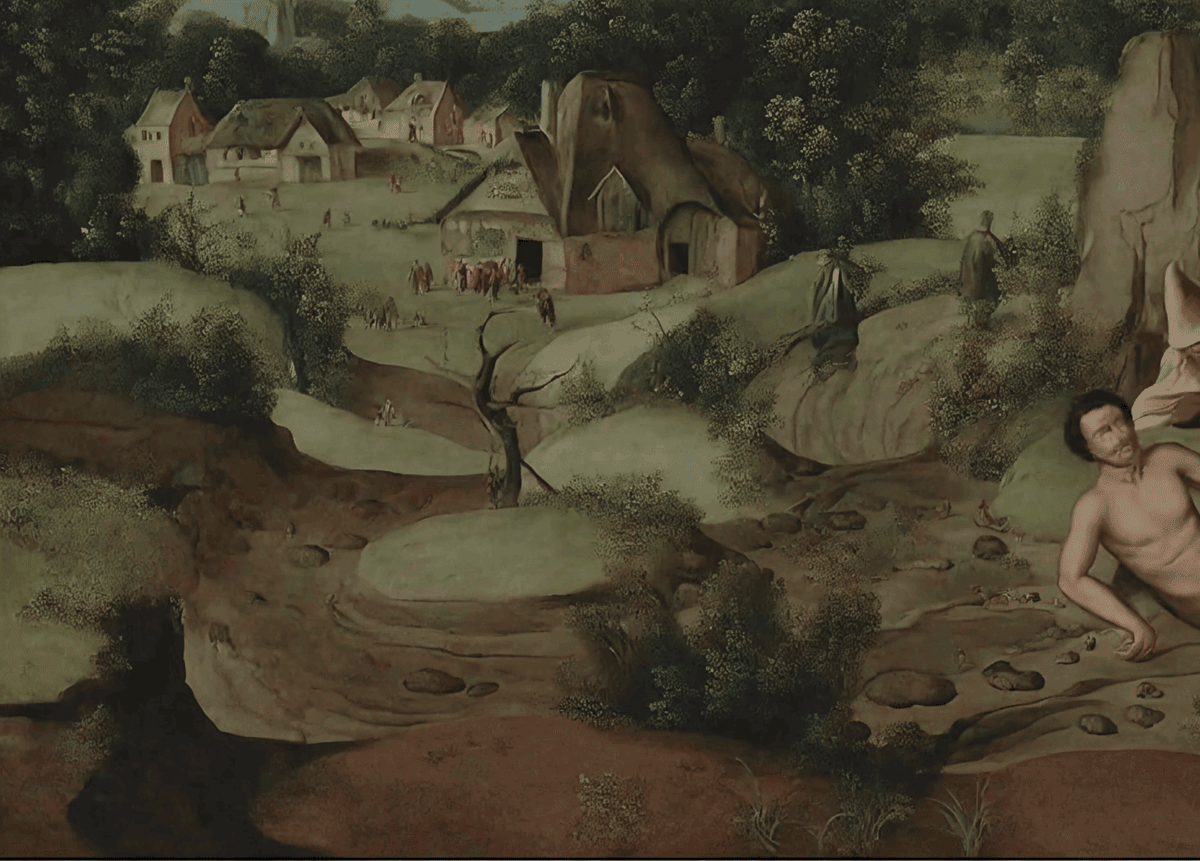} &
\includegraphics[width=0.30\textwidth]{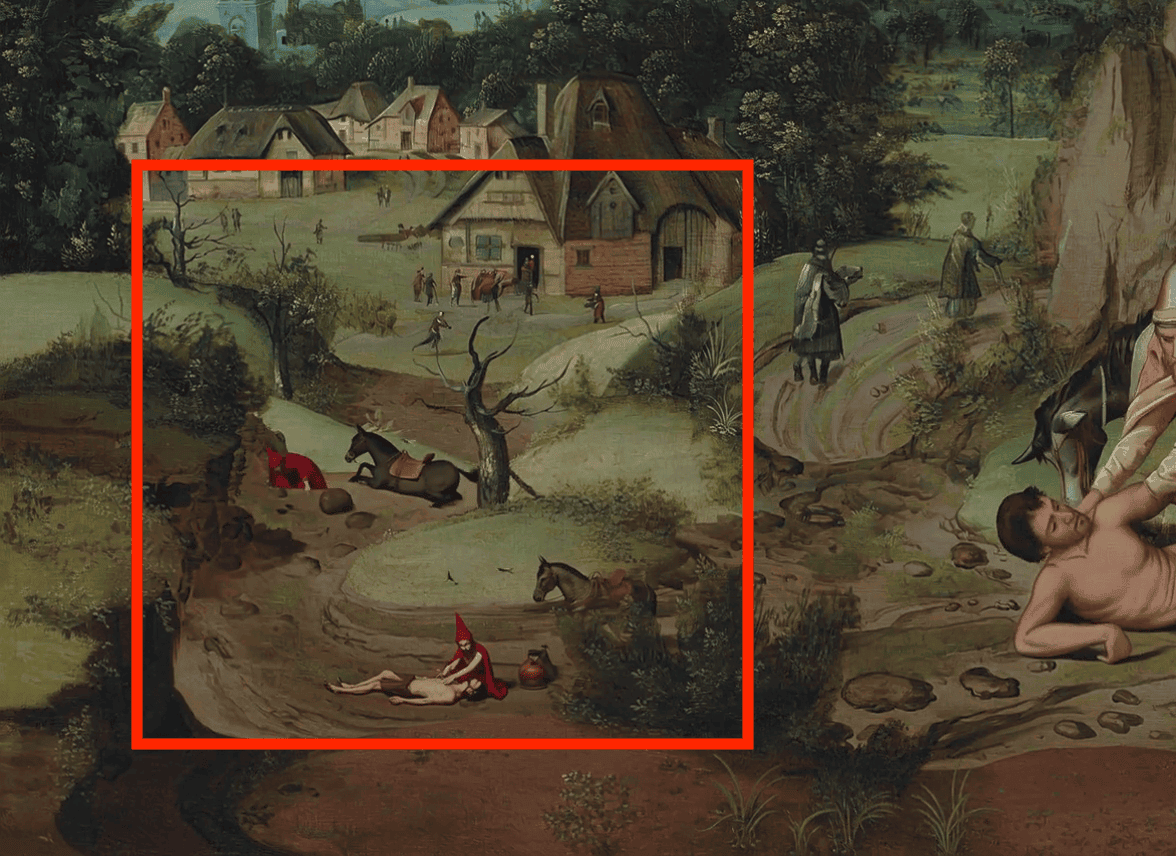} \\
\bottomrule
\end{tabular}

\caption{VSR baseline qualitative comparison. Top row: full frame for the low-resolution prior, FlashVSR, and FrescoDiffusion. Bottom row: zoom on the lower-left quadrant of each frame above. FlashVSR stays close to the prior's layout, while FrescoDiffusion introduces new content (e.g., the horses) absent from both the prior and the VSR output.}
\label{fig:vsr-comparison}
\end{center}

\clearpage
\section{Motion Quantity at Full Resolution}
\label{app:motion-quantity}

We measure optical-flow magnitude using RAFT-Large~\cite{teed2020raft} on frames resized to $2560\times1440$. For each method, we report the median flow magnitude after removing outliers above $Q_3+1.5\,\mathrm{IQR}$, quantifying how much net motion each method introduces at full resolution.

\begin{table}[H]
  \centering
  \caption{Optical flow magnitude (RAFT-Large, median over pixels, $Q_3+1.5\,\text{IQR}$ outliers removed) at $2560\times1440$ on FrescoArchive (FA) and VBench-I2V.}
  \label{tab:optical_flow}
  \begin{tabular}{lcc}
    \toprule
    Method & FA & VBench-I2V \\
    \midrule
    Wan (native resolution) & 0.288 & 5.837 \\
    FlashVSR                & 0.247 & 6.421 \\
    \midrule
    DemoFusion              & 1.251 & 4.638 \\
    DynamicScaler           & 24.085 & 17.358 \\
    MultiDiffusion          & 26.217 & 18.469 \\
    \midrule
    FrescoDiffusion         & 1.318 & 4.424 \\
    R-FrescoDiffusion       & 0.885 & 4.374 \\
    \bottomrule
  \end{tabular}
\end{table}

\Cref{tab:optical_flow} reveals three regimes. FlashVSR and native Wan produce near-zero flow on FA, below $0.3$~px: VSR refines texture but inherits the motion of its low-resolution prior. MultiDiffusion and DynamicScaler produce an order of magnitude more flow, reaching $26$ and $24$~px on FA, yet score lowest on motion preference in the user study (\cref{tab:user_study}), because the large flow reflects tiles drifting independently rather than coherent animation. FrescoDiffusion and R-FrescoDiffusion occupy the intermediate regime, with $1.3$ and $0.9$~px respectively, producing motion that users consistently prefer. Motion coherence, not motion quantity, drives preference. For the coherent-motion methods, namely native Wan, FlashVSR, DemoFusion, FrescoDiffusion, and R-FrescoDiffusion, flow magnitudes are lower on FrescoArchive than on VBench-I2V. This gap stems from the different content of the two image sets rather than from less motion. FrescoArchive frescoes spread many small subjects across a large composition, so each animated element is small relative to the frame and its displacement spans only a few pixels. VBench-I2V images instead center on one or a few large subjects that occupy much of the frame, so an equivalent proportional motion produces a larger pixel displacement. The two drift baselines, MultiDiffusion and DynamicScaler, are the exception, with higher flow on FrescoArchive: their motion is dominated by tiles drifting in uncorrelated directions, an effect amplified by the many independent scenes of a fresco rather than by genuine subject motion.

\clearpage
\section{FrescoDiffusion Additional Examples}
\label{app:extra-results}

\begin{center}
\captionsetup{type=figure}

\begin{tabular}{@{}ccc@{}}
\toprule
\textbf{frame 1} & \textbf{frame 48} & \textbf{frame 80} \\
\midrule
\includegraphics[width=0.30\textwidth]{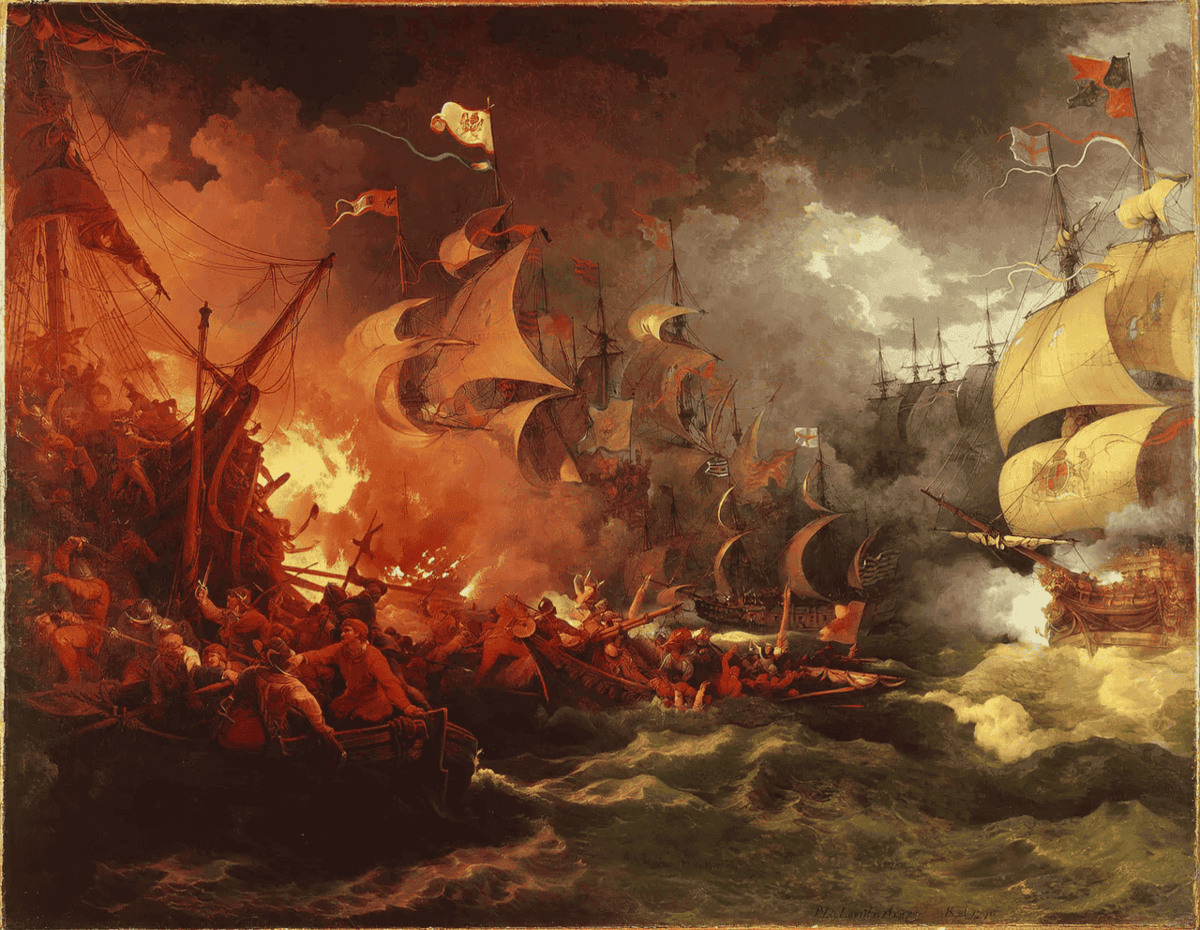} &
\includegraphics[width=0.30\textwidth]{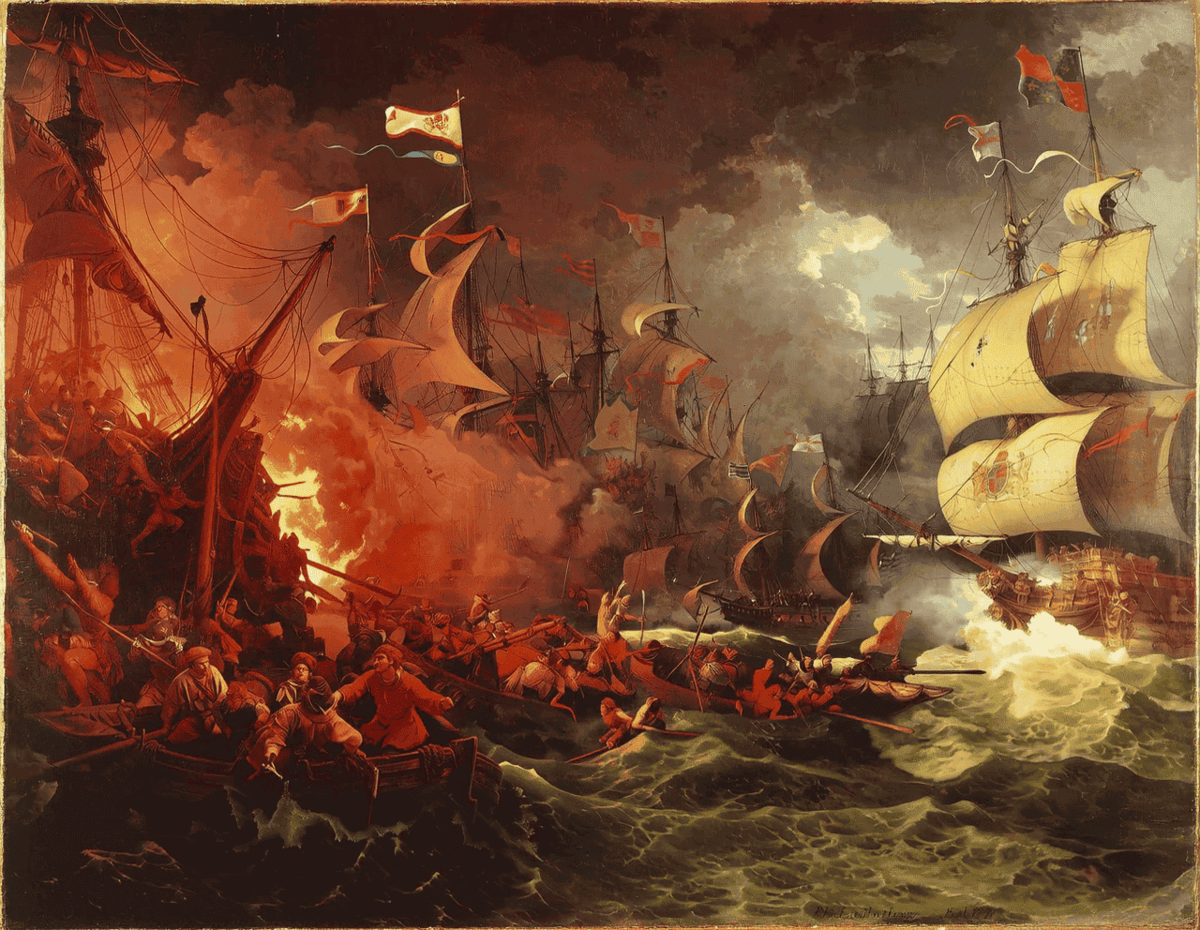} &
\includegraphics[width=0.30\textwidth]{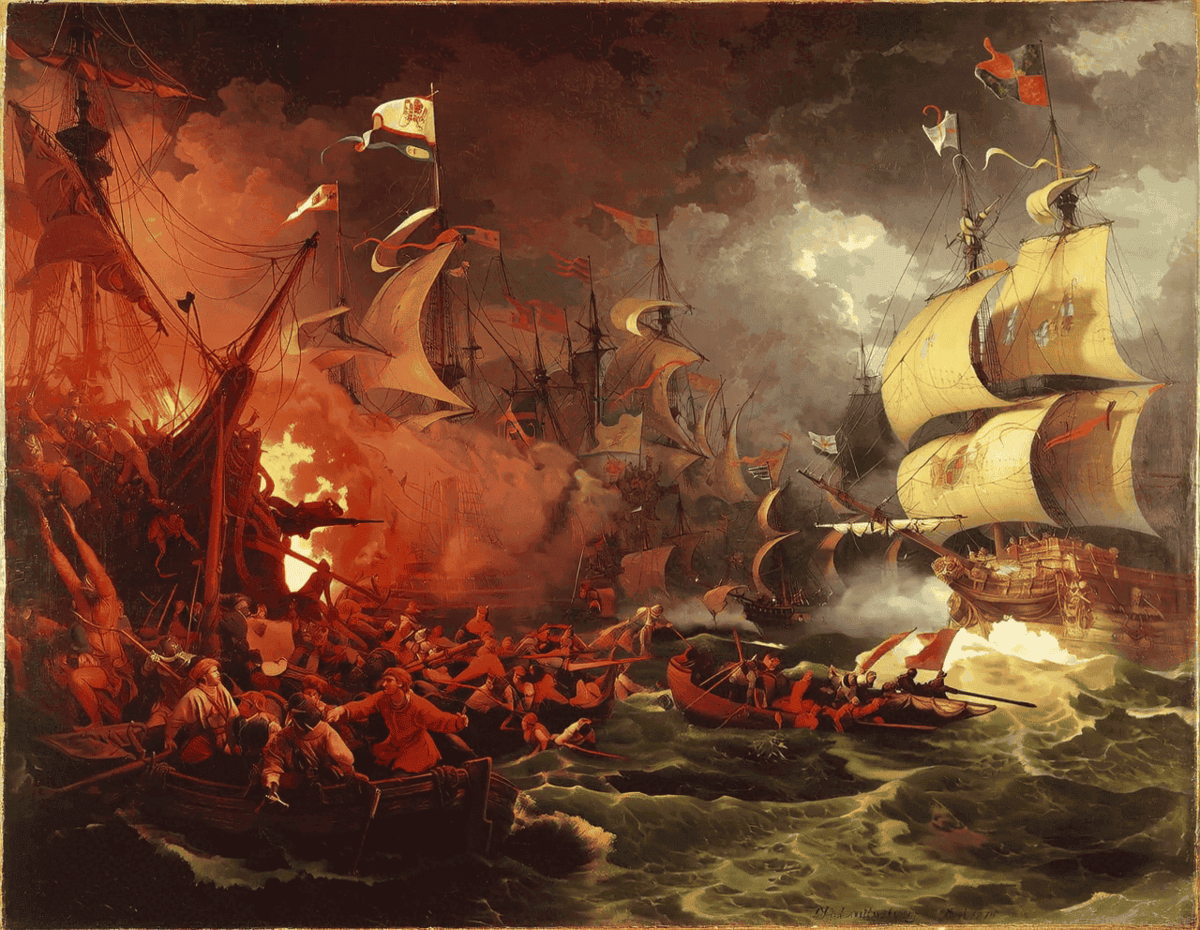} \\
\includegraphics[width=0.30\textwidth]{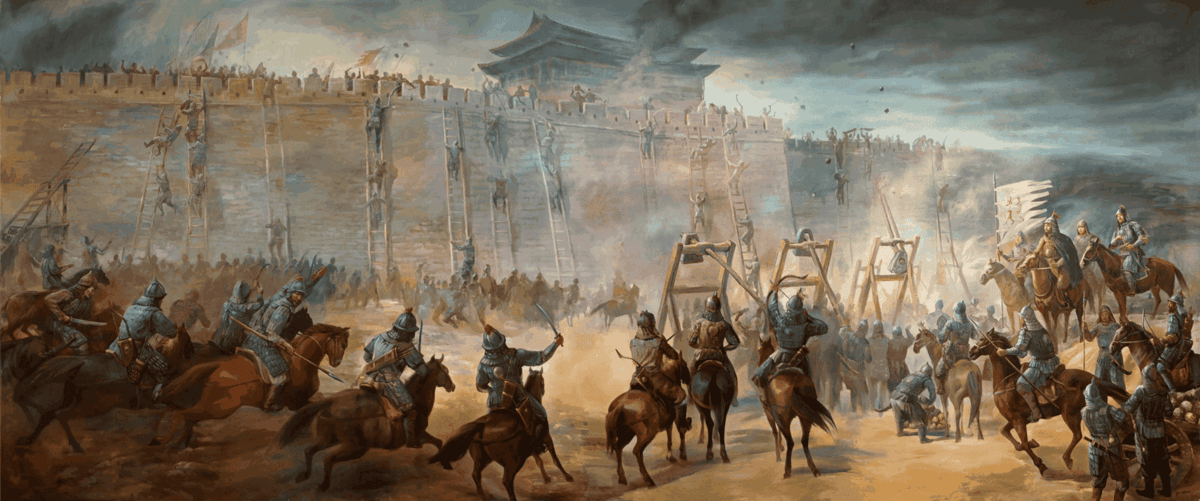} &
\includegraphics[width=0.30\textwidth]{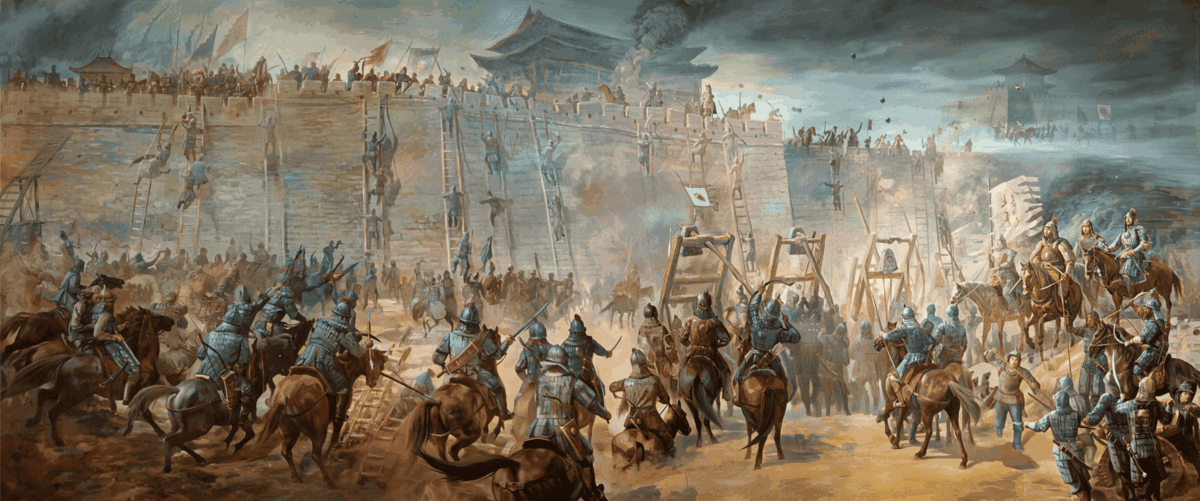} &
\includegraphics[width=0.30\textwidth]{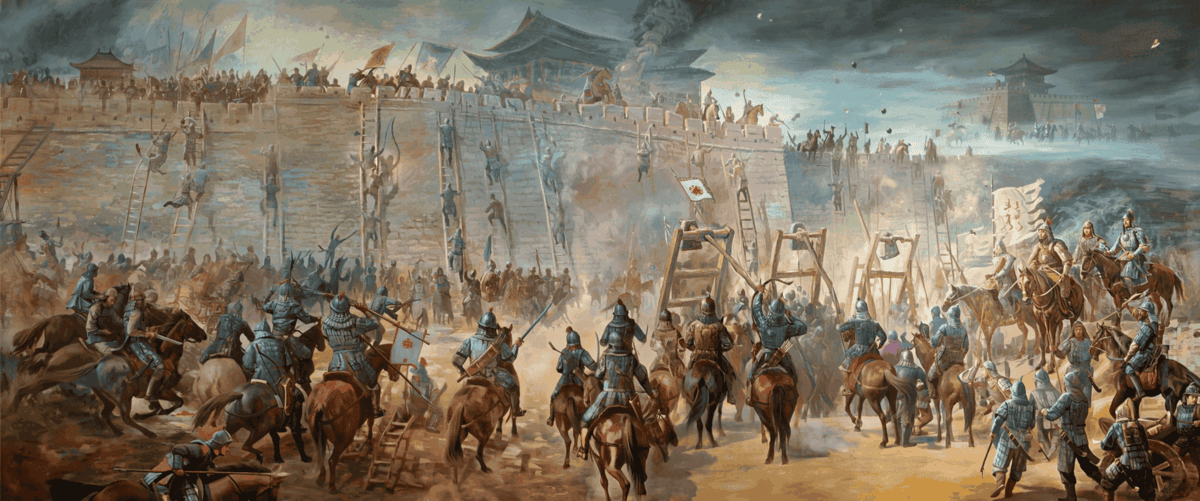} \\
\includegraphics[width=0.30\textwidth]{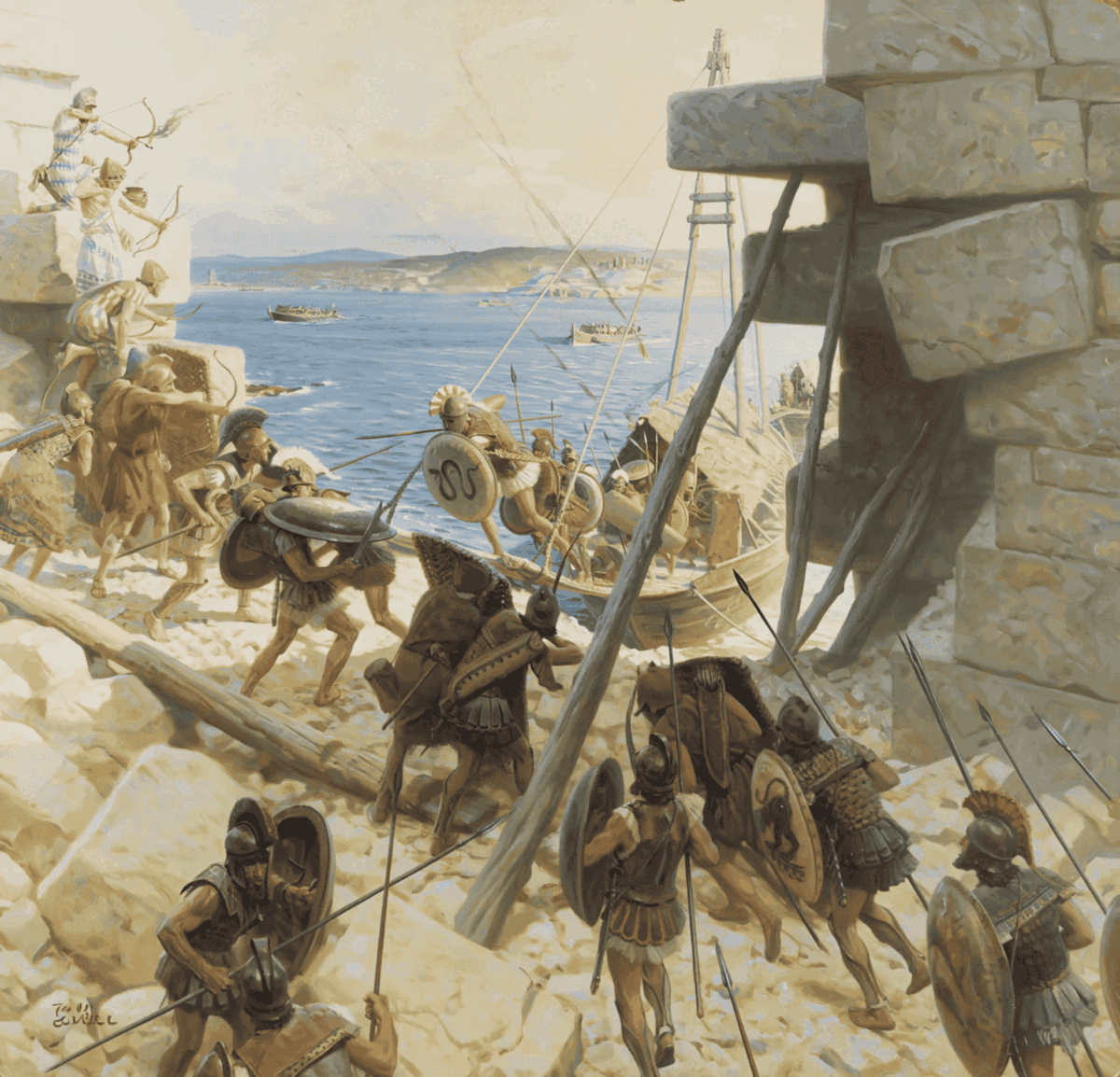} &
\includegraphics[width=0.30\textwidth]{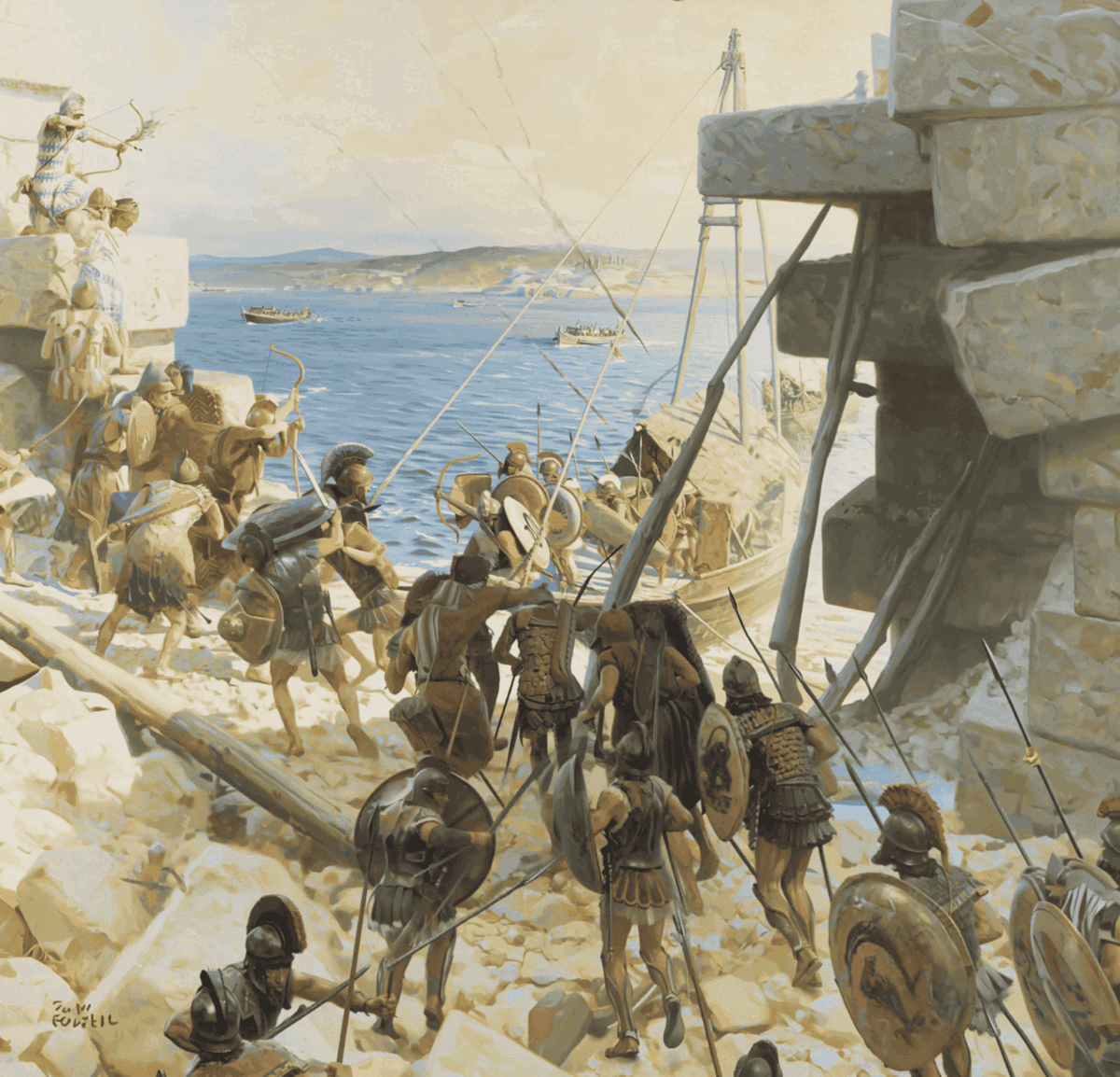} &
\includegraphics[width=0.30\textwidth]{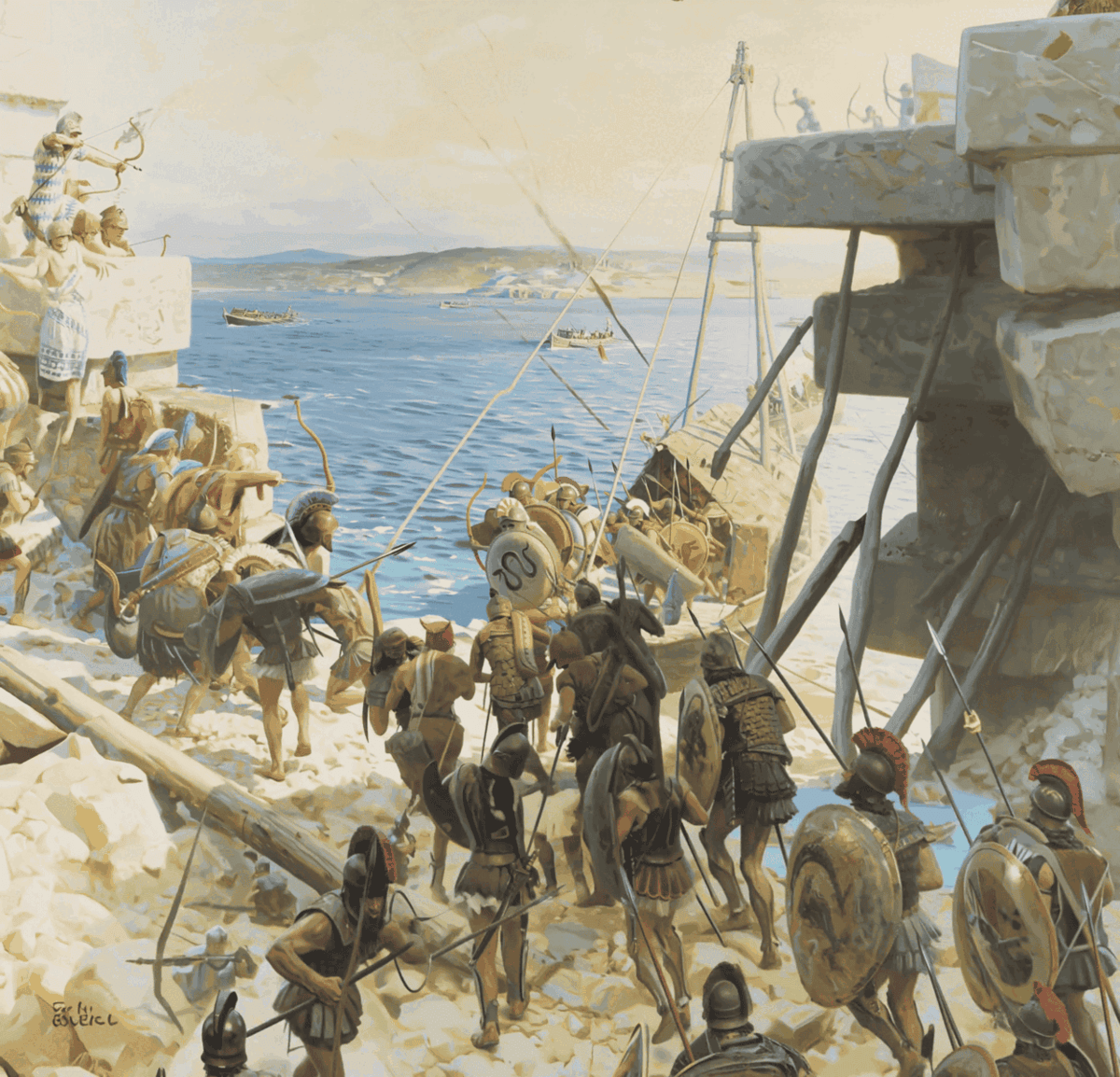} \\
\includegraphics[width=0.30\textwidth]{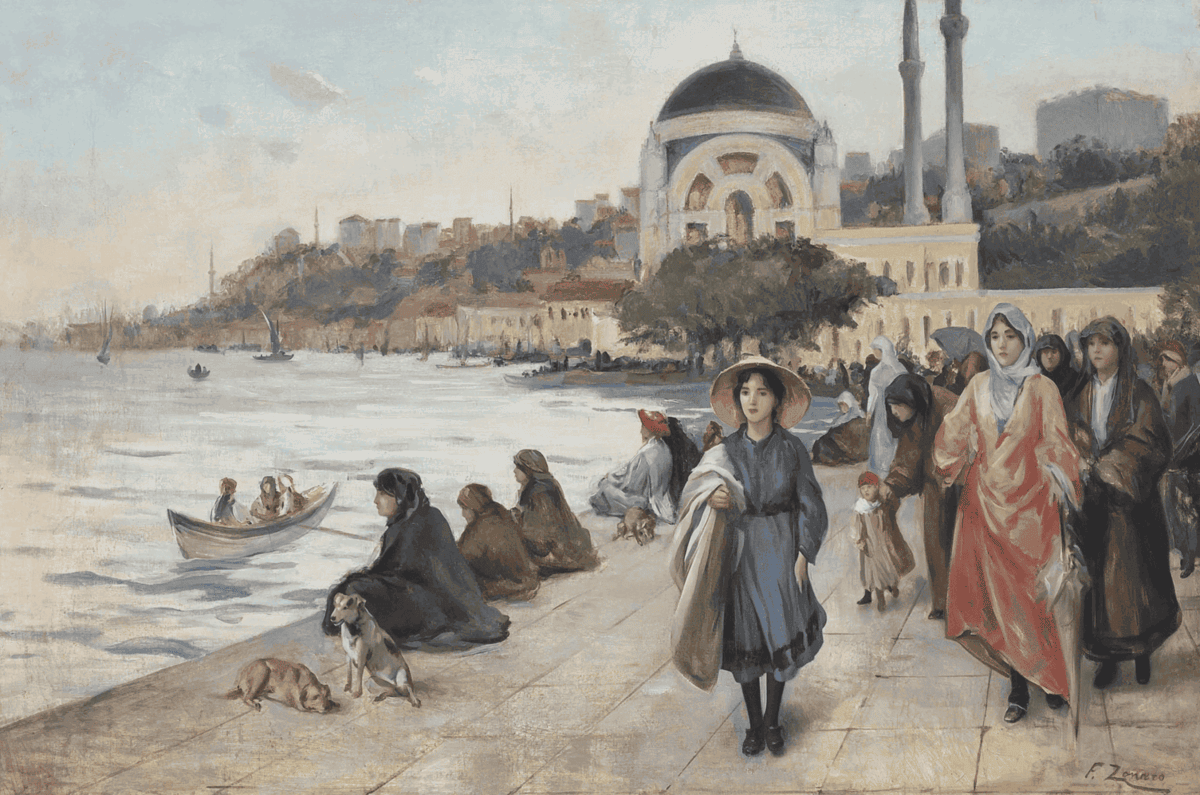} &
\includegraphics[width=0.30\textwidth]{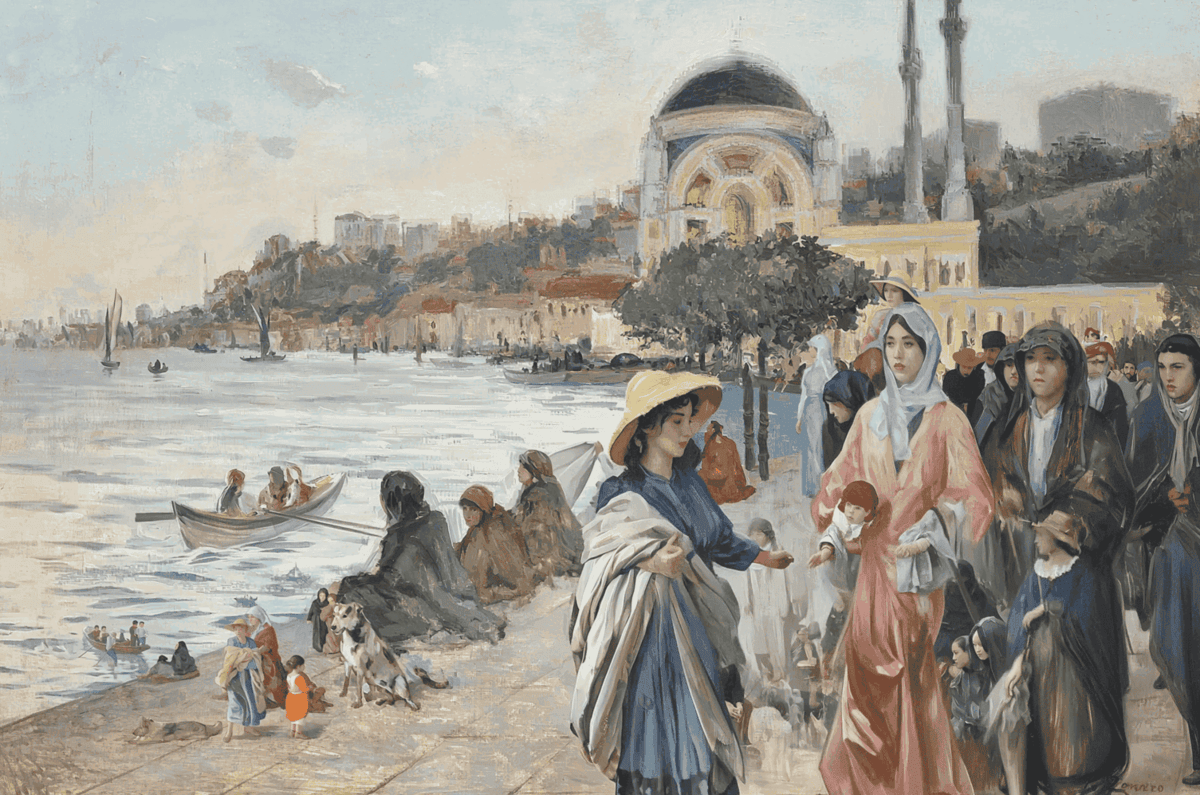} &
\includegraphics[width=0.30\textwidth]{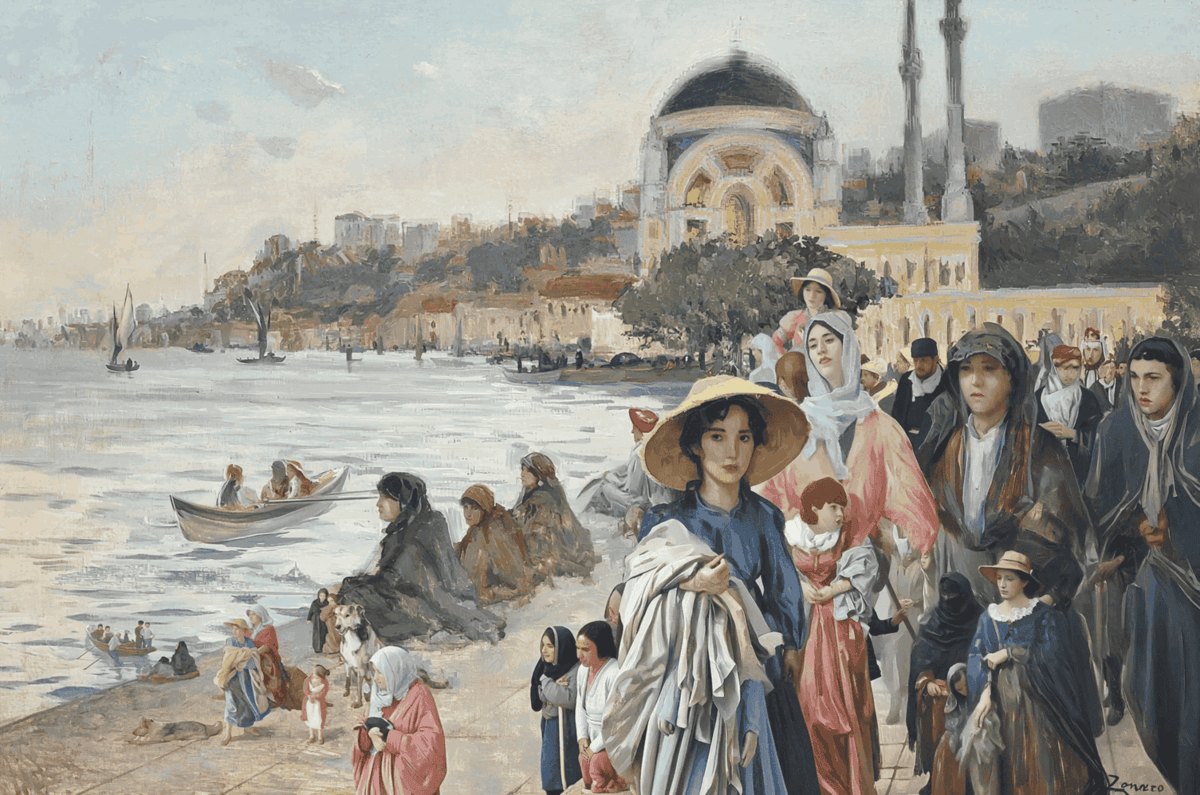} \\
\bottomrule
\end{tabular}

\caption{Additional examples obtained on our FrescoArchive dataset. First 2 rows obtained with FrescoDiffusion and last 2 rows obtained with R-FrescoDiffusion; columns show the same frame indices.}
\label{fig:additional}
\end{center}

\begin{center}
\captionsetup{type=figure}

\begin{tabular}{@{}ccc@{}}
\toprule
\textbf{frame 1} & \textbf{frame 48} & \textbf{frame 80} \\
\midrule
\includegraphics[width=0.30\textwidth]{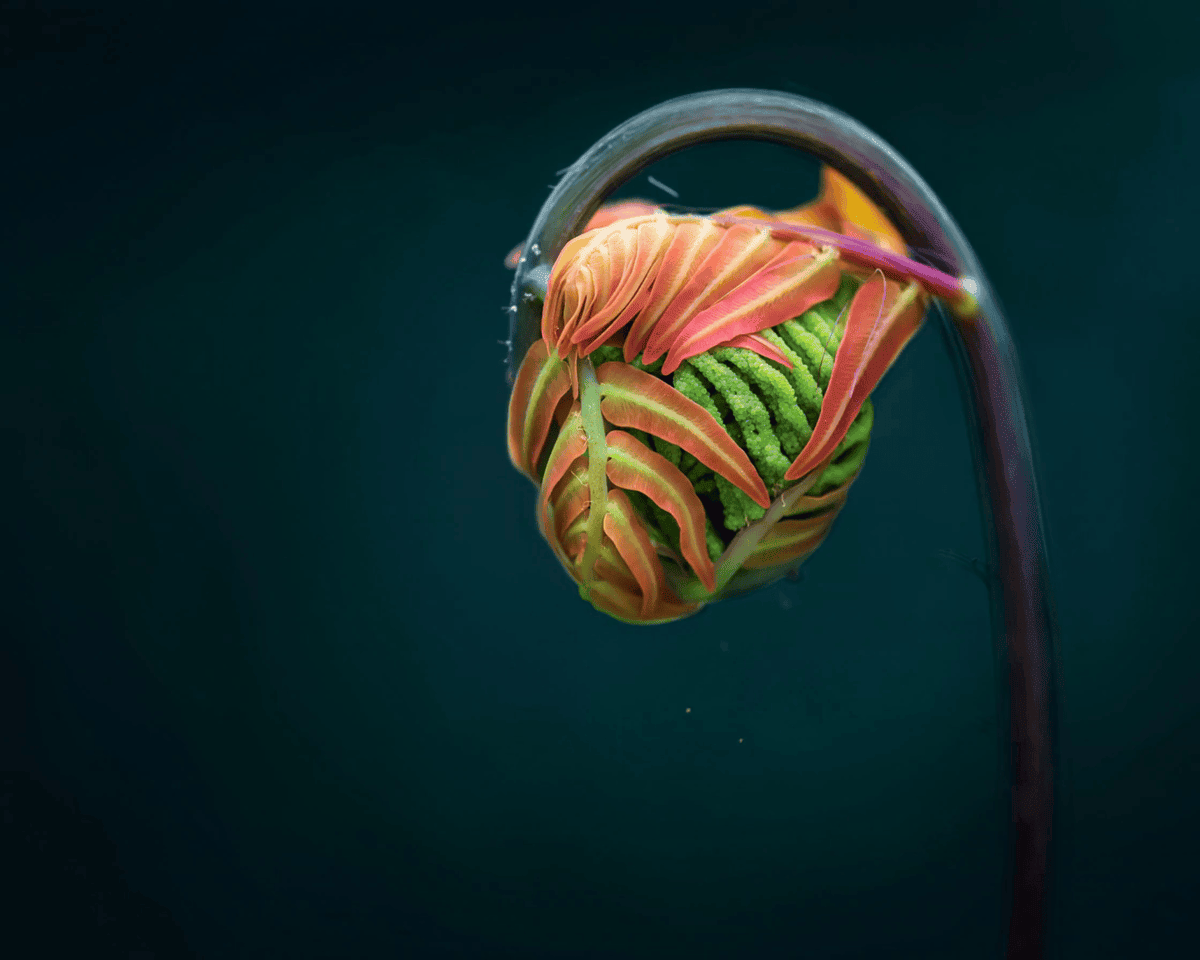} &
\includegraphics[width=0.30\textwidth]{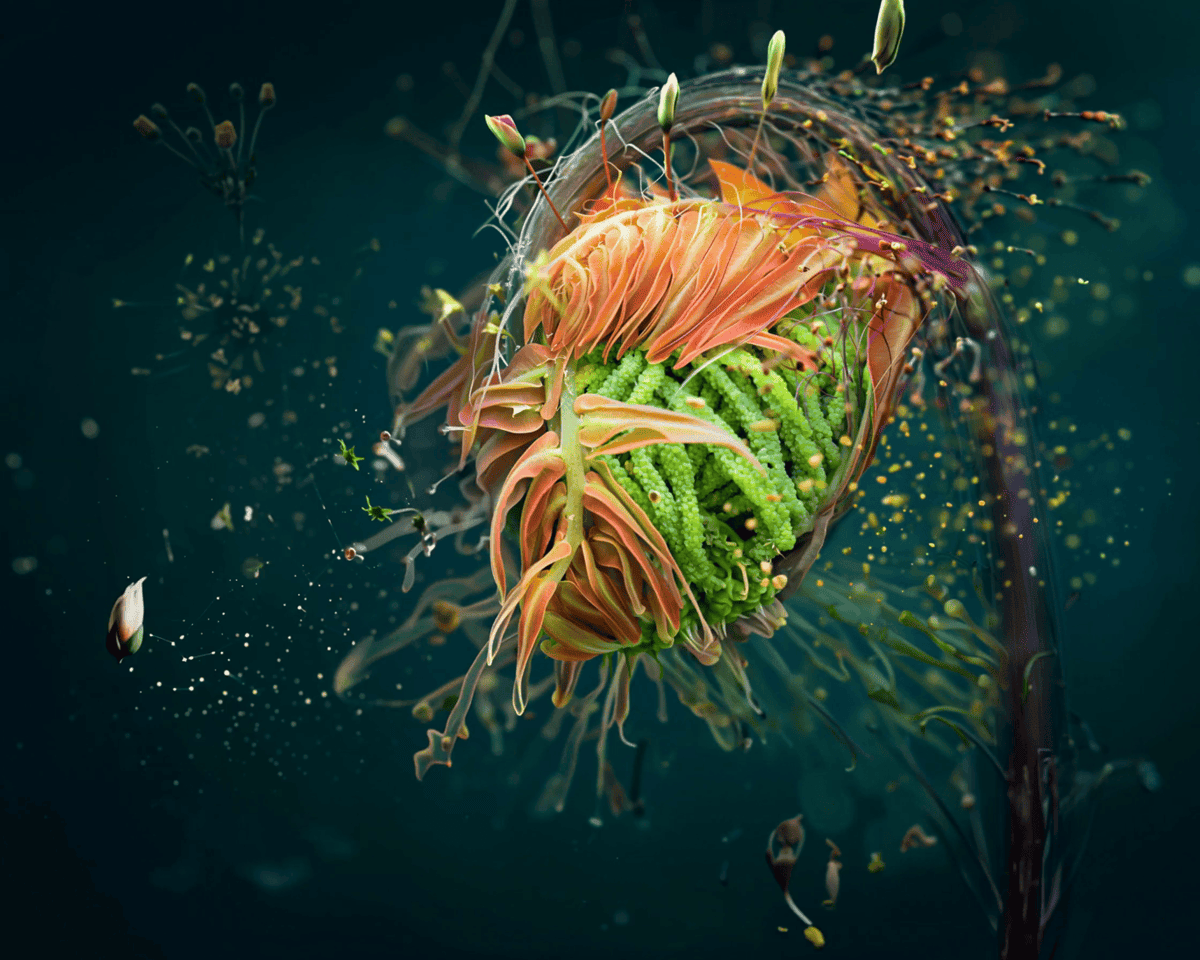} &
\includegraphics[width=0.30\textwidth]{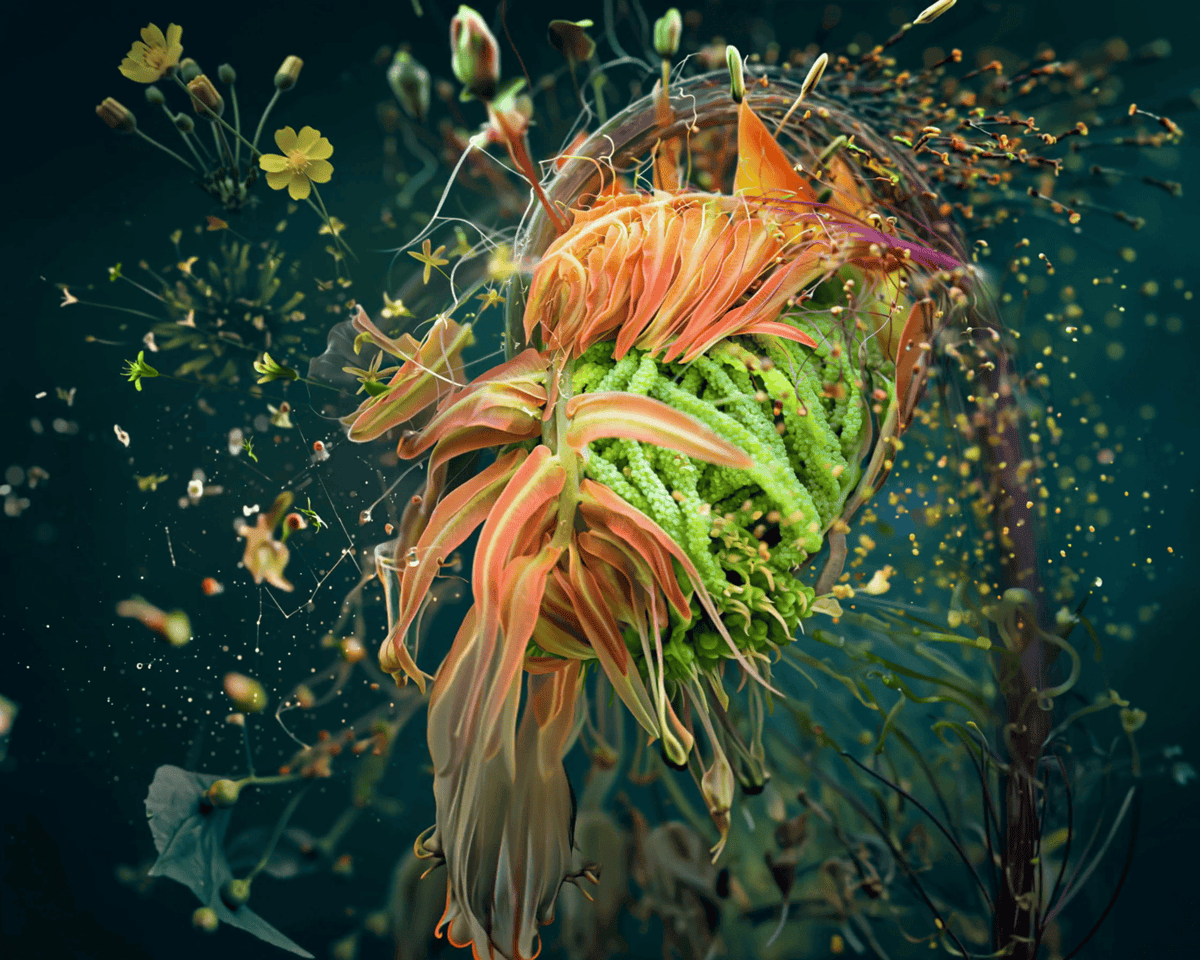} \\
\includegraphics[width=0.30\textwidth]{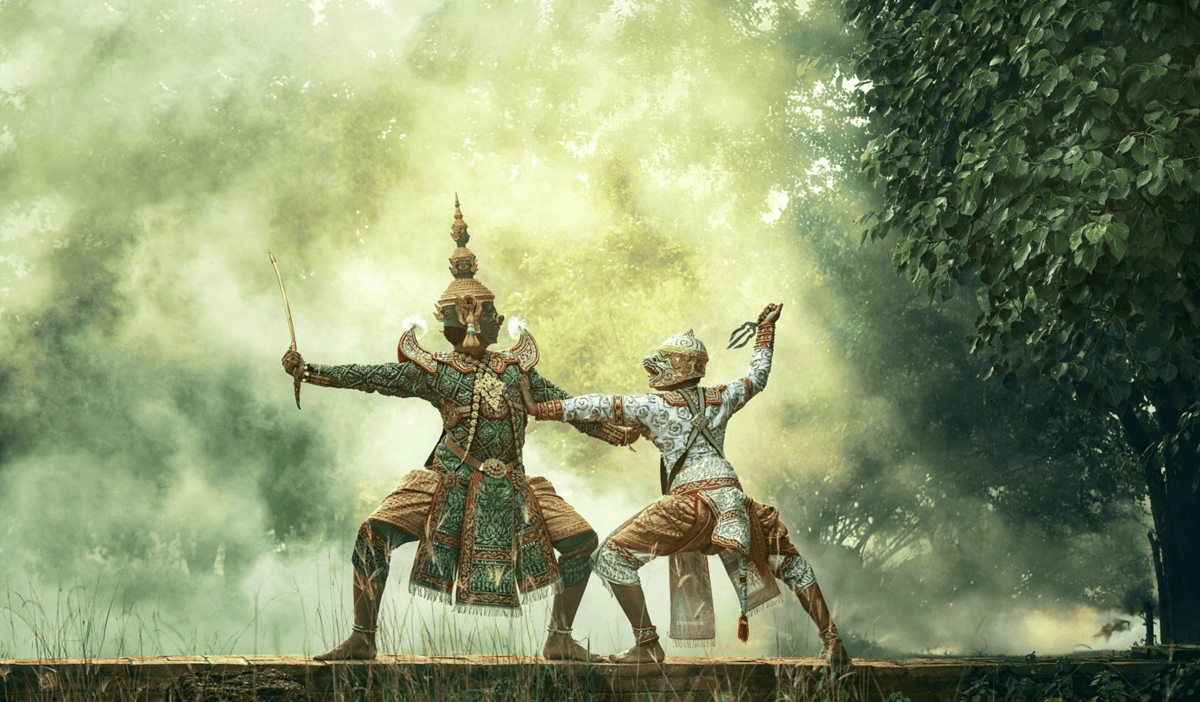} &
\includegraphics[width=0.30\textwidth]{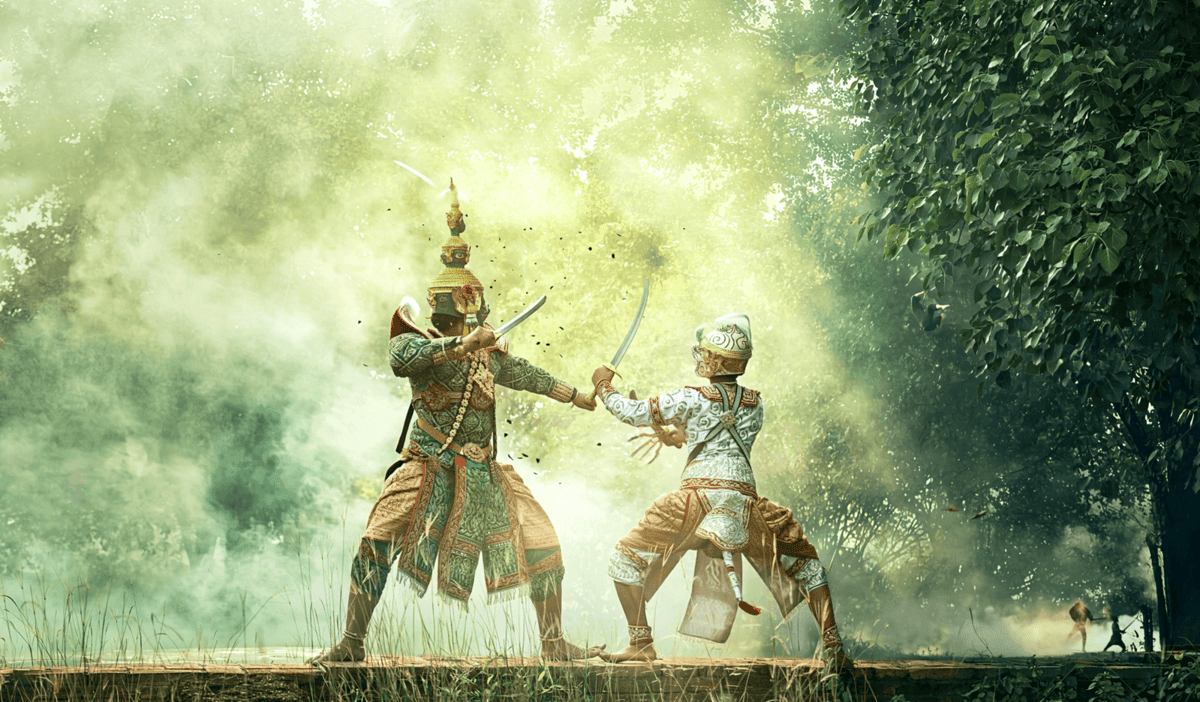} &
\includegraphics[width=0.30\textwidth]{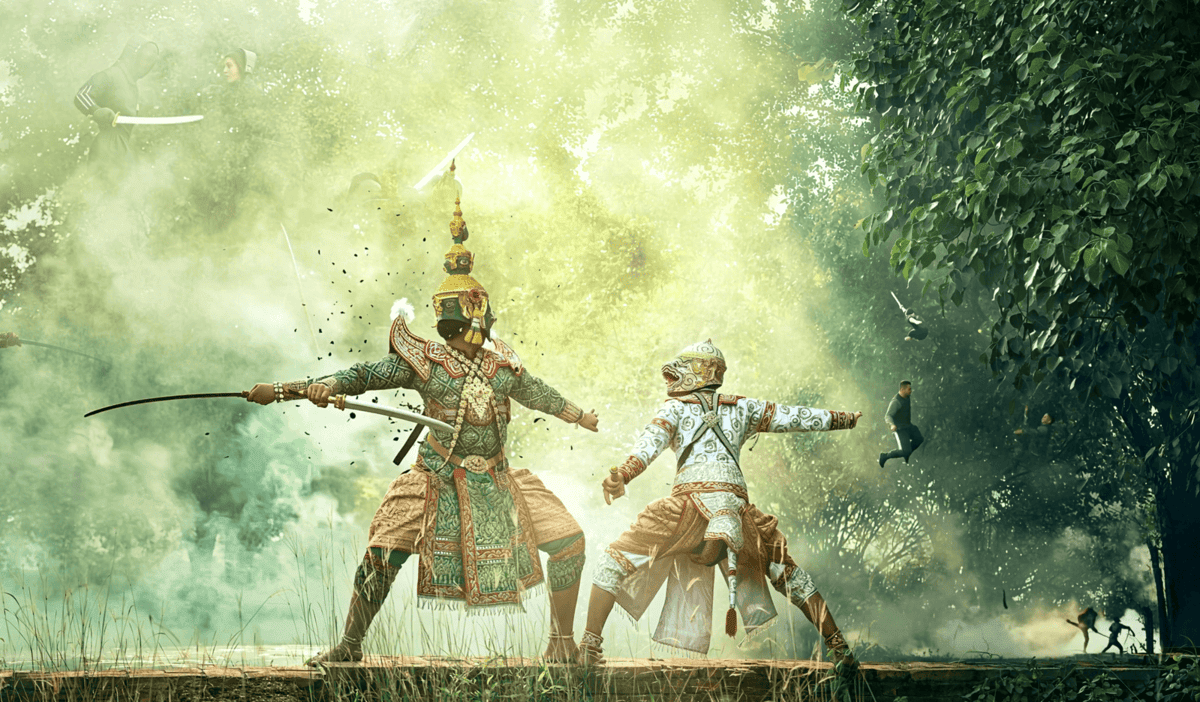} \\
\includegraphics[width=0.20\textwidth]{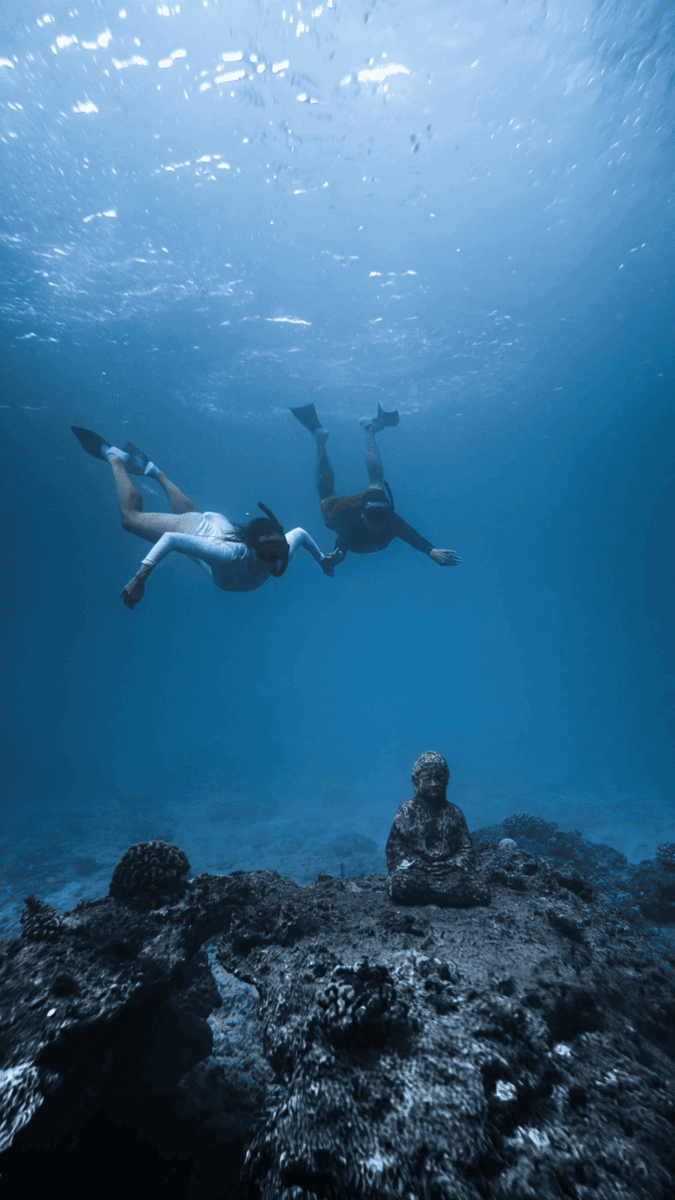} &
\includegraphics[width=0.20\textwidth]{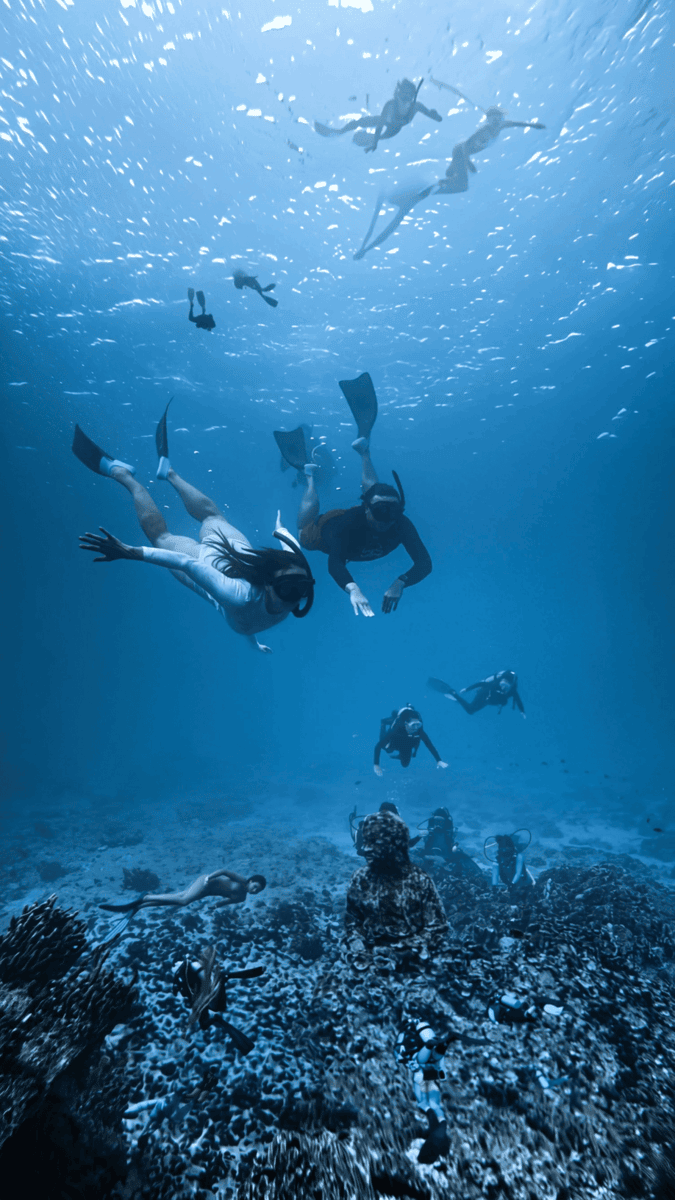} &
\includegraphics[width=0.20\textwidth]{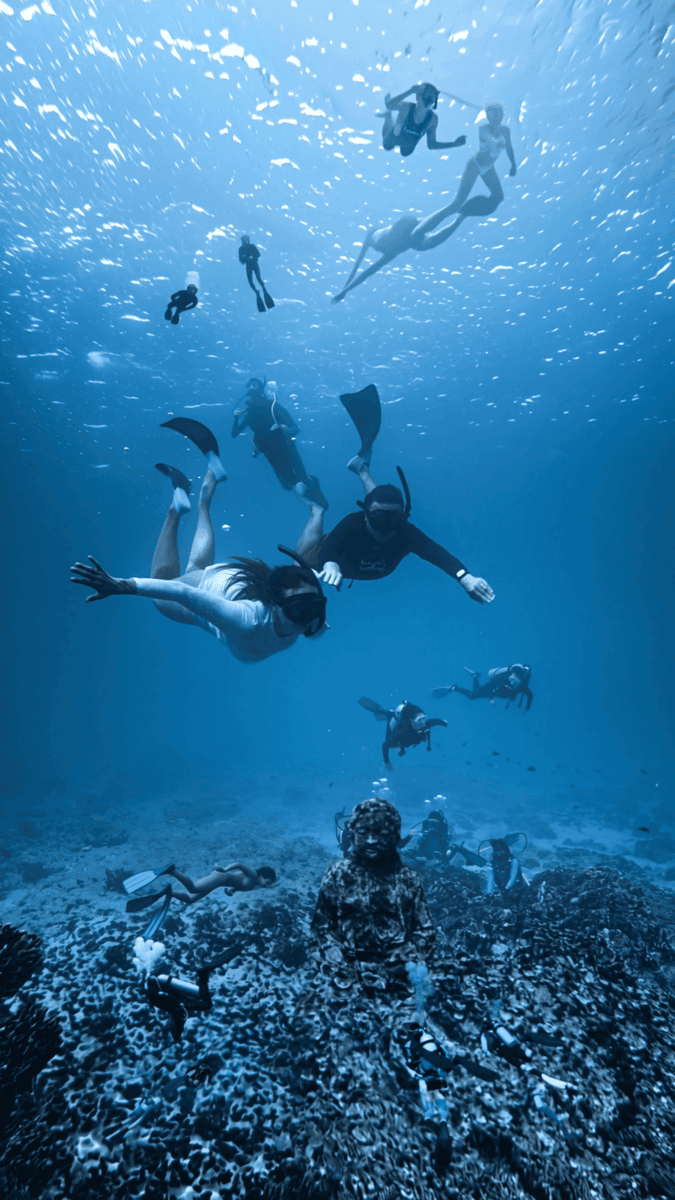} \\
\includegraphics[width=0.30\textwidth]{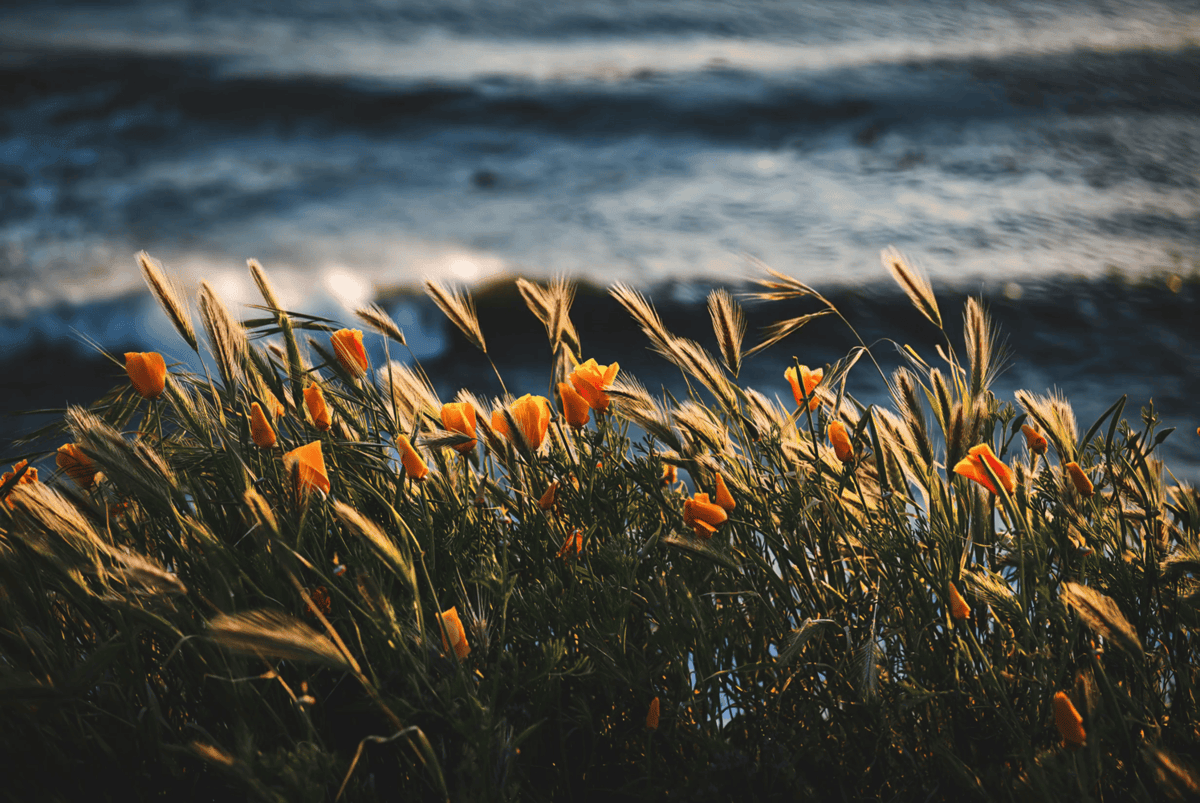} &
\includegraphics[width=0.30\textwidth]{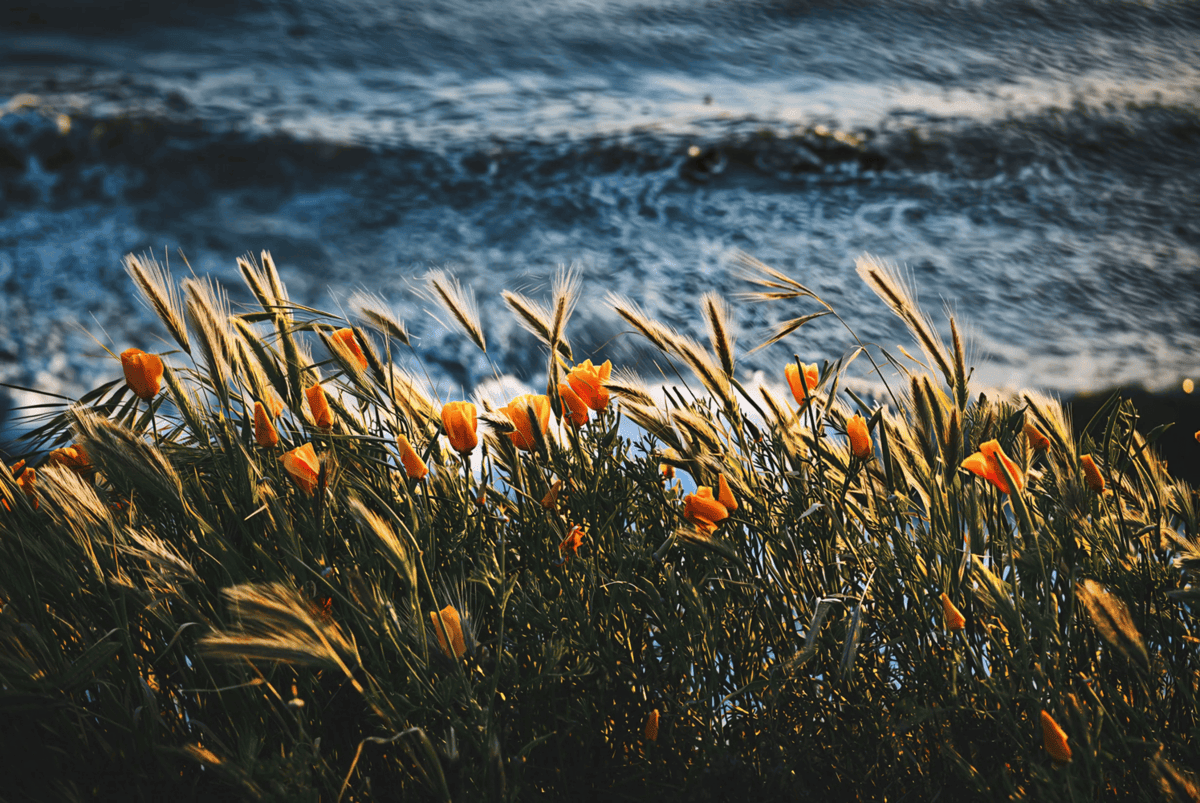} &
\includegraphics[width=0.30\textwidth]{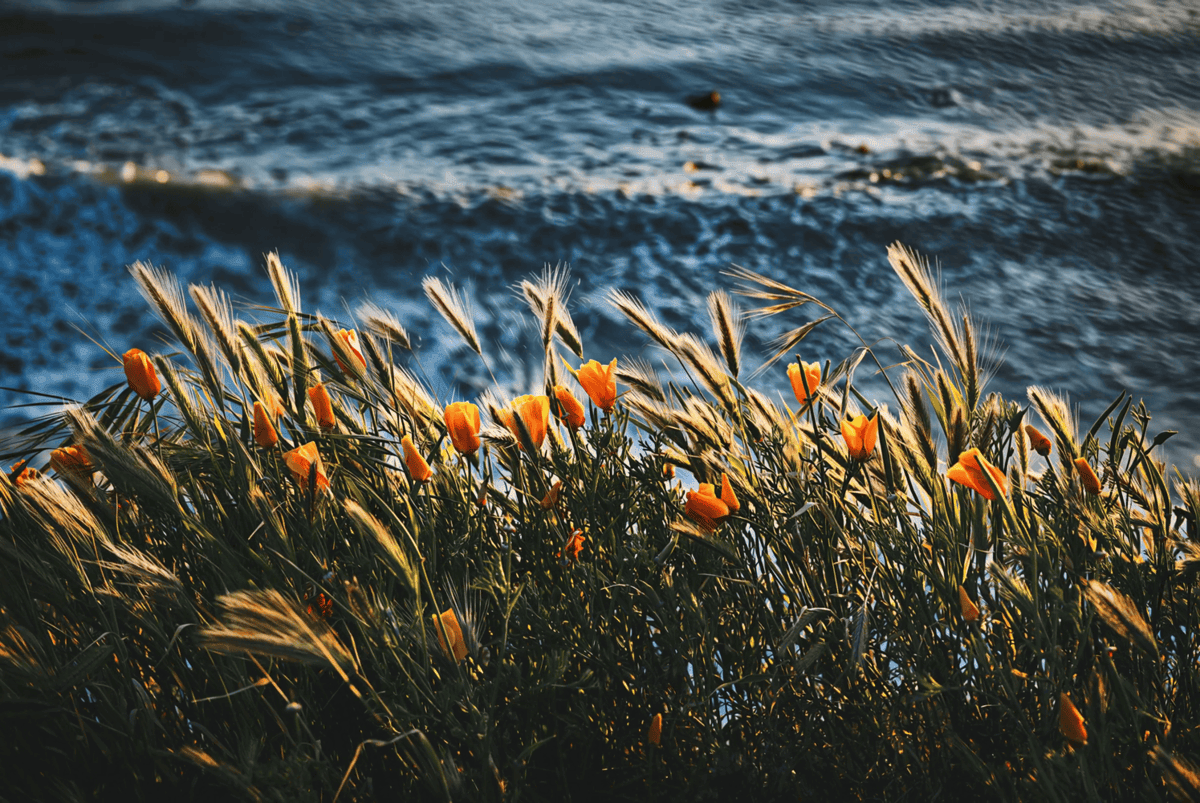} \\
\includegraphics[width=0.30\textwidth]{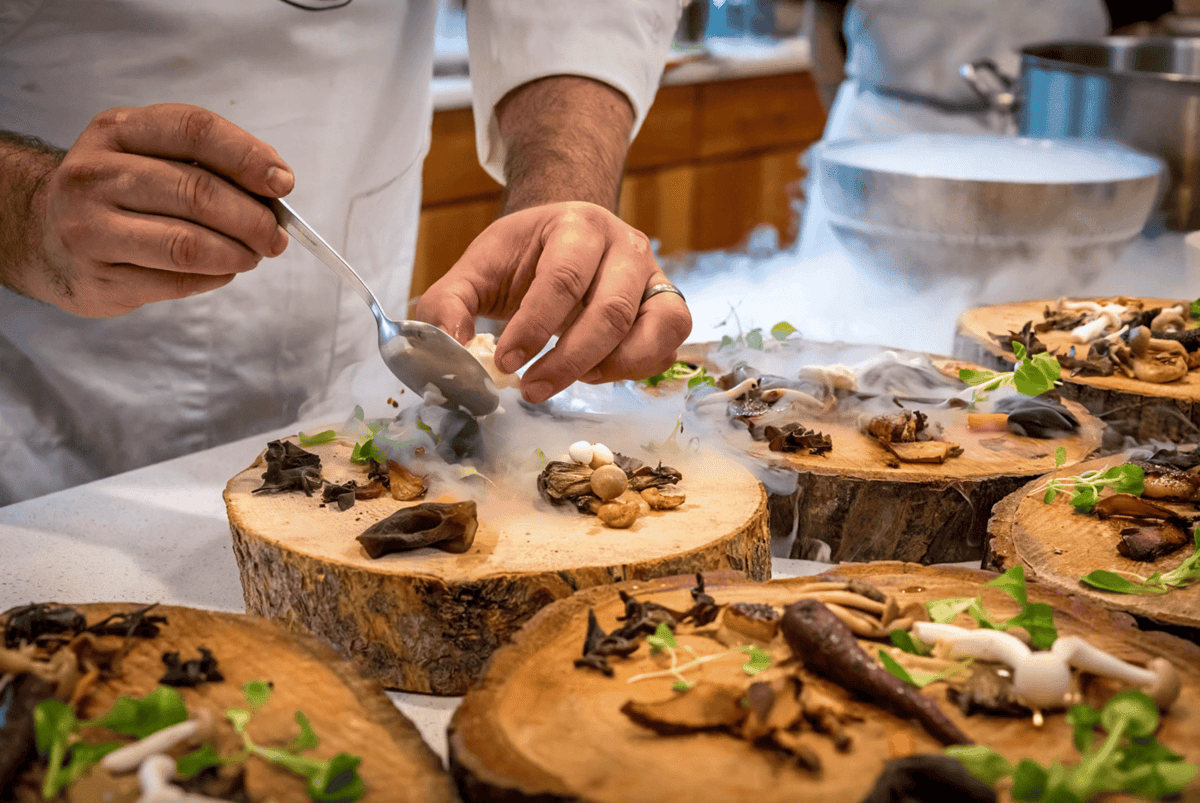} &
\includegraphics[width=0.30\textwidth]{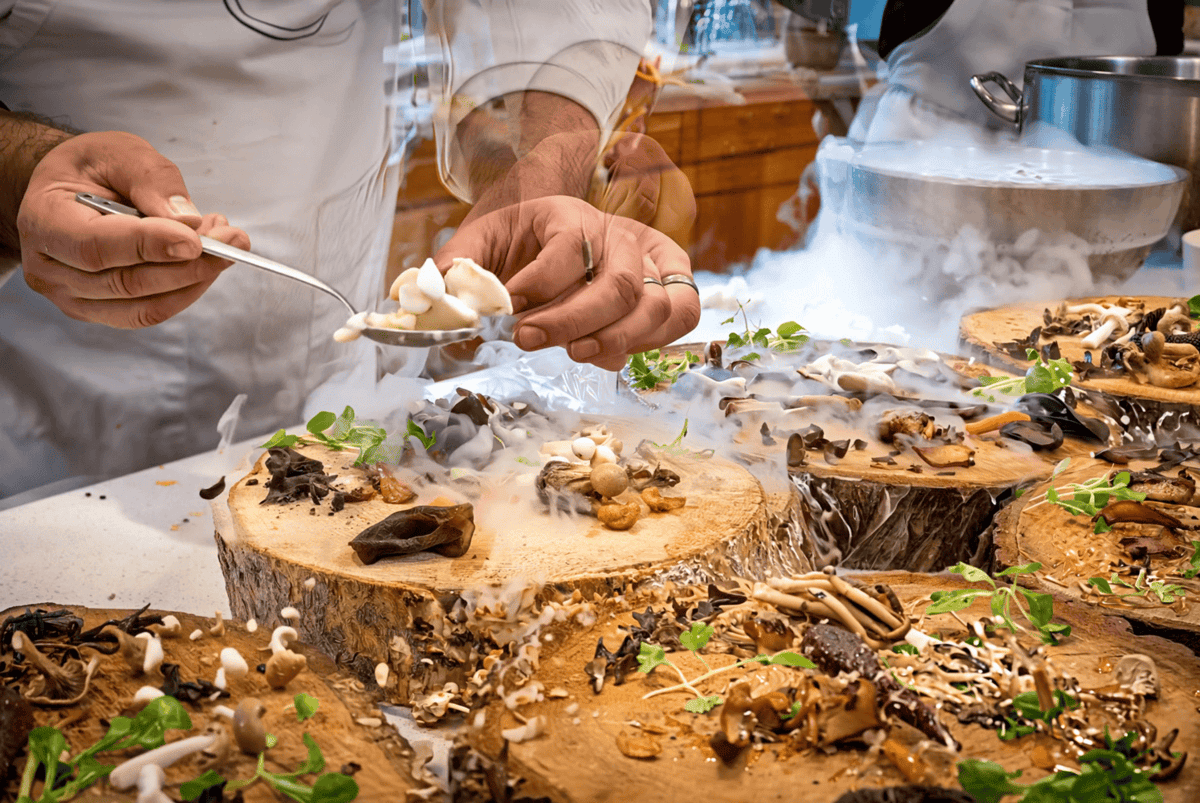} &
\includegraphics[width=0.30\textwidth]{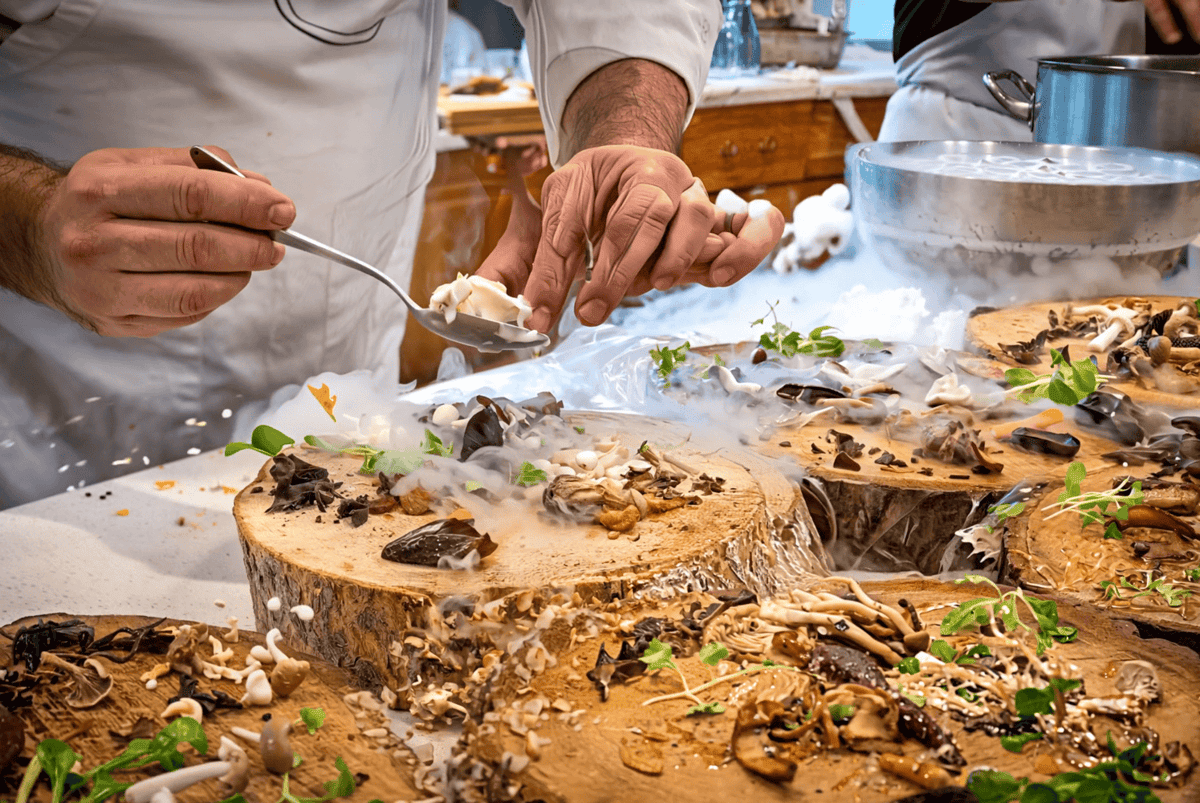} \\
\bottomrule
\end{tabular}

\caption{Additional examples obtained on VBench I2V dataset. First 3 rows obtained with FrescoDiffusion and 2 last rows obtained with R-FrescoDiffusion; columns show the same frame indices.}
\end{center}

\section{FrescoDiffusion closed-form solution in noise prediction setting}
\label{app:derivation-other}

Our proposed approach can be used with $\epsilon$-prediction diffusion models. We modify the FrescoDiffusion loss in \cref{eq:f2v-energy} to include the one-step approximation of the $\epsilon$-diffusion formulation:
\begin{equation}\label{eq:eps-loss}
    \ell_\text{FD}(y^\star; t) = \left\|\,\sqrt{\lambda}\,\odot\,\left[\frac{1}{\sqrt{\alpha_t}}\left(x_t^\text{4K} - \sqrt{1 - \alpha_t} \ y^\star\right) - x_{\text{prior}}\right]\right\|_2^2 + \ell_\text{MD}(y^\star; t),
\end{equation}
where $\alpha_t=\prod_{i=1}^t (1 - \beta_i)$ and $\beta_t$ are the schedule variances, and the one-step approximation is given by $\frac{1}{\sqrt{\alpha_t}}\left(x_t^\text{4K} - \sqrt{1 - \alpha_t} \ y^\star\right)$. Next, we set the derivative of \cref{eq:eps-loss} to 0 to solve for the optimal noise output. Hence, the optimal $\epsilon$ noise is:
\begin{equation}
    y_\text{FD}(x_t^\text{4K}) = \frac{\sqrt{\frac{1 - \alpha_t}{\alpha_t}} \, \lambda \odot\left(\frac{1}{\sqrt{\alpha_t}}x_t^\text{4K} - x_\text{prior}\right) + \displaystyle\sum_{i=1}^n w_i \odot y_i}{\frac{1 - \alpha_t}{\alpha_t}\lambda + \displaystyle\sum_{i=1}^n w_i}.
\end{equation}
We adopt the same variable definitions, as in the main text, for the current noisy state $x_t^\text{4K}$, the prior, $x_\text{prior}$, the weighting tensors $w_i$, and the prior regularization strength $\lambda$. As in the flow-matching formulation, when $\lambda=0$, $y_\text{FD}$ reduces to $y_\text{MD}$.

\end{document}